%% file: iclr2027_conference.tex
\pdfoutput=1
\documentclass{article}
\usepackage[T1]{fontenc}
\usepackage{iclr2027_conference,times}
\input{math_commands.tex}

\usepackage{amsmath,amssymb,booktabs,enumitem,float,graphicx,listings,microtype,tabularx,tikz,hyperref,url}
\usetikzlibrary{arrows.meta,positioning,fit,calc}
\usepackage{pgfplots}
\usepgfplotslibrary{groupplots}
\pgfplotsset{compat=1.18}
\newcommand{\benchmark}{\textsc{unsatz}}
\newcolumntype{Y}{>{\centering\arraybackslash}X}
\title{\benchmark: Evaluating Scientific Discovery\\Without a Given Ansatz}
\author{Ethan Ji, Wei Wang \& Yizhou Sun \\
Department of Computer Science\\
University of California, Los Angeles\\
\texttt{\{eji,weiwang,yzsun\}@cs.ucla.edu}
}
\iclrfinalcopy
\begin{document}
\maketitle
\lhead{Preprint}

\begin{abstract}
Before a scientist fits any parameter, they must decide what the model contains: which quantities exist, which are hidden, how they interact, and how they are measured. Physicists call such a trial form an \emph{ansatz}. Benchmarks for automated equation discovery mostly supply it, by fixing the variables, a library of candidate terms, or a law to be modified. We introduce \benchmark, a benchmark in which an LLM agent must construct the ansatz itself. Each task is a simulated system, generated from a private seed as a program in a small modeling language: ordinary, delay, or partial differential equations, discrete maps, static and causal models, or stochastic processes, possibly with latent states, distorted sensors, and uninformative measurement channels.  The agent buys experiments within a budget, writes candidate programs in the same language, obtains their coefficients from the benchmark's fitter, and is scored only on how its final program predicts experiments it never saw: conditions outside the ordinary ranges, interventions, and longer runs.
Across 80 systems in 20 domains, nine language models solve between 21\% and 60\%.
On the 64 deterministic systems, the two strongest agents solve 39 and 31, and classical and neural baselines together 14. The difficulty lies in finding the structure, not in the data: in 55\% of the agents' failed investigations, the experiments they bought suffice once the true program is given with unknown coefficients.
Because mechanisms recur across systems, we also start agents on a related system from the program of a first system, either the one they discovered or the true one; agents holding a correct program solve the related system in 15 of 42 investigations, and agents without one in 5 of 64.
Stronger agents make fewer unscientific errors, such as programs that omit a measured channel; beyond these, agents that find a system's variables still miss how they act, and prefer revisions that lower the training error to ones that predict better.
\end{abstract}

\section{Introduction}
\label{sec:introduction}

\begin{figure}[t]
\centering
{\raggedright\small\textbf{(a) A physical interpretation}\par}\smallskip
\includegraphics[width=\linewidth]{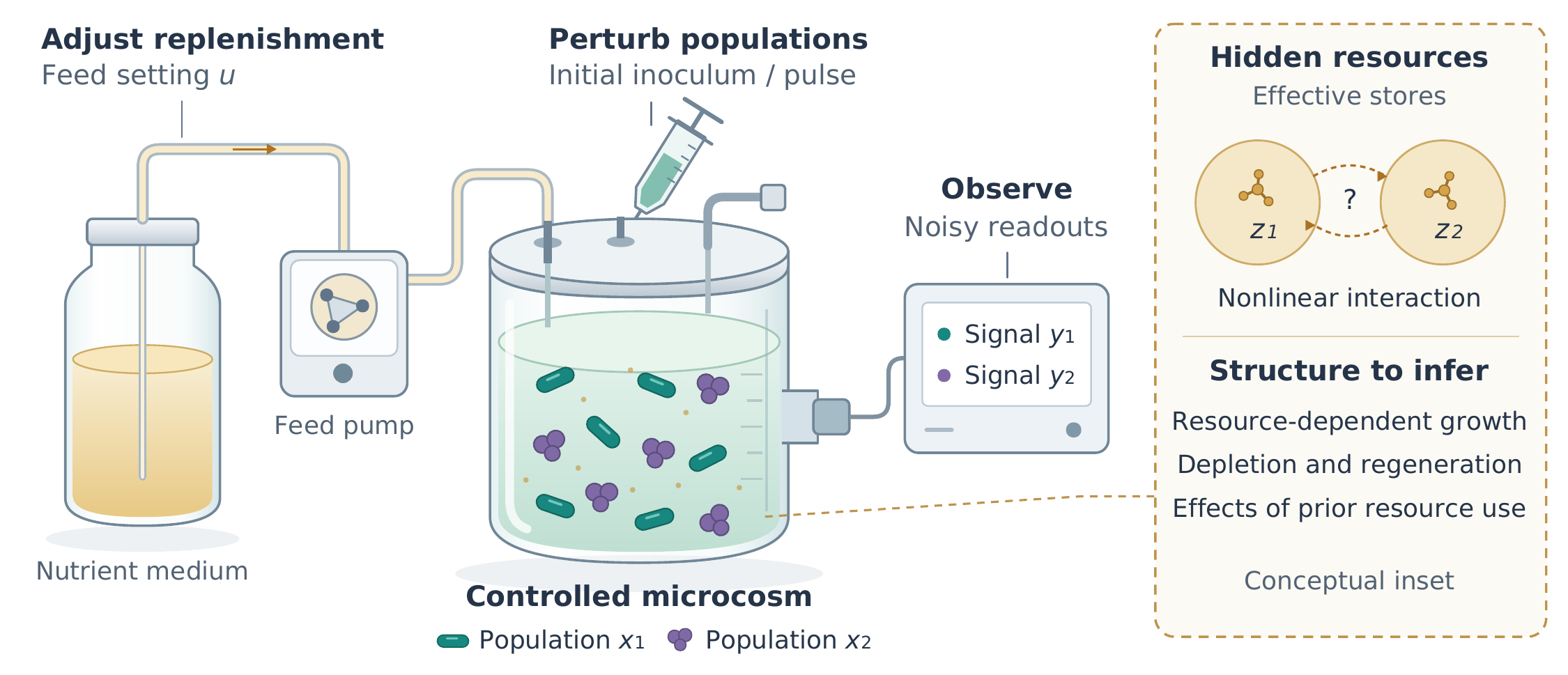}\par\medskip
\begin{minipage}[t]{0.3\linewidth}
{\raggedright\small\textbf{(b) Hidden structure}\par}\smallskip
\centering
\begin{tikzpicture}[font=\scriptsize, >={Stealth[length=1.5mm]},
  pop/.style={draw, circle, minimum size=6.5mm, inner sep=0pt, fill=blue!7},
  res/.style={draw, dashed, circle, minimum size=6.5mm, inner sep=0pt, fill=orange!12},
  sen/.style={draw, rounded corners=1pt, minimum width=6mm, minimum height=4mm, inner sep=1pt, fill=gray!10},
  lab/.style={font=\tiny, inner sep=1pt}]
\node (u) at (0, 1.75) {feed $u$};
\node[res] (z1) at (-1.15, 1.0) {$z_1$};
\node[res] (z2) at (1.15, 1.0) {$z_2$};
\node[pop] (x1) at (-1.15, -0.1) {$x_1$};
\node[pop] (x2) at (1.15, -0.1) {$x_2$};
\node[sen] (y1) at (-1.15, -0.85) {$y_1$};
\node[sen] (y2) at (1.15, -0.85) {$y_2$};
\draw[->] (u) -- (z1);
\draw[->] (u) -- (z2);
\draw[<->] (z1) -- node[lab, above] {exchange} (z2);
\draw[->] (z1) -- node[lab, left] {growth} (x1);
\draw[->] (z2) -- node[lab, right] {growth} (x2);
\draw[-{Bar[width=2mm]}, red!70!black] (x2) -- (z1);
\draw[-{Bar[width=2mm]}, red!70!black] (x1) -- (z2);
\node[lab, red!70!black] at (0, 0.1) {depletion};
\draw[->] (x1) -- (y1);
\draw[->] (x2) -- (y2);
\node[lab] at (0, -0.85) {dashed: hidden};
\end{tikzpicture}
\end{minipage}\hfill
\begin{minipage}[t]{0.68\linewidth}
{\raggedright\small\textbf{(c) Information available to the agent}\par}\smallskip
\fbox{\parbox{\dimexpr\linewidth-2\fboxsep-2\fboxrule\relax}{\scriptsize\raggedright\frenchspacing
\textbf{System message} (excerpt): A controlled ecological microcosm with 2 recorded population signals. You may choose the directly controllable starting populations, change the replenishment setting, apply an impulse, hold a population fixed, or purchase a wider-range run. Some recorded signals may be uninformative, and resource conditions may affect the dynamics. Budget: 243 units.\\[2pt]
\begin{tabular}{@{}l@{\enspace}l@{}}
\textbf{Agent} $\rightarrow$ \textbf{tool:} & \texttt{act(\{"action":\{"type":"observe"\}\})}\\
\textbf{Tool} $\rightarrow$ \textbf{agent:} & \texttt{\{"cost": 2, "remaining": 241.0,}\\
 & \texttt{\phantom{\{}"channels": ["y2", "y1"],}\\
 & \texttt{\phantom{\{}"t": [0.0, 0.3333, \ldots, 8.0],}\\
 & \texttt{\phantom{\{}"y": [[0.6738, 0.2380], \ldots], \ldots\}}
\end{tabular}}}
\end{minipage}
\caption{An ecology world. (a) A physical interpretation, for the reader. (b) The hidden structure of its true program: each population $x_i$ grows on a store $z_i$ that the feed $u$ replenishes and the \emph{other} population depletes (loss and turnover terms omitted). (c) The information available to the agent: an excerpt of its prompt and its first experiment (Appendix~\ref{app:ecology-prompt}).}
\label{fig:physical-ecology}
\end{figure}
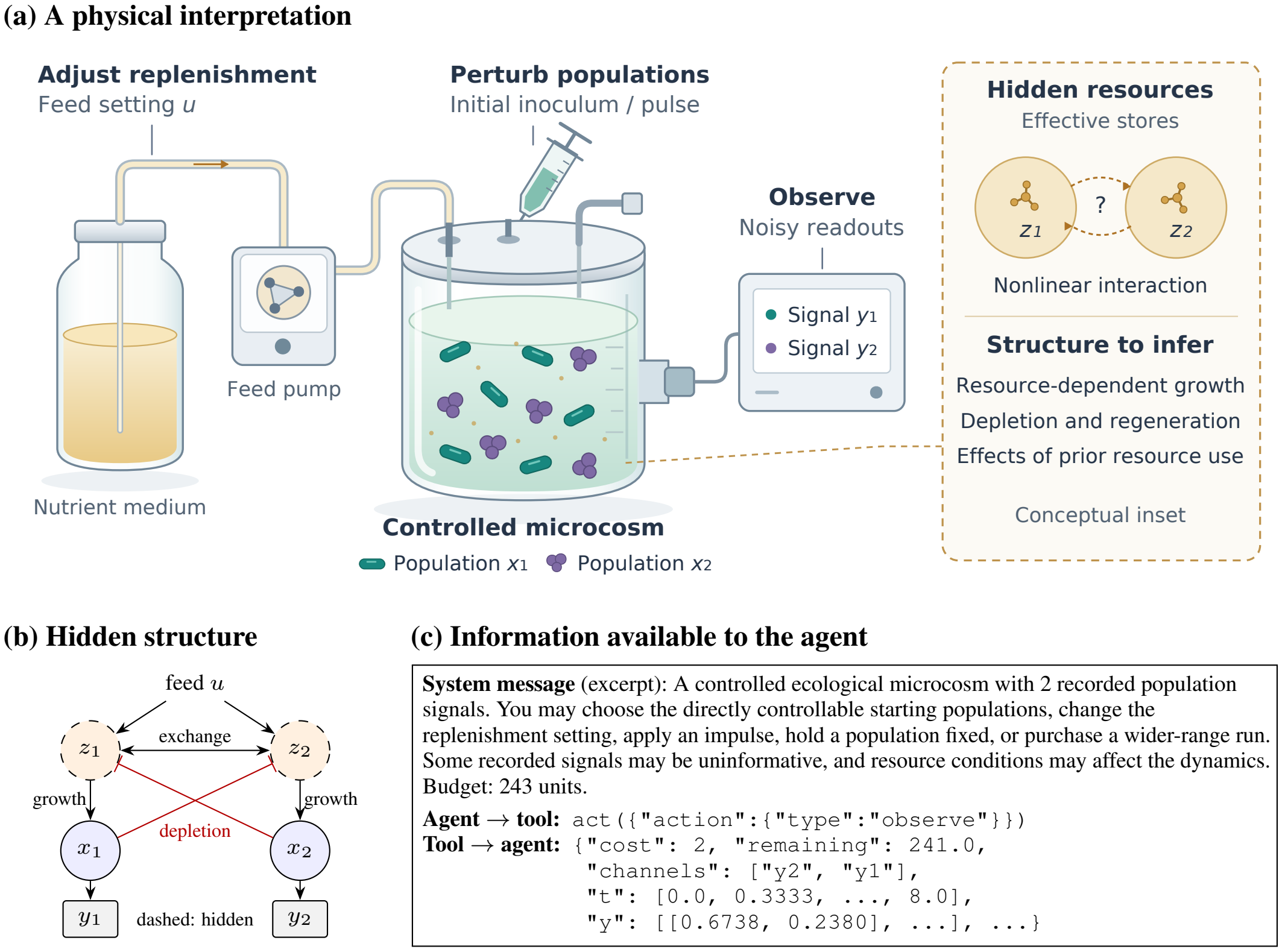

Consider a scientist presented with an unfamiliar ecological microcosm (Figure~\ref{fig:physical-ecology}). Two populations can be measured over time, the rate of a nutrient feed can be set, and only a limited number of runs is affordable. Before estimating a single coefficient, the scientist must decide what the model contains. Are two population variables enough, or is an unmeasured resource needed to explain why growth slows? Do the populations compete directly, or only through that resource? Does their response to the resource saturate? Experiments can answer these questions, since a pulse added to one population separates direct from resource-mediated interaction and a long run exposes slow depletion, but each experiment has to be chosen with a hypothesis in mind. Only once these decisions are made does fitting become a well-posed estimation problem. In the world of Figure~\ref{fig:physical-ecology}, the answer is two hidden resources, each feeding one population and depleted by the other, so that the populations compete only through them; the agent's prompt reveals none of this.

System identification calls this choice the selection of a model structure, as distinct from the estimation of its parameters \citep{ljung1999system}, and physicists call such a trial form an \emph{ansatz}. We use the latter term for the structure as a whole: which variables exist, which are latent, how they interact, and how they produce the measurements. Proposing a good ansatz is central to scientific modeling, yet evaluations of automated discovery largely take it as given. Symbolic-regression benchmarks provide datasets whose variables are already chosen \citep{udrescu2020aifeynman,lacava2021srbench,shojaee2025llmsrbench}, and sparse identification of nonlinear dynamics (SINDy) selects equations from a fixed library of candidate functions \citep{brunton2016sindy}. Interactive benchmarks let the agent design experiments, but draw their targets from a fixed set of probabilistic models, from mutations of canonical laws, or from textbook problems \citep{gandhi2025boxinggym,zheng2026newtonbench,chen2025physgym}, and the closest, DiscoverPhysics, asks for force laws in 22 particle worlds \citep{wiemann2026discoverphysics}. In each case, a person wrote or selected the hidden systems, and in most the variables, or the form the answer can take, are fixed in advance (Table~\ref{tab:comparison}).

\begin{table}[t]
\caption{Benchmarks for discovering scientific models, compared with \benchmark{}.}
\label{tab:comparison}
\centering\scriptsize\setlength{\tabcolsep}{4pt}
\begin{tabularx}{\linewidth}{@{}l l l l X@{}}
\toprule
Benchmark & Data & Hidden systems & Variables & Scored on \\
\midrule
SRBench & Fixed & Known equations, real datasets & Given & Symbolic match, test fit \\
LLM-SRBench & Fixed & Rewritten and synthetic equations & Given & LLM-judged match, out-of-domain error \\
BoxingGym & Chosen, $\le$10 & 10 models from the literature & Given & Prediction error, explanation \\
DiscoveryWorld & Actions, $\le$1{,}000 & 24 task templates, 5 seeds each & Given & Task completion, LLM-graded report \\
NewtonBench & Chosen, $\le$200 & 108 mutations of 12 canonical laws & Given & LLM-judged match, data error \\
PhysGym & Chosen, $\le$100 & 97 PHYBench physics problems & Given or masked & Symbolic or LLM-judged match \\
Gravity-Bench & Chosen, $\le$100 & 16 simulated binary systems & Given & Answers within tolerance \\
DiscoverPhysics & Chosen, $\le$16 rounds & 22 non-canonical N-body worlds & Some hidden & Held-out trajectories, LLM rubric \\
\benchmark{} & Priced, budgeted & Unlimited, from private seeds & Partly latent & Out-of-range, intervention, long-run tests \\
\bottomrule
\end{tabularx}

\smallskip\parbox{\linewidth}{\scriptsize \emph{Partly latent}: the agent is told the measured channels and inputs, but not how many states lie behind them, which are latent, how the sensors read them, or which channels are uninformative. \emph{Some hidden}: DiscoverPhysics measures the positions and velocities of the visible particles, and some worlds add hidden species or sources.}
\end{table}

This gap matters because large language model (LLM) agents are increasingly built as research assistants that design experiments, analyze data, and propose hypotheses \citep{boiko2023coscientist,lu2024aiscientist,chen2025scienceagentbench}. Whether they can propose an ansatz when neither the variables nor the form of the answer is given is unknown, and measuring it requires a benchmark with four properties. (i)~The equations that generate the system must be known to the evaluator, absent from the agent's training data, and expressible by the agent. (ii)~A failure must be attributable to the agent's structure or experiments, not to the way its coefficients were estimated. (iii)~A model must be judged by its predictions rather than its formula, since different formulas can describe the same system. (iv)~Experiments must cost enough for their choice to matter, yet suffice for a good investigator.

\benchmark{} provides these properties as follows. It generates each system, which we call a \emph{world}, from a private seed as a program in a small modeling language, and LLM agents write their hypotheses in the same language~(i). One fitter estimates the coefficients of every proposed program, and a world is accepted only if its true program, fitted to affordable experiments starting from its true coefficients, passes its hidden tests~(ii). These tests alone score the final model: conditions outside the ordinary ranges, interventions, including some the agent could not perform, and runs several times longer than a standard run~(iii). Experiments are priced, and each budget is a small multiple of the cost of a set of experiments from which the true program can be fitted~(iv). The benchmark thus isolates three steps of an investigation (choosing experiments under a budget, proposing a model that may involve latent states, and revising it against data) and holds the modeling language and the fitter fixed, so that success and failure at these steps can be attributed.

Discovery also accumulates across systems. The same saturating law describes enzyme kinetics, nutrient uptake by microbes, and the feeding of predators, and a scientist who has found a mechanism in one system tries it in the next, but must drop it where it does not act. \benchmark{} therefore also contains \emph{transfer problems}, which embed one generated mechanism in a first and a related world. After an agent investigates the first, a fresh instance of it investigates the related world, starting from the program the first investigation produced or from the true program of the first world. Two comparisons show whether a gain comes from the mechanism itself: a fresh agent on the related world without a program, and a control world that differs from the related one only in lacking the mechanism, where a program should not help.

We evaluate nine language models on 80 worlds from 20 domains and on four transfer problems (36 transfer studies).
GPT~6 Astra solves 48 of the 80 worlds and Claude Fable~5.1 solves 38, each significantly more than the next agent on a paired test; the remaining seven solve between 17 and 24 and are not separable from one another (Table~\ref{tab:headline}; Section~\ref{sec:llm-results}).
On the 64 deterministic worlds, GPT solves 39 and Claude 31, and classical and neural baselines, given up to three datasets per world, together solve 14. Most of the agents' failures are failures of structure, not of data: in 268 of their 490 failed episodes, the true program with unknown coefficients, fitted to the experiments the agent bought, passes the hidden tests, whereas in worlds GPT solved, a failing agent's program fitted to GPT's experiments passes in 11 of 197.
Stronger agents make fewer unscientific errors, such as final programs that omit a measured channel, but for every agent most failures are scientific: agents that find the latent states or the interaction graph of a system still miss the equations that govern them, and agents select revisions by training error rather than by prediction.
Recording the latent states does not make a world easy: when four agents repeat 18 dynamical worlds with every latent state recorded, 55 of their 72 episodes still fail.
A correct program for one world, whether the agent discovered it or was given it, helps on a related world that shares its mechanism: agents holding one solve the related world in 15 of 42 episodes and agents without a correct one in 5 of 64, and on a control world that lacks the mechanism the counts are 2 of 42 and 1 of 64.

\section{The \benchmark{} Benchmark}
\label{sec:benchmark}
\begin{figure}[t]
\centering
\begin{tikzpicture}[
  font=\scriptsize, >={Stealth[length=1.5mm]},
  box/.style={draw, rounded corners=2pt, align=center, inner sep=3pt, fill=#1},
  box/.default=white,
  note/.style={font=\scriptsize\itshape, align=center, inner sep=1pt}]
\node[box=gray!8] (world) {\textbf{World} (from a private seed)\\[1pt]
  \emph{hidden:} program $P^\star$, coefficients $\theta^\star$,\\ tests $T_1,T_2,T_3$\\[1pt]
  \emph{public:} channels, settings,\\ experiment prices, budget};
\node[box=blue!6, right=2.5cm of world] (agent) {\textbf{Agent}\\ language model};
\node[box=orange!8, right=2.1cm of agent] (score) {\textbf{Scoring}\\[1pt]
  $E=\max_\tau E_\tau$ over\\ out-of-range, intervention,\\ and long-horizon tests;\\ solved if $E<0.1$};
\node[box, below=0.7cm of agent] (fitter) {\textbf{Fitter}\\ $\hat\theta(P,\mathcal D)$ and $L(\hat\theta)$};
\draw[->] ([yshift=3pt]agent.west) -- node[note, above=1pt] {experiment, price} ([yshift=3pt]world.east);
\draw[->] ([yshift=-3pt]world.east) -- node[note, below=1pt] {noisy data} ([yshift=-3pt]agent.west);
\draw[->] ([xshift=-5pt]agent.south) -- node[note, left=2pt] {fit or submit $P$} ([xshift=-5pt]fitter.north);
\draw[->] ([xshift=5pt]fitter.north) -- node[note, right=2pt] {coefficients, training error} ([xshift=5pt]agent.south);
\draw[->, dashed] (agent.east) -- node[note, above=1pt] {last submission} (score.west);
\draw[->, dotted] (world.south) |- ($(fitter.south)+(0,-0.3)$) -| (score.south);
\node[note, fill=white] at ($(fitter.south)+(0,-0.3)$) {hidden tests, never shown to the agent};
\end{tikzpicture}
\caption{An episode. The agent buys experiments, the fitter estimates the coefficients of its programs, and its last submission is scored on the hidden tests.}
\label{fig:overview}
\end{figure}
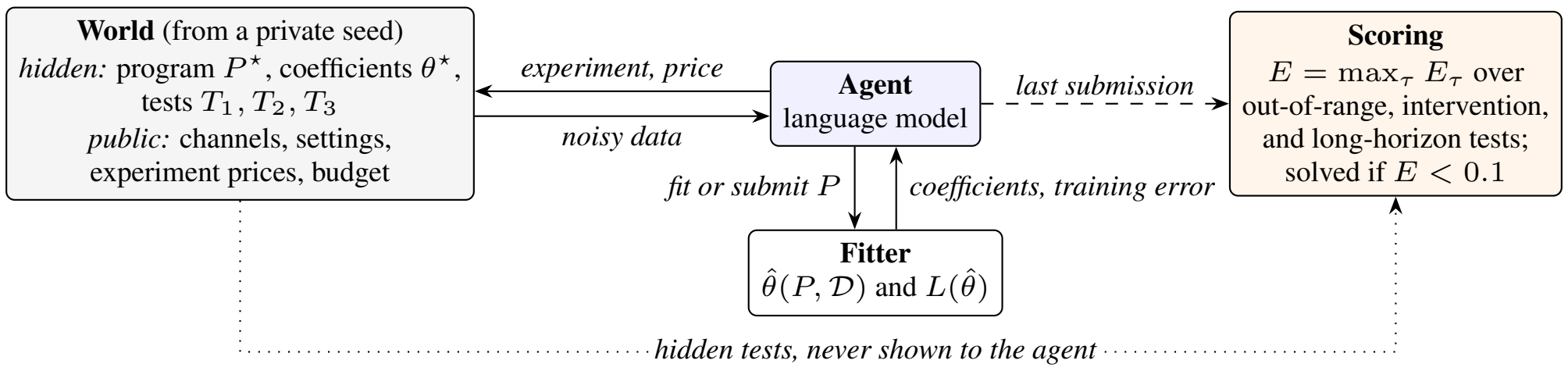
An \emph{episode} is one agent's investigation of one world (Figure~\ref{fig:overview}): the agent reads the world's description, buys experiments, receives fitted coefficients for the programs it proposes, and commits a final program, scored afterwards on tests it never sees.

\subsection{Worlds}
\label{sec:formulation}
A world consists of a hidden program $P^\star$ with coefficient values $\theta^\star$, a public description, and hidden test experiments. We call $P^\star$ with its coefficients left free the world's \emph{true structure}: the correct ansatz, which leaves only $\theta^\star$ to be estimated.
Most worlds are dynamical systems in state-space form
\begin{equation}
    \dot x=f(x,z,u;\theta),\qquad
    \dot z=g(x,z,u;\theta),\qquad
    y=h(x,z;\theta)+\varepsilon,
\label{eq:system}
\end{equation}
where the dot is the derivative with respect to time, $x$ are states that experiments can prepare or perturb, $z$ are \emph{latent} states that no channel reads directly, $u$ are inputs set by the experimenter, and $\theta$ are coefficients. The \emph{observation map} $h$ turns the states into the measured signals, or \emph{channels}, $y$, which carry Gaussian sensor noise $\varepsilon$. Other worlds replace the differential equations by discrete-time maps or by delay or stochastic differential equations (SDEs); in spatial worlds, $x$ and $z$ are fields over one dimension and $f$ and $g$ contain spatial derivatives (partial differential equations); the remaining worlds are static causal and input--output models, or jump, agent-based, and lattice processes (Table~\ref{tab:domains}). If $f$, $g$, and $h$ were known, the task would reduce to estimating $\theta$; the agent must instead find these functions, including the latent states they involve.

The description, which forms the agent's system message (Figure~\ref{fig:physical-ecology}c; Appendix~\ref{app:ecology-prompt}), contains a two-sentence narrative of the setting, the channels and inputs with their settable ranges, the standard initial condition and run length, the experiment types with their prices, the budget, the grammar of the modeling language, and the scoring rule. It states that every channel is noisy and that some may be uninformative: in nine domains, a world can contain up to two unlabeled, unscored auxiliary channels carrying only noise or a transformed or delayed copy of a real signal. It does not reveal the equations, the coefficients, the latent states or their number, the noise level, or which channels are auxiliary.

\subsection{Experiments}
\label{sec:experiments}
Each experiment is a fresh simulation of the hidden system and returns noisy measurements of every channel at fixed times. A dynamical world offers up to seven types (Appendix~\ref{app:tools}): (1)~a \emph{standard run} from the standard initial condition, (2)~a run from initial values of the measured states chosen within an ordinary range, (3)~a run at a chosen value of an input, (4)~a run in which one measured state receives an \emph{impulse} of chosen size at a chosen time, (5)~a run in which one measured state is \emph{clamped} at a chosen value over a chosen interval, (6)~a \emph{probe}, which may start from a wider range of initial values and last up to three times as long as a standard run, and (7)~an \emph{assay}, which reads the exact value of one latent state at one time of the most recent run. Each type answers a different question. A clamp cuts the feedback through the held state, so the response of the others shows how that state acts on them, and an impulse shows how a perturbation spreads and decays. Probes and input changes expose nonlinearities and slow processes that the ordinary range hides, such as saturation or the depletion of a resource. Assays show whether a latent state exists and, repeated at several times, how it evolves; a request for a latent state beyond the last one returns nothing, so even the number of latent states can be learned, at a price. Assay values guide the agent's reasoning but do not enter the fit. Static worlds answer queries at chosen inputs, causal worlds return samples of all variables with or without one of them clamped, and stochastic worlds return ensembles of independent paths. Each experiment has a price in budget units that depends on its type and, for samples or paths, on their number and duration. The budget caps the total price of the experiments an agent buys, and fits and submissions are free.

\subsection{The modeling language}
\label{sec:dsl}
The agent states each hypothesis as a program in the modeling language that also defines the worlds. A program names its class, declares its states, which of them are latent, its inputs, and the coefficients to be fitted, and gives one evolution equation per state and one observation equation per channel it predicts. For instance, the program
\begin{center}\small
\begin{tabular}{@{}l@{\qquad\qquad}l@{}}
\texttt{class ODE} & \texttt{param alpha, beta, gamma}\\
\texttt{state a, r} & \texttt{d a/dt = alpha*a*r - beta*a}\\
\texttt{latent r} & \texttt{d r/dt = u - gamma*a*r}\\
\texttt{input u} & \texttt{obs y1 = a}
\end{tabular}
\end{center}
proposes that a measured population $a$ grows by consuming a latent resource $r$, which is replenished at the feed rate $u$; \texttt{obs y1 = a} states that channel \texttt{y1} reads $a$ directly, so that an impulse or a clamp applied to \texttt{y1} acts on $a$. Expressions combine arithmetic with a fixed vocabulary of responses, namely signed powers, saturations, exponentials, logarithms, hyperbolic tangents, sines, and threshold switches, and some classes add delays, spatial derivatives, event rates, or interaction energies (Appendix~\ref{app:dsl}). A program may have at most 16 fitted coefficients, including the initial value of each latent state, which no experiment can set. The language fixes this vocabulary but not the model: the number of states, which of them are latent, how they interact, and how they are measured are the agent's decisions. Because every world is itself a program in this language, a correct ansatz is always expressible, and because programs are executable, the scorer can simulate any hypothesis under experiments that were never run. Expressibility also makes a failure interpretable: the true program can be fitted to any dataset an agent bought, which separates a failure of the structure from a failure of the experiments (Section~\ref{sec:classical}).

\subsection{The fitter}
\label{sec:fitter}
Coefficients are always estimated by the fitter. When it fits or submits a program $P$, the fitter simulates $P$ under every experiment $e$ in the purchased dataset $\mathcal D$ and chooses the coefficients $\theta$ that minimize the training error
\begin{equation}
    L(\theta)=\frac1N\sum_{e\in\mathcal D}\sum_{c,t}\operatorname{clip}_{[-10,10]}\!\left(\frac{\hat y_c(t;P,\theta,e)-y_c(t;e)}{s_c}\right)^{\!2},
\label{eq:fit}
\end{equation}
where $c$ ranges over the channels that $P$ predicts, $s_c$ is the scale of channel $c$ in $\mathcal D$ (its standard deviation plus 5\% of its mean absolute value), and $N$ is the number of residuals. The scaling makes channels of different magnitude count equally, and the clipping keeps one diverging simulation from dominating the fit. The minimizer is trust-region least squares \citep{virtanen2020scipy} from six starts of at most 60 residual evaluations each, all random unless the agent supplies one. The agent receives the fitted coefficients, the training error, and the error on each experiment. Programs with random output are fitted to the distribution of the data instead (Appendix~\ref{app:tools}). The agent acts through three tools, which perform an action (an experiment, a fit, or a submission), re-read earlier responses, and end the episode, and it cannot run code of its own. The fitter's output is therefore its only quantitative feedback on a hypothesis, and the coefficients of every agent and every symbolic baseline are estimated in the same way, so a failure reflects the structure or the data rather than the estimation~(ii). A much larger fitting budget does not change the outcome (Section~\ref{sec:classical}).

\subsection{Scoring}
\label{sec:episode}
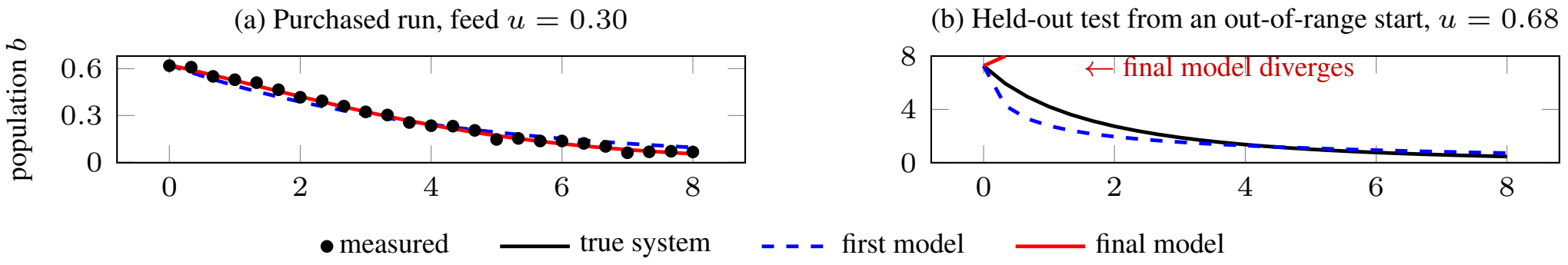
\begin{figure}[t]
\centering
\input{figures/selection-case.tex}
\caption{Two submissions of DeepSeek on an ecology world, showing the second population. (a)~On a purchased run at the lowest feed rate, both fit the data, and over all eight purchased runs the final model's training error is 76\% below the first's. (b)~On a hidden test starting both populations near 7.3, beyond the range the agent could prepare, the final model diverges.}
\label{fig:selection-case}
\end{figure}

Only the last submitted program is scored, after the episode, on hidden tests in up to three groups: ($T_1$)~runs from initial conditions or inputs outside the ordinary ranges, ($T_2$)~runs under interventions and altered conditions, including types and magnitudes of intervention that the agent could not buy, and ($T_3$)~runs several times longer than a standard run, ten times in the ODE and map worlds. These tests separate a mechanism from a curve fit. A wrong structure can fit the purchased data closely, by overfitting them or by agreeing with the truth only within the explored range, yet fail to predict how the system behaves far from those data, under an unfamiliar intervention, or over a much longer time (Figure~\ref{fig:selection-case}). For each group $\tau$,
\begin{equation}
    E_\tau=\frac{1}{|T_\tau|\,|C|}\sum_{e\in T_\tau}\sum_{j\in C}\frac{\operatorname{RMSE}_t\bigl(\hat y_j(t;e),\,y^\star_j(t;e)\bigr)}{n_{\tau j}},\qquad
    E=\max_\tau E_\tau,
\label{eq:error}
\end{equation}
averages, over the group's experiments and the real channels $C$, the root-mean-square error of the prediction $\hat y_j$ against the noiseless true trajectory $y^\star_j$, divided by the spread $n_{\tau j}$ of that channel over the group (its standard deviation plus 5\% of its mean absolute value). The world's error $E$ is the worst of its groups, and a deterministic world is \emph{solved} when $E<0.1$: even on its hardest group of tests, the model's error is below a tenth of the system's own variation. A program that diverges, omits a real channel, or is never submitted fails. Stochastic worlds are scored on distributions: at checkpoints along each test run, the continuous ranked probability score (CRPS) of each predicted channel may exceed that of the true distribution only by a small margin \citep{gneiting2007scoring}, and in SDE worlds the predicted probability of crossing a fixed barrier must be close to the true one \citep{brier1950verification} (Appendix~\ref{app:scoring}). Because the score compares behavior rather than formulas, any program equivalent to the true program passes.
Of the 230 final programs that pass the hidden tests, 224 differ from the true program in their terms, 200 express at least one dependence of the true program through a different type of term, and 25 have a different number of latent states (Appendix~\ref{app:programs-truth}).

\subsection{World generation}
\label{sec:generation}
A \emph{domain} is a generator of worlds; each seed draws a new program from it, so the worlds of a domain share a setting but not their equations. Table~\ref{tab:domains} lists the 20 domains, grouped by the kind of system. In five of them (mechanics, kinetics, discrete dynamics, causal systems, and scaling relations), the generator draws a random interaction graph and writes each equation as a sum of one to four products of responses from the language's vocabulary, applied to the variables that feed it, with log-uniform coefficient magnitudes and random signs; up to two states are latent, and the sensors may mix or saturate states. The other domains instantiate templates of their own, such as the consumer--resource dynamics of the ecology world or random directed graphs of 5, 12, or 32 nodes for the networks, whose response functions are again drawn from the vocabulary (Appendices~\ref{app:scenarios} and~\ref{app:extensions}). In the three spatial domains, the agent also chooses each run's spatial extent, boundaries, initial and forcing profiles, and sensor positions. A candidate world is discarded if its simulations are numerically unstable or its channels barely vary.
In twelve domains and in the transfer problems it is also discarded if its equations match one of 300 named laws, such as the Lotka--Volterra equations, up to the coefficients and the naming of the variables. The comparison rarely acts: about one candidate in 20{,}000 from the kinetics and mechanics generators is a named law (Appendix~\ref{app:named-laws}).

\label{sec:calibration}
Budgets are calibrated against the true program, so that a good investigator can succeed but must still choose. For each candidate world, a procedure that knows the true program buys, at the world's prices, designs of 8 to 60 experiments that mix standard runs, probes, input changes, and interventions, and fits the true program to each design, starting from the true coefficients, under three noise seeds. The cheapest design on which all three fits reach $E\le0.08$, a margin below the pass threshold, defines the world's reference cost, and the agent's budget is that cost times a factor drawn uniformly from $[2.5,4]$. The world is accepted only if the true program, fitted to a design that spends the whole budget (the \emph{calibration design}), again reaches $E\le0.08$ on five further noise seeds. A budget thus pays for two and a half to four sufficient designs, which leaves room for exploration and for a wrong first hypothesis, while every experiment still competes with the others. Every accepted world is therefore solvable within its budget by an investigator who finds its true structure and starts the fit near the right coefficients; fits from random coefficients are examined in Section~\ref{sec:classical}.

\subsection{Transfer problems}
\label{sec:transfer}
A transfer problem has three ODE worlds. The generator draws an expression subtree, the \emph{mechanism}, and embeds it in a \emph{first world} and in a \emph{related world}, where it acts on different variables within different equations; a \emph{control world} is identical to the related one except that the mechanism is replaced. In a transfer study, one agent investigates the first world, and the related and control worlds are each investigated from three starting points with equal budget and noise seed: by a fresh instance of the agent given the first investigation's final program as a starting hypothesis, by one given the true program of the first world in the same way, and by one that starts without a program.

\begin{table}[t]
\caption{Domains by kind of system. ODE, SDE, DDE, and PDE denote ordinary, stochastic, delay, and partial differential equations; S1 and S2 are the 12- and 32-node network domains. $^\dagger$ marks the eleven domains on which the known-structure baseline runs on a random design (Section~\ref{sec:setup}).}
\label{tab:domains}
\centering\scriptsize\setlength{\tabcolsep}{3pt}
\begin{tabularx}{\linewidth}{@{}l l X@{\hspace{10pt}}l l X@{}}
\toprule
Domain & Class & What must be inferred & Domain & Class & What must be inferred \\
\midrule
\multicolumn{3}{@{}l}{\emph{Deterministic dynamics}} & \multicolumn{3}{l@{}}{\emph{Static and causal systems}} \\
Mechanics$^\dagger$ & ODE & Forces, driving, and couplings & Causal$^\dagger$ & Causal & Dependencies under intervention \\
Kinetics$^\dagger$ & ODE & Reactions with hidden species & Scaling$^\dagger$ & Static & Relations via intermediate quantities \\
Discrete$^\dagger$ & Map & Nonlinear update rules & Metrology$^\dagger$ & Static & A latent response, two sensors \\
Ecology$^\dagger$ & ODE & Two hidden resource stores & Spectroscopy$^\dagger$ & Static & Peaks under four settings \\
Network$^\dagger$, S1$^\dagger$, S2 & ODE & Directed couplings of nodes & \multicolumn{3}{l@{}}{\emph{Stochastic systems}} \\
Reservoir & ODE & Exchanges with hidden stores & Stochastic dyn.$^\dagger$ & SDE & Drift, noise, barrier crossing \\
Delay & DDE & Delayed feedback and its lag & Jump & Jump & Event rates and state changes \\
Transport & PDE & Fluxes, diffusion, losses & Agents & Agent rule & Update rules on a graph \\
Reaction--diffusion & PDE & Reactions and diffusivities & Lattice & Spin lattice & Local interaction energies \\
Wave & PDE & Restoring and damping forces & & & \\
\bottomrule
\end{tabularx}
\end{table}

\section{Experimental Setup}
\label{sec:setup}
\paragraph{Worlds.} For every domain we use the first four worlds that pass generation and calibration, from private seed streams fixed before any agent ran on them (Appendix~\ref{app:named-laws}). All agents see the same 80 worlds and four transfer problems with the same budgets and measurement-noise seeds, so they differ only in the experiments they choose and the programs they propose.

\paragraph{Agents.} We evaluate GPT~6 Astra, Claude Fable~5.1, Muse Spark~1.3, GLM~5.3, Grok~4.6, Kimi~K3, Qwen~3.8 Max, Gemini~3.8 Flash, and DeepSeek~v4.1 Flash, each at its highest supported reasoning effort with default sampling parameters (Appendix~\ref{app:configuration}). The whole conversation stays in context, and an episode ends when the agent finishes or reaches its context or output limit. Each agent runs one episode per world and up to five per transfer problem, 896 episodes in total.
Three further experiments add 544 episodes: second and third episodes, under new measurement-noise seeds, of DeepSeek and Gemini on every world and of GPT and Claude on one world of each domain (Appendix~\ref{app:repeats}); 18 worlds rerun with their latent states recorded (Section~\ref{sec:case-study}); and transfer studies that start from the true program (Section~\ref{sec:extension-results}).

\paragraph{Baselines.} Classical methods cannot choose experiments, so each receives up to three datasets per world, all bought within the world's budget: the calibration design of Section~\ref{sec:calibration}, the experiments GPT bought in its episode, and a random design that buys affordable experiment types with uniformly drawn settings.
Nine methods run on the 64 deterministic worlds, each where it applies: sparse regression in four variants (SINDy with a polynomial library, a library of the language's responses with and without a latent coordinate, and PDE-FIND), evolutionary symbolic regression (PySR), our grammar search over complete programs of the modeling language including latent states, a multilayer perceptron, and neural ODEs with and without latent coordinates.
A baseline's program is submitted, refitted, and scored as an agent's is; the counts of solved worlds also include smaller PySR and grammar-search runs on the random design (Appendices~\ref{app:classical} and~\ref{app:classical-all}). The \emph{known-structure baseline} fits the true structure, from random starting coefficients, to the random design on the 44 worlds of the eleven domains marked in Table~\ref{tab:domains} (40 deterministic and 4 SDE; Appendix~\ref{app:classical}) and, for every failed episode, to the experiments the agent bought (Appendix~\ref{app:experiments-hypothesis}). It is a handicapped reference, not an upper bound: unlike the agents, it cannot choose starting coefficients.

\section{Results}
\label{sec:results}
\subsection{The structure is the bottleneck}
\label{sec:classical}
Structure search without an agent rarely finds a model that predicts. On the calibration design under the first of its three noise seeds, the baselines together solve 3 of the 64 deterministic worlds, and over all datasets and noise seeds 14; GPT solves 39 and Claude 31 in one episode per world (Appendix~\ref{app:classical-all}). The baselines keep up only where no dynamics must be modeled: on the 16 static and causal worlds they solve 10, against 9 each for GPT and Claude, and on the 48 dynamical worlds 4, against 30 and 22. Of the baselines' failing submissions, 96\% fail the out-of-range tests.

Data are not what the agents' failed investigations lack. For each of their 490 failed episodes, we fit the true structure to the experiments the agent bought (Appendix~\ref{app:experiments-hypothesis}). It passes the hidden tests from random starting coefficients in 268 episodes (55\%; 95\% interval 44--65\% when worlds are resampled) and from the true coefficients in a further 180 (37\%); in the remaining 42 (9\%), the experiments do not determine the coefficients. In more than half of its failures, then, an agent held data from which a correct structure can be fitted. The same separation appears without the true program. In worlds that GPT solved, GPT's final structure, fitted to a failing agent's experiments, passes in 193 of 210 episodes, whereas the failing agent's structure, fitted to GPT's experiments, passes in 11 of the 197 with a final program (136 and 7 from random starts alone).
Even a random design usually suffices: on the 44 marked worlds, the true structure passes from random starting coefficients on 27 of the 40 deterministic worlds and 3 of the 4 SDE worlds (Appendix~\ref{app:classical}). Nor does the fitting budget explain the failures: refitting the agents' 248 executable but failing final models on the 40 deterministic marked worlds with 24 starts of 200 evaluations turns none into a pass (Appendix~\ref{app:classical}).

\subsection{Two agents lead the ranking}
\label{sec:llm-results}
Table~\ref{tab:headline} summarizes the 720 episodes. GPT~6 Astra solves 48 of the 80 worlds and Claude Fable~5.1 solves 38; no other agent solves more than 24, and the top-two ranking holds on paired tests: GPT solves 12 worlds that Claude does not and Claude 2 that GPT does not (exact McNemar test, $p=0.013$), Claude outperforms Muse 16 to 2 ($p=0.001$), and no pair among the other seven differs significantly (smallest $p=0.09$). At least one agent solves 54 of the 80 worlds.
The benchmark measures ability in aggregate over its worlds: with the 20 domains fixed, the standard error of an agent's percentage solved on 80 worlds is 3.7 points and that of a difference between two agents 4.6 points, whereas GPT and Claude differ by 12.5 points and Claude and Muse by 17.5 (Appendix~\ref{app:ranking-stability}).
Over three episodes on all 80 worlds, DeepSeek solves 18, 18, and 16 and Gemini 17, 16, and 15; on one world of each domain, GPT solves 11, 13, and 13 and Claude 9, 8, and 12. A further episode has the outcome of the first on 88\% of worlds for DeepSeek and Gemini and on 78\% for GPT and Claude (Appendix~\ref{app:repeats}).
Spending does not separate agents: DeepSeek and Kimi use 90\% of their budgets on average, near GPT's 93\%, yet solve about a third as many worlds (Appendix~\ref{app:domain-results}).

\begin{table}[t]
\caption{Results of the nine agents on the 80 worlds. A world is solved when its final model passes every group of hidden tests; the interval is a 95\% Wilson score interval over independent worlds. ``Median'' is the median worst-group error $E$ and ``Invalid'' counts divergent, missing, or incomplete predictions (entered as $+\infty$ in the median), both over the 64 deterministic worlds. ``Truncated'' counts episodes that ended at the model's context or output limit. API costs are in Appendix~\ref{app:domain-results}.}
\label{tab:headline}
\centering\scriptsize\setlength{\tabcolsep}{3.5pt}
\input{tables/results.tex}
\end{table}

\begin{figure}[t]
\centering
\input{figures/outcome-profiles.tex}
\caption{Outcomes of the final models; ``finite error'' marks executable models that predict poorly.}
\label{fig:outcomes}
\end{figure}
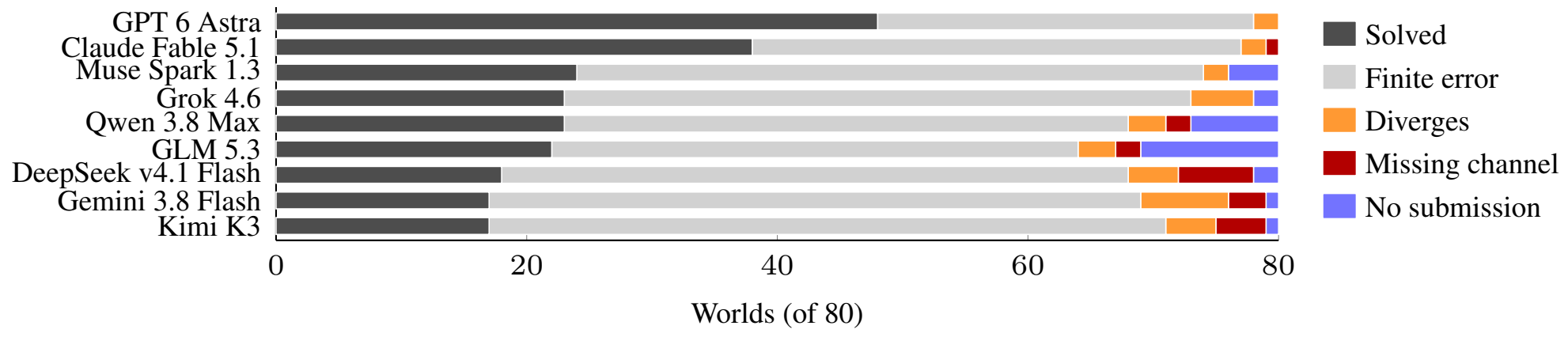

\subsection{Unscientific errors explain only part of the gap}
\label{sec:unscientific}
Unscientific errors are mistakes that have nothing to do with the system under study. GPT and Claude, the top two, end 1 of their 160 episodes with no submission or a final model that omits a measured channel; Gemini and Kimi, the bottom two, end 9 of 160 this way. The top pair solves 86 worlds and the bottom pair 34, so the 8 extra unscientific errors could close at most 8 of that 52-world gap; for every agent, most failures are models that predict poorly or diverge (Figure~\ref{fig:outcomes}).
In one spectroscopy world, Claude, Kimi, and DeepSeek all begin with programs that lack the declarations binding the four experimental settings to program variables (such as \texttt{obs field = fi}), so their predictions ignore the settings. Claude tests a binding on a minimal example and fixes its program (test error 0.016), whereas Kimi attributes the failure to the fitter, DeepSeek substitutes a hand-derived formula, and both submit constants they declare validated.

\subsection{Agents that find the variables still miss their equations}
\label{sec:case-study}
Performance varies strongly by domain (Appendix~\ref{app:domain-results}), and among the 48 deterministic dynamical worlds those with latent states are solved less often: excluding the networks, whose nodes are all measured, agents solve the worlds without a latent state in 71\% of episodes (96 of 135) and the worlds with at least one in 23\% (43 of 189), falling to 12\% (14 of 117) with two or three. To separate the effect of hiding a state from the other differences between worlds, we gave GPT, Claude, DeepSeek, and Gemini the 18 ODE and map worlds with latent states a second time, with every latent state recorded on an additional channel and nothing else changed.
With every latent state recorded, they solve 17 of the 72 episodes (24\%), against 8 with the latent states hidden (Appendix~\ref{app:latent-recorded}); on the worlds without a latent state, the same four agents solve 45 of 60 (75\%). Most of the difference between the two kinds of world therefore lies elsewhere than in the hidden states.
The networks are hard too: GPT solves 8 of their 12 worlds, the other eight agents 4 of 96.

The investigation records (Appendices~\ref{app:error-comparisons} and~\ref{app:qualitative}) locate the scientific errors: agents that find which variables a model needs and which interact still miss how they act, and the training error hides this.

\paragraph{Finding the latent states is not enough.}
On the same 18 worlds, all with assays, the nine agents buy an assay in 122 of 162 episodes and, in 76, ask for a latent state beyond the last one, which reveals how many exist; 69 final models declare the true number. Yet only 19 of the 162 episodes solve the world, and final models with the true number of latent states succeed no more often (8 of 69) than the others (11 of 93). Detecting latent states is the easier part of the ansatz; writing down how they evolve is harder. GPT's successful ecology episode shows what this requires: once its assays establish exactly two latent states, it fits one- and two-resource models side by side as experiments accumulate and submits one that predicts the truth (Figure~\ref{fig:physical-ecology}b) within tolerance ($E=0.091$), although its program and the true one differ in which population depletes which resource.

\paragraph{The right graph with the wrong equations.}
The same gap appears where every variable is measured. In a twelve-node network world, GPT and DeepSeek both recover all sixteen directed edges of the true graph. GPT submits nonlinear node dynamics with saturating couplings $\tanh(x_j-x_i)$ and reaches $E=0.094$; DeepSeek keeps linear dynamics and couplings and reaches 0.97. Recovering the interaction graph, the goal of causal discovery, is not enough to predict the system.

\paragraph{Every channel is treated as signal.}
The prompt warns that some channels may be uninformative, yet 115 of the 127 final models on worlds with auxiliary channels predict at least one of them. Modeling such a channel costs coefficients and latent states that the real dynamics then lack: every latent state in the four programs DeepSeek submitted in the episode of Figure~\ref{fig:selection-case} (two shown there) reproduces its third channel, the second population delayed by two sampling intervals, and none represents a resource. Deciding what not to model is part of an ansatz, and agents rarely make it.

\paragraph{The training error misleads revisions.}
DeepSeek's final ecology model, fitted to the same eight runs as its first, lowers the training error by 76\%, but its test error rises from 0.29 to 126 (Figure~\ref{fig:selection-case}). The revision replaces the stabilizing term $-0.054\,b^3$ of the second population $b$ by a feed-dependent quadratic, so that at the lowest feed rate $\dot b=-0.427\,b+0.441\,b^2$ grows without bound once $b>0.97$, which the purchased low-feed runs never reach; at the feed rate of the test in Figure~\ref{fig:selection-case}b the threshold is 1.8, and the test starts at 7.3.
Selecting by training error, the only quantitative feedback agents receive, rarely turns a solved world into a failure (7 of the 720 episodes), but in 235 of the 510 episodes with more than one submission the final model predicts worse than an earlier one.
On the 44 marked worlds, 48 of these episodes have the final and the better earlier model fitted to the same data, and in 35 of them the final model has the lower training error.

\subsection{Do agents reuse a discovered mechanism?}
\label{sec:extension-results}
Table~\ref{tab:transfer-true} (Appendix~\ref{app:transfer-true}) reports the 36 studies. In one, the first investigation submitted no program, and only the true-program starting point was run. Agents starting fresh solve the related world in 4 of 35 studies (95\% interval 5--26\%), agents given their own program in 6 of 35 (8--33\%), and agents given the true program in 10 of 36 (16--44\%); on the control world they solve 1, 0, and 2. An agent's own program helps when it is correct: where the first world was solved, it leads to a solved related world in 5 of 6 studies, against 2 of 6 for agents starting fresh; where it was not, in 1 of 29, against 7 of 30 for the true program. Pooled over the three starting points, agents that hold a correct program, their own after solving the first world or the true one, solve the related world in 15 of 42 episodes and agents without a correct one in 5 of 64 (permutation test within study, $p=0.002$); on the control world the counts are 2 of 42 and 1 of 64 ($p=0.44$). A correct mechanism helps where it acts.

\section{Related Work}
\label{sec:related}

\paragraph{Equation discovery.}
Bongard and Lipson reverse-engineered dynamical systems by choosing experiments, and Schmidt and Lipson distilled free-form laws from motion data \citep{bongard2007reverse,schmidt2009distilling}. SINDy selects equations from a library of candidate terms \citep{brunton2016sindy}, with extensions to spatial derivatives, learned coordinates, and delay embeddings \citep{rudy2017pdefind,champion2019coordinates,bakarji2023delay}. AI Feynman and PySR advance symbolic regression, SRBench and LLM-SRBench benchmark it, and LLM-SR and BoxLM bring language models to equation and program search \citep{udrescu2020aifeynman,cranmer2023pysr,lacava2021srbench,shojaee2025llmsr,shojaee2025llmsrbench,li2024boxlm}. Most start from fixed data with chosen variables.

\paragraph{Interactive scientific discovery.}
Bayesian experimental design chooses experiments for a given model family \citep{huan2013design}. Table~\ref{tab:comparison} compares \benchmark{} with the closest benchmarks \citep{jansen2024discoveryworld,gandhi2025boxinggym,zheng2026newtonbench,chen2025physgym,koblischke2025gravitybench,wiemann2026discoverphysics}. COAT proposes causal variables from unstructured data \citep{liu2024coat}. Like LiveBench, FrontierMath, and DyVal \citep{white2025livebench,glazer2024frontiermath,zhu2024dyval}, \benchmark{} counters contamination \citep{deng2024contamination} with fresh items from private seeds.

\section{Conclusion}
\label{sec:discussion}
\label{sec:conclusion}
We introduced \benchmark{}, a benchmark in which agents must construct the structure of a scientific model, and not only its coefficients, by experimenting on generated systems.
The strongest of nine language models solves 48 of 80 worlds.
Unscientific errors are rare for the strongest agents, but scientific errors remain for all: agents that recover a system's latent states or interaction graph still write inadequate equations for them and accept revisions that lower the training error. Both failures would be caught by one habit, testing a candidate where it could fail rather than where it was fitted.

\subsection*{AI use statement}
Generative AI tools assisted with software implementation, experiment orchestration, literature search, illustration, and manuscript preparation. The authors are responsible for the content, experimental claims, and accompanying software.

\subsection*{Ethics statement}
The benchmark uses simulated systems without human-subject or personal data. Model evaluation consumes commercial inference resources. The intended use is scientific capability measurement and methodological research.

\subsection*{Reproducibility statement}
The evaluation is determined by the generator, world and measurement seeds, experiment records, submitted programs, fitted parameters, and scorer. These records are retained for all reported episodes, and the generator, the evaluation code, the seeds of the evaluated worlds, the prompts, and the records of all 1{,}440 episodes will be released. Section~\ref{sec:benchmark} and Table~\ref{tab:protocol} specify the protocol, and the appendices give per-setting generators, tools, the modeling language, model configurations, and the submitted programs discussed in the text. Replaying a recorded scientific evaluation does not depend on reproducing a proprietary endpoint's future responses.

\subsection*{Acknowledgments}
This work was partially supported by NSF 2303037, 2312501, 2531008, 2609582, and 2617441; DARPA HR001126CE054; NIH OT2OD038003, U54OD036472, U54DK097771, and U54HG012517; the SRC JUMP 2.0 Center; CDSC Industry Partners; Amazon Research Awards; Amazon; Google; Google Gifts; NEC; Optum; and Snapchat.

\raggedbottom
\bibliography{iclr2027_conference}
\bibliographystyle{iclr2027_conference}
\appendix

\section*{Contents of the appendices}

{\footnotesize\setlength{\tabcolsep}{4pt}
\begin{tabularx}{\linewidth}{@{}l X@{}}
\toprule
\ref{app:protocol} & Protocol configuration \\
\ref{app:scenarios} & Scenario specifications of the first eleven domains \\
\ref{app:tools} & Agent tools and experimental actions \\
\ref{app:dsl} & The modeling language, and the named models within the limits of a program \\
\ref{app:extensions} & Delay, spatial, jump, agent-rule, lattice, reservoir and larger-network domains; the transfer problems, their results, and transfer with the true program \\
\ref{app:configuration} & Model configurations \\
\ref{app:scoring} & Continuous error definitions \\
\ref{app:raw-errors} & Per-world errors and calibration errors \\
\ref{app:classical} & Classical comparisons: every method on every applicable world, and the matched comparison on 44 worlds \\
\ref{app:error-comparisons} & Paired error comparisons \\
\ref{app:qualitative} & Submitted structures in the case studies \\
\ref{app:ecology-prompt} & Ecology prompt and observation example \\
\ref{app:domain-results} & Per-domain results, termination, and cost \\
\ref{app:failures} & Why episodes fail: experiments and hypotheses, final programs and the true program, latent states recorded \\
\ref{app:ranking-stability} & Uncertainty, stability of the ranking, and repeated episodes \\
\ref{app:named-laws-all} & Named laws in the generated worlds and in the agents' programs \\
\bottomrule
\end{tabularx}}

\section{Protocol configuration}
\label{app:protocol}
Table~\ref{tab:protocol} collects the constants of the protocol described in Section~\ref{sec:benchmark}.
\begin{table}[H]
\caption{Protocol configuration. Appendices~\ref{app:scenarios}--\ref{app:extensions} give per-domain details.}
\label{tab:protocol}
\centering\scriptsize
\begin{tabularx}{\linewidth}{@{}l X@{}}
\toprule
Component & Configuration \\
\midrule
Programs & At most 16 fitted parameters, including latent initial values; at most 400 expression nodes (1{,}200 for Network-S2) \\
Generators & 1--4 terms per equation and 1--2 factors per term; power exponents $\{-1,-\tfrac12,\tfrac13,\tfrac12,\tfrac23,1,\tfrac32,2,3\}$; up to 2 latent states; 0--2 auxiliary channels; signal-to-noise ratio log-uniform in $[30,1000]$ in mechanics, kinetics, discrete, causal, and scaling \\
Fitter & Trust-region reflective least squares, 6 starts of at most 60 residual evaluations each (finite-difference Jacobian columns are not counted); starting coefficients have random sign and log-uniform magnitude in $[0.03,8]$, latent initial values are uniform in $[-2,2]$, and an agent-supplied start replaces one of the six; stochastic processes use Nelder--Mead, 6 starts of 60 evaluations \\
Held-out tests & $(|T_1|,|T_2|,|T_3|)$: $(8,6,3)$ for ODE and map worlds, $(4,3,2)$ for delay and spatial worlds, $(3,1,1)$ for jump, agent, and lattice worlds; causal and scaling 10 and 6; metrology and spectroscopy 16 or 24 and 8; SDE 4 conditions; stochastic scores use 512 simulated paths per condition \\
Stochastic scores & Excess CRPS $\mathbb E|X-Y|-\tfrac12\mathbb E|X-X'|-\tfrac12\mathbb E|Y-Y'|$ for predicted $X,X'$ and true $Y,Y'$, divided by the true standard deviation (at least 0.1) and maximized over recorded times; excess Brier score $(\hat p-p)^2$ \\
Calibration & True-coefficient fits on designs of 8, 12, 16, 24, 32, 44, or 60 experiments on three noise seeds; five further seeds at the full budget \\
Long-horizon tests & Ten times the standard run in the ODE and map domains, four times in the delay and spatial domains, and three times in the jump, agent, and lattice domains \\
\bottomrule
\end{tabularx}
\end{table}

\input{appendix-protocol.tex}

\input{appendix-extensions.tex}

\section{Model configurations}
\label{app:configuration}
All runs use the same experimental interface and fitting harness. The highest supported reasoning setting is \texttt{xhigh} for Qwen, \texttt{high} for Gemini, and \texttt{max} for the other models. Sampling temperature, top-$p$, and provider seed use endpoint defaults. The context and output allowances are listed below; conversation history is retained until the agent finishes or an endpoint limit ends the episode.
\begin{table}[H]
\caption{Endpoint configurations. Token counts are context/output ceilings.}
\centering\scriptsize\setlength{\tabcolsep}{3pt}
\input{tables/configuration.tex}
\end{table}
\begin{table}[H]
\caption{Model identifiers used for inference.}
\centering\scriptsize
\input{tables/model-ids.tex}
\end{table}

\section{Continuous error definitions}
\label{app:scoring}
For each deterministic test group $\tau$, concatenate the generating system's noiseless trajectories and define the channel normalizer
\begin{equation}
    n_{\tau j}=\operatorname{std}(y^\star_{\tau j})
       +0.05\operatorname{mean}|y^\star_{\tau j}|+10^{-9}.
\end{equation}
For each hidden condition, compute timewise RMSE on each real channel, divide by $n_{\tau j}$, and average across channels. The group error $E_\tau$ averages these condition-level errors; Equation~\ref{eq:error} takes the worst group. Divergent simulations receive an internal failure penalty, which is separated from finite prediction errors in the reported continuous summaries. Missing submissions and invalid predictions remain in the evaluation denominator.

For stochastic systems, the evaluator simulates independent predicted and reference path ensembles with fixed evaluation seeds. Distributional error uses normalized energy/CRPS excess, and first-passage error uses excess Brier loss. The maximum excess over the corresponding hidden conditions determines each reported error. A stochastic world passes when both are below their respective thresholds, $0.04$ and $0.03$.

\section{Per-world errors and calibration errors}
\label{app:raw-errors}

\begin{table}[H]
\caption{Worst-group NRMSE for every deterministic model--world pair in the domains marked $^\dagger$ in Table~\ref{tab:domains}. World numbers identify the same paired instances across models. D, I, and N are the failure categories of Figure~\ref{fig:outcomes}: diverges, missing channel or invalid predictions, and no submission.}
\centering\scriptsize\setlength{\tabcolsep}{2.6pt}
\input{tables/deterministic-errors.tex}
\end{table}

\begin{table}[H]
\caption{Stochastic prediction errors in their native score units. Each world has a maximum energy/CRPS excess and maximum first-passage Brier excess; thresholds are 0.04 and 0.03. A dash denotes no submission.}
\centering\scriptsize\setlength{\tabcolsep}{3pt}
\input{tables/stochastic-errors.tex}
\end{table}

\begin{table}[H]
\caption{Calibration errors for deterministic worlds (Section~\ref{sec:calibration}). Each entry is the worst NRMSE across the five further noise seeds at the assigned budget, fitting the true program from its true coefficients on a design spanning the held-out regimes. The known-structure baseline of Section~\ref{sec:classical} instead starts from random coefficients on the random design. Values retain the precision recorded during calibration.}
\centering\small
\input{tables/calibration-errors.tex}
\end{table}

\section{Classical comparisons}
\label{app:classical}

Two comparisons are reported: every classical method on every world it applies to, on three datasets (Appendix~\ref{app:classical-all}), and a matched comparison on 44 worlds, with the known-structure baseline (Appendix~\ref{app:classical-matched}).

\input{appendix-classical-all}

\subsection{Matched comparison on 44 worlds}
\label{app:classical-matched}

Table~\ref{tab:classical} compares methods on their shared subsets of the 44 worlds in the domains marked $^\dagger$ in Table~\ref{tab:domains}. Each entry gives the pass count and median worst-group NRMSE; simulation failures and missing predictions enter the median as $+\infty$. The deterministic panel comprises 24 dynamical worlds, twelve static-function worlds, and four causal worlds. All classical methods on a given world share the same random-acquisition dataset, while language models choose their own experiments under the same acquisition budget.

\begin{table}[H]
\caption{Classical and language-model results on matched world subsets. Dashes denote settings outside a method's scope. All 512 grammar proposals are evaluated in each of its forty worlds.}
\label{tab:classical}
\centering\footnotesize\setlength{\tabcolsep}{3pt}
\input{tables/classical-comparison.tex}
\end{table}

PySR's median error on the static worlds is 0.146, compared with 0.084 for GPT and 0.113 for Claude. The known-structure baseline attains median NRMSE 0.019 over the 40 deterministic worlds. Its remaining failures reflect estimation from the random design with random starting coefficients, whereas calibration uses a design spanning the held-out regimes and starts from the true coefficients. The stochastic known-structure results are:

\begin{table}[H]
\caption{Known-structure baseline on stochastic worlds, using the shared random design and random starting coefficients.}
\centering\small
\input{tables/known-structure-sde.tex}
\end{table}

\paragraph{Fitter ceiling.}
To test whether the agents' failures reflect the fitter's budget rather than their structures, we refit every executable but failing final deterministic model of these 44 worlds (248 episodes) on the agent's own purchased data, reproducing each archived training error exactly, with 24 starts of at most 200 residual evaluations (the protocol uses 6 starts of 60) and the submitted coefficients as one of the starts. None passes the held-out tests; 12 improve their training error by more than 10\%, and 5 their held-out error.
Fitting the true program on each world's calibration design from random starts passes 24 of the 40 deterministic worlds with the protocol fitter and 26 with 24 starts of 200 evaluations, against all 40 from the true coefficients and 27 on the random design (Table~\ref{tab:classical}). Of the 16 protocol failures, 13 end above the truth's training error, so local minima rather than the design limit the known-structure baseline; the other three match it with coefficients the design does not pin down. The agents' submitted structures, by contrast, gain nothing from the larger budget.

\section{Paired error comparisons}
\label{app:error-comparisons}

\subsection{Final-outcome categories}
\begin{table}[H]
\caption{Error profiles across the 44 worlds of the domains marked $^\dagger$ in Table~\ref{tab:domains}. Categories are disjoint. ``Finite error'' denotes predictions exceeding a scoring threshold. ``Simulation'' denotes at least one divergent, nonfinite, or shape-invalid hidden prediction. ``Missing channel'' denotes an omitted required readout. An episode without a submission is classified separately.}
\centering\scriptsize\setlength{\tabcolsep}{3pt}
\input{tables/error-profiles.tex}
\end{table}

\subsection{Spectroscopy: executable verification and response to feedback}
The four models in Table~\ref{tab:spectroscopy-errors} investigate the same world with a 138-unit budget and finish voluntarily. Each final program is replayed at the settings in its own training acquisitions, using the exact submitted parameter vector. This replay reproduces the recorded fitting cost. Kimi's constant prediction spans observations from 0.0311 to 0.1966; DeepSeek's zero prediction spans observations from 0.0305 to 0.1759.

\begin{table}[H]
\caption{Paired spectroscopy comparison. Measurements are those available at the final submission. The prediction range is evaluated on those measurements; hidden error is worst-group NRMSE. The Kimi and DeepSeek programs omit the input/readout bindings and therefore ignore the experimental settings.}
\label{tab:spectroscopy-errors}
\centering\footnotesize\setlength{\tabcolsep}{3pt}
\input{tables/spectroscopy-error-comparison.tex}
\end{table}

The binding rule appears in the supplied language description: a direct readout links a channel to a state, so an intervention on that channel acts on that state. For static inputs, the following declarations connect the measured settings to the program:
\begin{lstlisting}[basicstyle=\ttfamily\small,frame=single]
obs frequency = fq
obs field = fi
obs temperature = te
obs resolution = re
\end{lstlisting}
Claude first tests a minimal binding example, identifies the required declarations, and refits its spectral model on the same 79 observations. Its training error falls from 0.83490 for the initial fit without these declarations to 0.01368 after adding them. Kimi instead concludes that the harness evaluates a ``fixed zero context,'' pins its coefficients, and finishes with the claim ``The model is confirmed.'' DeepSeek similarly attributes the uninformative fits to the fitting procedure and substitutes a hand-derived expression. The distinguishing behavior is whether contradictory diagnostics lead to a corrected executable.

DeepSeek's expression also illustrates a mismatch between mathematical intent and execution. Its prose describes a Gaussian factor, but the submitted expression uses \texttt{exp(-pow(z,2))}. The disclosed signed-power convention makes this factor $e^{4}\approx54.60$ at $z=-2$, whereas a Gaussian square gives $e^{-4}\approx0.0183$. Explicit multiplication, as used by GPT and Claude, preserves the intended square. This semantic discrepancy is identifiable directly from the submitted expression and the public operator definition.

\subsection{Network: identical recovered topology, different governing equations}
Mapping states to observed channels verifies that GPT's and DeepSeek's final programs both contain exactly the generating graph's sixteen directed edges. Their fourth-node equations illustrate the remaining difference:
\begin{align}
\text{DeepSeek:}\quad \dot x_4 &= -a x_4 + wu(x_5+x_8-2x_4)+c_4,\\
\text{GPT:}\quad \dot x_4 &= a x_4-bx_4^3
 +gu\bigl[\tanh(x_5-x_4)+k\tanh(x_8-x_4)\bigr]+h.
\end{align}
DeepSeek tests one saturating-coupling variant but retains an affine final model. GPT combines saturation with nonlinear self-dynamics, including different response forms across nodes. Both identify the relevant neighbors; their scientific hypotheses differ in the responses assigned to those interactions. The final errors, 0.9655 and 0.0942, make that distinction consequential under held-out conditions.

\section{Submitted structures in the case studies}
\label{app:qualitative}
The following programs are the actual submitted structures. Their parameters are estimated by the shared fitting harness; the main text reports the relevant fitted coefficients and observed errors.

\subsection{DeepSeek: ecology}
The first submitted program is:
\begin{lstlisting}[basicstyle=\ttfamily\scriptsize,columns=fullflexible,keepspaces=true,breaklines=true,showstringspaces=false,frame=single]
class ODE
state a, b, c1, c2
latent c1, c2
input u
param p1, q1, r1, s1, p2, q2, r2, s2, kk, g2
d a/dt = -(p1 + q1/u)*a - r1*a*a - s1*a*a*a
d b/dt = -(p2 + q2/u)*b - r2*b*b - s2*b*b*b
d c1/dt = kk*(b - c1) + g2*(b - c1)*step(t, 0.67, 0, 1)
d c2/dt = kk*(c1 - c2) + g2*(c1 - c2)*step(t, 0.67, 0, 1)
obs y1 = a
obs y2 = b
obs y3 = c2
\end{lstlisting}
The fourth and final program is:
\begin{lstlisting}[basicstyle=\ttfamily\scriptsize,columns=fullflexible,keepspaces=true,breaklines=true,showstringspaces=false,frame=single]
class ODE
state a, b, c1, c2
latent c1, c2
input u
param p1, q1, v1, r1, p2, q2, v2, r2, kk, g2, w1, w2, w4
d a/dt = -(p1 + q1/u + v1*(a/u))*a - r1*a*a
d b/dt = -(p2 + q2/u + v2*(b/u))*b - r2*b*b
d c1/dt = kk*(b - c1) + g2*(b - c1)*step(t, 0.67, 0, 1)
d c2/dt = kk*(c1 - c2) + g2*(c1 - c2)*step(t, 0.67, 0, 1)
obs y1 = a
obs y2 = b
obs y3 = c2 + (w1 + w2*step(t, 0.67, 0, 1))*(b - c2) + w4*(b - c1)
\end{lstlisting}
The earlier second-population equation contains a negative cubic term. In the final model, its quadratic coefficient is $0.189415/u-0.190566$, yielding the unstable low-input dynamics analyzed in the main text. The final observation map also introduces a gated blend of measured and latent states.

\subsection{Grok: mechanics}
The second submission is:
\begin{lstlisting}[basicstyle=\ttfamily\scriptsize,columns=fullflexible,keepspaces=true,breaklines=true,showstringspaces=false,frame=single]
class ODE
state x, u, x2
latent x, u
param w2, g, k, k3, p, q, c
d x/dt = u
d u/dt = -w2 * x - g * u + p * x2 * x2
d x2/dt = -k * x2 - k3 * x2 * x2 * x2 + q * x
obs y1 = x
obs y4 = x
obs y2 = x2
obs y3 = c
\end{lstlisting}
The third and final submission is:
\begin{lstlisting}[basicstyle=\ttfamily\scriptsize,columns=fullflexible,keepspaces=true,breaklines=true,showstringspaces=false,frame=single]
class ODE
state x, u, x2
latent x, u
param w2, g, k, k3, p, ks, q, c
d x/dt = u
d u/dt = -w2 * x - g * u + p * sat(x2 * x2, ks, 1)
d x2/dt = -k * x2 - k3 * x2 * x2 * x2 + q * x
obs y1 = x
obs y4 = x
obs y2 = x2
obs y3 = c
\end{lstlisting}
The respective OOD, intervention, and long-horizon errors are $(0.3935,0.2276,0.1780)$ and $(0.5118,0.2555,0.1638)$. The revised saturation improves long-horizon prediction but worsens both other groups.

\subsection{Muse: stochastic dynamics}
Both submissions use the same structure:
\begin{lstlisting}[basicstyle=\ttfamily\scriptsize,columns=fullflexible,keepspaces=true,breaklines=true,showstringspaces=false,frame=single]
class SDE
state x1, x2
input u
param a, p, q, k1, s0, s1, c0, c1, k, au1, bu2
d x1/dt = -a*x1 + au1*u
d x2/dt = p*pow(x1, 0)*sat(x1, k1, 2) + q*x2 + bu2*pow(u, 0)*u
diffusion x1 = s0 + s1*x1
diffusion x2 = c0 + c1*pow(x2, 0)*sat(x2, k, 1)
obs y1 = x1
obs y2 = x2
\end{lstlisting}
The language defines signed power as $\operatorname{pow}(x,r)=\operatorname{sign}(x)|x|^r$; hence $\operatorname{pow}(x,0)$ represents the sign function. Muse submits this program after 31 experiments and again, refitted, after 44. The first submission solves the world with a maximum excess CRPS of 0.023; the refitted final one fails at 0.071, above the 0.04 threshold, with the structure unchanged.

\subsection{GPT: twelve-node network}
The acquisition sequence contains twelve single-node clamps, four long-regime probes, two input changes, three initial-condition experiments, and one ordinary observation. The final program is:
\begin{lstlisting}[basicstyle=\ttfamily\scriptsize,columns=fullflexible,keepspaces=true,breaklines=true,showstringspaces=false,frame=single]
class ODE
state x1, x2, x3, x4, x5, x6, x7, x8, x9, x10, x11, x12
input u
param a, b, c, d, g, h, k, a1, b1, q
d x1/dt = a1*x1-b1*pow(x1,2)+g*u*(k*tanh(x7-x1)+q*tanh(x9-x1))-h
d x2/dt = a*x2-b*pow(x2,3)+g*u*tanh(x1-x2)+h
d x3/dt = a*x3-b*pow(x3,3)+g*u*tanh(x2-x3)+h
d x4/dt = a*x4-b*pow(x4,3)+g*u*(tanh(x5-x4)+k*tanh(x8-x4))+h
d x5/dt = a*x5-b*pow(x5,3)+g*u*tanh(x12-x5)-h
d x6/dt = a*x6-b*pow(x6,3)+g*u*(k*tanh(x2-x6)+tanh(x11-x6))+h
d x7/dt = a*x7-c*pow(x7,2)+g*u*tanh(x10-x7)-h
d x8/dt = a*x8-b*pow(x8,3)+g*u*(tanh(x6-x8)+k*tanh(x10-x8))+h
d x9/dt = a*x9-b*pow(x9,3)+g*u*tanh(x4-x9)+h
d x10/dt = a*x10-c*pow(x10,2)+g*u*tanh(x3-x10)+h
d x11/dt = a*x11-d*pow(x11,1.5)+g*u*tanh(x7-x11)+h
d x12/dt = a*x12-b*pow(x12,3)+g*u*tanh(x8-x12)+h
obs y1 = x1
obs y2 = x2
obs y3 = x3
obs y4 = x4
obs y5 = x5
obs y6 = x6
obs y7 = x7
obs y8 = x8
obs y9 = x9
obs y10 = x10
obs y11 = x11
obs y12 = x12
\end{lstlisting}

\section{Ecology prompt and observation example}
\label{app:ecology-prompt}
The system message introduces the task as follows:
\begin{quote}\small
You are the participant in the benchmark described by the JSON below. Use the supplied JSON tools to interact with it. Large action responses may name an archive entry; response\_archive retrieves the exact response.
\end{quote}
It then supplies the domain narrative, channels, budget, controllable settings, acquisition actions, hypothesis grammar, scoring instructions, and resource policy. Selected fields from the ecology example are:
\begin{lstlisting}[basicstyle=\ttfamily\footnotesize,columns=fullflexible,keepspaces=true,breaklines=true,showstringspaces=false,frame=single]
{
  "domain": "ecology",
  "channels": ["y2", "y1"],
  "budget": 243.0,
  "inputs": {
    "u": {"box": [0.3, 1.7], "default": 1.5858456350818932}
  },
  "submission": {
    "classes": ["ODE", "Map", "StructuralDAG", "StaticFunction", "SDE"],
    "interface": "program text (see grammar) or Python model class",
    "max_ast_nodes": 400,
    "max_params": 16
  }
}
\end{lstlisting}
Neither the illustration nor the hidden structure of Figure~\ref{fig:physical-ecology} is part of the prompt; the number of resources and their equations are withheld. The ranking rule quoted to the agent orders an episode's own diagnostics; the paper reports only the pass outcome and the worst-group error. The user message instructs the agent to begin, use the available actions, and call \texttt{finish} when finished.

The first archived action is
\begin{lstlisting}[basicstyle=\ttfamily\footnotesize,frame=single]
act({"action":{"type":"observe"}})
\end{lstlisting}
Its response contains 25 observation rows at times from zero to eight, with channels ordered \texttt{["y2","y1"]}, and the updated remaining budget. Subsequent hypotheses are supplied through \texttt{fit} and \texttt{submit}; complete measurements remain accessible through the observation archive.

The seven experiment types of this world, with their prices in budget units and their argument ranges, are: \texttt{observe} (2), a standard run; \texttt{set\_initial} (3), starting populations in $[0.15,1.25]$; \texttt{kick} (6), an impulse of size 0.165--0.825 at a time in $[0,8]$; \texttt{clamp} (8), a held value in $[0.05,2.5]$ over a window in $[0,8]$; \texttt{set\_input} (10), a feed rate $u\in[0.3,1.7]$; \texttt{probe\_regime} (14), starting populations in $[0.05,2.5]$ and optionally three times the standard duration; and \texttt{deep\_assay} (15), one latent state at one time of the most recent run. The standard preparation is $\texttt{y1}=0.26$, $\texttt{y2}=0.67$ with $u=1.59$, and a standard run records 25 times over a horizon of 8.

GPT's episode on this world (Section~\ref{sec:case-study}) consists of 18 experiments (one standard run, five chosen preparations, two input changes, two impulses, four clamps, and four probes), seven assays, 32 fits, three logged hypothesis sets, and one submission. Its final two calls, verbatim except for rounding and elisions (\ldots), are:
\begin{lstlisting}[basicstyle=\ttfamily\scriptsize,columns=fullflexible,keepspaces=true,breaklines=true,frame=single]
LLM  act({"action": {"type": "submit", "init": [1.24, 0.36, ...], "bounds": [[0, 5], ...],
      "program": "class ODE
      state x, z, r, s
      latent r
      latent s
      input u
      param b, c, q, p, e, f, A, h, v, B, k
      d x/dt = c*sat(x,q,1)+x*(r-b)
      d z/dt = z*(s-e-f*pow(z,2))
      d r/dt = A*u-(h+v*x+p*z)*r
      d s/dt = B*u-(k+v*x+p*z)*s
      obs y1 = x
      obs y2 = z"}})
tool {"ok": true, "accepted": true, "cost": 0.0, "remaining": 1.0, "experiments_run": 18,
      "params": ["b", "c", "q", "p", "e", "f", "A", "h", "v", "B", "k", "r_0", "s_0"],
      "fit": {"theta": [1.239, 0.356, 0.947, 0.009, 0.372, 0.141, 0.080, 0.130, 0.012, 0.028,
                        0.199, 0.620, 0.196],
              "cost": 0.00063, "per_sample_rmse": [0.028, 0.027, 0.026, ..., 0.024],
              "n_starts": 6, "success": true}}
LLM  finish({})
\end{lstlisting}
The benchmark fits the eleven declared coefficients and the initial values \texttt{r\_0} and \texttt{s\_0} of the two latent resources on the 18 purchased runs and returns them with the training error and one RMSE per run. The hidden test error of this model, $E=0.091$, is computed after the episode and is never shown to the agent.

\section{Per-domain results, termination, and cost}
\label{app:domain-results}
\begin{table}[H]
\caption{Worlds solved out of four per domain, by agent.}
\centering\scriptsize\setlength{\tabcolsep}{2pt}
\input{tables/settings-solved.tex}
\end{table}
Of the 36 episodes per domain, the most often solved are reaction--diffusion in 34, delay in 31, and transport in 31. The least often solved are kinetics, reservoir, and lattice, in one episode each, and the 32-node network and metrology, in two each.
\begin{table}[H]
\caption{Budget use, termination, model selection, and API cost over the 80 worlds. ``Budget used'' is the mean share of the experiment budget spent. ``Truncated'' counts episodes that ended at the model's context or output limit. ``Final worse'' counts episodes whose final model has a higher held-out error than an earlier submission, and ``Solve lost'' those in which an earlier submission solved the world and the final one does not; earlier submissions are scored only for this analysis. Cost, in USD, is the API charge reported by OpenRouter. Requests served by an endpoint that reports only token counts are priced from those counts, at the provider's published rates for Claude and GPT and at the rates OpenRouter charged for Kimi's other requests. These estimates make up about half of the three models' totals.}
\label{tab:selection}
\centering\scriptsize\setlength{\tabcolsep}{4pt}
\input{tables/selection.tex}
\end{table}
Of the 720 episodes, 86 end at a context or output limit, and 14 of those still solve the world.
Truncation also separates the agents: Qwen ends 40 of its 80 episodes at its context or output limit and GLM 20, whereas GPT and Gemini end none this way.
On the SDE worlds, 29 of 30 submitted models predict the barrier-crossing probability within threshold but only 14 the distribution of the measured values: agents capture whether an event occurs far more often than the process's variability.
Among the 321 executable deterministic models that fail, the worst-scoring test group is out-of-distribution in 154 cases and long-horizon in 145; interventions are worst in only 22.

\input{appendix-analyses}
\end{document}

%% file: math_commands.tex
\usepackage{amsmath,amsfonts,bm}

\def\eqref#1{equation~\ref{#1}}

\def\1{\bm{1}}

\DeclareMathAlphabet{\mathsfit}{\encodingdefault}{\sfdefault}{m}{sl}
\SetMathAlphabet{\mathsfit}{bold}{\encodingdefault}{\sfdefault}{bx}{n}

%% file: figures/selection-case.tex
\begin{tikzpicture}
\begin{groupplot}[group style={group size=2 by 1, horizontal sep=1.6cm}, width=0.5\linewidth, height=2.5cm,
  tick label style={font=\scriptsize}, label style={font=\scriptsize}, title style={font=\scriptsize, yshift=-4pt},
  legend style={font=\scriptsize, draw=none, fill=none}, every axis plot/.append style={line width=0.9pt}]
\nextgroupplot[title={(a) Purchased run, feed $u=0.30$}, ylabel={population $b$}, ytick={0,0.3,0.6},
  legend to name=selectioncaselegend, legend columns=4, legend style={/tikz/every even column/.append style={column sep=10pt}}]
\addplot[only marks, mark=*, mark size=1.1pt, black] coordinates {(0.000,0.6191) (0.333,0.6094) (0.667,0.5504) (1.000,0.5290) (1.333,0.5102) (1.667,0.4649) (2.000,0.4166) (2.333,0.3938) (2.667,0.3600) (3.000,0.3231) (3.333,0.3030) (3.667,0.2549) (4.000,0.2345) (4.333,0.2315) (4.667,0.2044) (5.000,0.1464) (5.333,0.1530) (5.667,0.1363) (6.000,0.1366) (6.333,0.1209) (6.667,0.1017) (7.000,0.0612) (7.333,0.0670) (7.667,0.0707) (8.000,0.0661)};
\addlegendentry{measured}
\addlegendimage{black}
\addlegendentry{true system}
\addplot[blue, dashed] coordinates {(0.000,0.6237) (0.333,0.5755) (0.667,0.5314) (1.000,0.4909) (1.333,0.4536) (1.667,0.4193) (2.000,0.3876) (2.333,0.3584) (2.667,0.3315) (3.000,0.3066) (3.333,0.2836) (3.667,0.2623) (4.000,0.2426) (4.333,0.2244) (4.667,0.2076) (5.000,0.1920) (5.333,0.1776) (5.667,0.1643) (6.000,0.1520) (6.333,0.1405) (6.667,0.1300) (7.000,0.1202) (7.333,0.1112) (7.667,0.1028) (8.000,0.0951)};
\addlegendentry{first model}
\addplot[red] coordinates {(0.000,0.6237) (0.333,0.5915) (0.667,0.5583) (1.000,0.5243) (1.333,0.4899) (1.667,0.4555) (2.000,0.4214) (2.333,0.3879) (2.667,0.3553) (3.000,0.3239) (3.333,0.2940) (3.667,0.2657) (4.000,0.2392) (4.333,0.2145) (4.667,0.1917) (5.000,0.1707) (5.333,0.1516) (5.667,0.1343) (6.000,0.1187) (6.333,0.1046) (6.667,0.0921) (7.000,0.0809) (7.333,0.0709) (7.667,0.0621) (8.000,0.0543)};
\addlegendentry{final model}
\nextgroupplot[title={(b) Held-out test from an out-of-range start, $u=0.68$}, ymin=0, ymax=8.0, ytick={0,4,8}]
\addplot[black] coordinates {(0.000,7.2618) (0.333,5.9326) (0.667,4.9550) (1.000,4.2044) (1.333,3.6102) (1.667,3.1291) (2.000,2.7331) (2.333,2.4028) (2.667,2.1244) (3.000,1.8877) (3.333,1.6849) (3.667,1.5100) (4.000,1.3582) (4.333,1.2258) (4.667,1.1097) (5.000,1.0075) (5.333,0.9171) (5.667,0.8368) (6.000,0.7652) (6.333,0.7013) (6.667,0.6439) (7.000,0.5922) (7.333,0.5456) (7.667,0.5035) (8.000,0.4652)};
\addplot[blue, dashed] coordinates {(0.000,7.2618) (0.333,4.2765) (0.667,3.3018) (1.000,2.7666) (1.333,2.4129) (1.667,2.1553) (2.000,1.9557) (2.333,1.7944) (2.667,1.6601) (3.000,1.5457) (3.333,1.4464) (3.667,1.3590) (4.000,1.2812) (4.333,1.2112) (4.667,1.1477) (5.000,1.0897) (5.333,1.0364) (5.667,0.9871) (6.000,0.9414) (6.333,0.8988) (6.667,0.8590) (7.000,0.8217) (7.333,0.7866) (7.667,0.7535) (8.000,0.7222)};
\addplot[red] coordinates {(0.000,7.2618) (0.333,8.0000)};
\node[font=\scriptsize, red!80!black, anchor=west] at (axis cs:1.3,7.0) {$\leftarrow$ final model diverges};
\end{groupplot}
\end{tikzpicture}\\[1pt]
\ref{selectioncaselegend}

%% file: tables/results.tex
\begin{tabular}{@{}lrcrrrrrrr@{}}
\toprule
 & & & \multicolumn{4}{c}{Solved by kind of system} & & & \\
\cmidrule(lr){4-7}
Model & Solved & 95\% CI & Dyn. (36) & Spatial dyn. (12) & Static, causal (16) & Stoch. (16) & Median & Invalid & Truncated \\
\midrule
GPT 6 Astra & 48/80 & 49--70\% & 18 & 12 & 9 & 9 & 0.045 & 2 & 0 \\
Claude Fable 5.1 & 38/80 & 37--58\% & 10 & 12 & 9 & 7 & 0.106 & 3 & 1 \\
Muse Spark 1.3 & 24/80 & 21--41\% & 4 & 12 & 7 & 1 & 0.202 & 5 & 7 \\
Grok 4.6 & 23/80 & 20--39\% & 5 & 9 & 6 & 3 & 0.301 & 3 & 3 \\
Qwen 3.8 Max & 23/80 & 20--39\% & 8 & 10 & 3 & 2 & 0.276 & 9 & 40 \\
GLM 5.3 & 22/80 & 19--38\% & 7 & 8 & 6 & 1 & 0.265 & 9 & 20 \\
DeepSeek v4.1 Flash & 18/80 & 15--33\% & 4 & 9 & 0 & 5 & 0.455 & 10 & 3 \\
Gemini 3.8 Flash & 17/80 & 14--31\% & 3 & 7 & 6 & 1 & 0.348 & 7 & 0 \\
Kimi K3 & 17/80 & 14--31\% & 4 & 9 & 3 & 1 & 0.425 & 7 & 12 \\
\bottomrule
\end{tabular}

%% file: figures/outcome-profiles.tex
\begin{tikzpicture}
\begin{axis}[xbar stacked, scale only axis, width=0.6\linewidth, height=1.95cm, bar width=4pt, xmin=0, xmax=80,
  xtick={0,20,40,60,80}, xlabel={Worlds (of 80)}, xlabel style={font=\scriptsize},
  ytick=data, symbolic y coords={{Kimi K3},{Gemini 3.8 Flash},{DeepSeek v4.1 Flash},{GLM 5.3},{Qwen 3.8 Max},{Grok 4.6},{Muse Spark 1.3},{Claude Fable 5.1},{GPT 6 Astra}}, y tick label style={font=\scriptsize},
  x tick label style={font=\scriptsize}, enlarge y limits=0.07, axis x line*=bottom, axis y line*=left,
  legend style={font=\scriptsize, draw=none, at={(1.03,0.5)}, anchor=west, legend columns=1,
  /tikz/every even column/.append style={column sep=6pt}}, legend cell align=left]
\addplot[fill=black!70, draw=white, line width=0.3pt] coordinates {(17,{Kimi K3}) (17,{Gemini 3.8 Flash}) (18,{DeepSeek v4.1 Flash}) (22,{GLM 5.3}) (23,{Qwen 3.8 Max}) (23,{Grok 4.6}) (24,{Muse Spark 1.3}) (38,{Claude Fable 5.1}) (48,{GPT 6 Astra})};
\addlegendentry{Solved}
\addplot[fill=black!18, draw=white, line width=0.3pt] coordinates {(54,{Kimi K3}) (52,{Gemini 3.8 Flash}) (50,{DeepSeek v4.1 Flash}) (42,{GLM 5.3}) (45,{Qwen 3.8 Max}) (50,{Grok 4.6}) (50,{Muse Spark 1.3}) (39,{Claude Fable 5.1}) (30,{GPT 6 Astra})};
\addlegendentry{Finite error}
\addplot[fill=orange!80, draw=white, line width=0.3pt] coordinates {(4,{Kimi K3}) (7,{Gemini 3.8 Flash}) (4,{DeepSeek v4.1 Flash}) (3,{GLM 5.3}) (3,{Qwen 3.8 Max}) (5,{Grok 4.6}) (2,{Muse Spark 1.3}) (2,{Claude Fable 5.1}) (2,{GPT 6 Astra})};
\addlegendentry{Diverges}
\addplot[fill=red!70!black, draw=white, line width=0.3pt] coordinates {(4,{Kimi K3}) (3,{Gemini 3.8 Flash}) (6,{DeepSeek v4.1 Flash}) (2,{GLM 5.3}) (2,{Qwen 3.8 Max}) (0,{Grok 4.6}) (0,{Muse Spark 1.3}) (1,{Claude Fable 5.1}) (0,{GPT 6 Astra})};
\addlegendentry{Missing channel}
\addplot[fill=blue!55, draw=white, line width=0.3pt] coordinates {(1,{Kimi K3}) (1,{Gemini 3.8 Flash}) (2,{DeepSeek v4.1 Flash}) (11,{GLM 5.3}) (7,{Qwen 3.8 Max}) (2,{Grok 4.6}) (4,{Muse Spark 1.3}) (0,{Claude Fable 5.1}) (0,{GPT 6 Astra})};
\addlegendentry{No submission}
\end{axis}
\end{tikzpicture}

%% file: appendix-protocol.tex
\section{Scenario specifications}
\label{app:scenarios}

This appendix specifies the mechanics, kinetics, discrete, causal, scaling, ecology, network, Network-S1, metrology, spectroscopy, and stochastic-dynamics domains, the domains marked $^\dagger$ in Table~\ref{tab:domains}; Appendix~\ref{app:extensions} specifies the others. The specifications below describe their generative distributions and experimental interfaces. Each evaluated agent receives a domain narrative, channel names, standard preparation, controllable ranges, action prices, acquisition budget, and the common hypothesis language. The sampled equations, parameters, latent variables, sensor assignments, and hidden test conditions remain withheld during discovery. Table~\ref{tab:scenario-panel} gives the public numerical configuration of the 44 evaluated worlds; world numbering agrees with the per-world result tables.

\begin{table}[H]
\caption{Public configuration of the 44 worlds in these domains. Channels include any auxiliary signals. $T$ is the standard simulated duration, measured in ticks for discrete maps; extended runs use longer grids. A stochastic run returns an ensemble with the indicated default number of samples per path. Budgets are in budget units, distinct from API expenditure.}
\label{tab:scenario-panel}
\centering\small\setlength{\tabcolsep}{3pt}
\input{tables/scenario-panel.tex}
\end{table}

\subsection{Common construction and observation model}

The mechanics, kinetics, discrete, causal, and scaling generators construct equations as sums of products of elementary responses. Before the parameter allowance is imposed, an equation draws one to four interaction terms with probabilities $(0.25,0.45,0.25,0.05)$; each term contains one or two factors with probabilities $(0.65,0.35)$. The dynamical generators reserve parameters for characteristic scales, hidden initial conditions, and optional switching behavior, then restrict the term count to fit the resulting allowance. Coefficient magnitudes are sampled on a logarithmic scale, with signs selected independently for interaction terms. Characteristic scales are shared across terms involving the same coordinate.

The continuous-dynamics atom pool comprises signed powers, saturation, exponentials, logarithmic responses, hyperbolic tangents, oscillatory forcing, and compositions. Static equations omit explicit time dependence; the map profile emphasizes saturating and other nonlinear update responses. Power exponents are drawn from
\begin{equation}
\left\{-1,-\tfrac12,\tfrac13,\tfrac12,\tfrac23,1,\tfrac32,2,3\right\},
\end{equation}
and saturation exponents from $\{\tfrac12,1,2,3,4\}$. In the mechanics, kinetics, and discrete domains, an equation additionally receives a negative self-response with probability 0.85. Composed atoms apply an exponential, saturation, logarithm, or hyperbolic tangent to a powered coordinate. The exact meanings of these operators are given in Appendix~\ref{app:dsl}.

Measurements combine a generating observation function with independent Gaussian sensor noise. For the mechanics, kinetics, discrete, causal, and scaling domains, the signal-to-noise ratio is sampled logarithmically between 30 and 1,000; channel noise scales are determined from reference observations with a scale floor. Zero, one, or two auxiliary channels are added with probabilities $(0.40,0.35,0.25)$. They can be noise-dominated signals, transformed combinations of genuine channels, or lagged channels in dynamical domains. Channel order is shuffled, and the agent receives no label identifying auxiliary signals. Behavioral scoring uses the genuine observation channels.

Each acquisition specifies a fresh experiment: an initial-condition change or input setting applies to that run, rather than permanently altering subsequent runs. A standard observation returns to the disclosed standard preparation. The most recent trajectory is retained for an optional internal assay. Acquired measurements accumulate for fitting throughout the episode.

Deterministic ODEs use fixed-step, classical fourth-order Runge--Kutta integration. Maps advance synchronously by one update per tick. Stochastic systems use the separate Euler--Maruyama configuration specified below.

For mechanics, kinetics, discrete dynamics, ecology, and both network domains, the eight OOD tests draw preparations beyond the ordinary box. A signed coordinate uses its box midpoint plus or minus three to six half-widths; a positive-preparation coordinate uses three to six times its ordinary upper bound. Half of these conditions use extended input values when an input exists. Intervention tests combine clamps, impulses, and input changes, including unavailable intervention types or values outside the ordinary action bands. Public long probes reach three times the standard duration, while hidden long tests reach ten times that duration with proportionally longer observation grids. Causal and scaling OOD root settings use the same shell construction. Metrology and spectroscopy instead draw within their extended boxes while placing at least one coordinate outside its ordinary range.

\subsection{Mechanics}

The scenario is an instrumented mechanical assembly with unknown effective coordinates and couplings. Its generating model is an ODE with two to four states. Up to two states are initially designated unobserved, subject to retaining at least two observed coordinates. A directed graph defines candidate interactions: edge probability is selected from $\{0.35,0.50,0.65\}$, and proposals require incoming connections and a path from each latent coordinate to an observed one. The equations select and compose responses from these permitted neighbors. A constant external drive $u$ is included with probability 0.5 and acts on one or two states. Coefficient magnitudes lie in $[0.2,2.5]$, and explicit oscillatory forcing can depend on time.

Coordinate scales lie in $[0.5,2]$. Ordinary preparations span $[-s,s]$ and wider preparations $[-2s,2s]$. When present, the ordinary drive range is $[0.5,1.5]$. Observations can mix coordinates through a signed-square or multiplicative term, and a channel can have a saturating readout. Such transformations make the underlying state distinct from the measured signal. An optional thresholded response, proposed with probability 0.4, becomes active beyond the ordinary regime.

The standard horizon is chosen from the characteristic rate at the default preparation and clipped to $[0.5,60]$. Each standard trajectory contains 41 recorded times, with four Runge--Kutta substeps between adjacent observations. Actions include ordinary runs and chosen preparations, with impulses, clamps, drive changes, and wider/longer probes offered according to the world's disclosed action list. An internal assay is always listed. Hidden tests comprise eight OOD conditions, six interventions or altered inputs, and three runs at ten times the standard horizon. This domain tests coupled dynamics, hidden state, nonlinear readouts, and mechanisms that emerge outside short ordinary runs.

\subsection{Kinetics}

The scenario is a reaction vessel whose assayed signals represent deviations from a reference state. It uses the same two-to-four-state ODE construction and observation transformations as mechanics, with coefficient magnitudes in $[0.05,5]$ and no explicitly time-dependent forcing atoms. Ordinary initial preparations lie in $[0.1s,s]$, and extended preparations in $[0.1s,2s]$; the state variables themselves represent deviations and can evolve through signed values. An optional input represents a feed setting.

Experiments correspond to preparing species, injecting a bolus, holding a coordinate with a feed, changing the external setting, or assaying an unobserved species. The horizon, numerical integration, optional action availability, and $8/6/3$ hidden-test counts follow the continuous mechanics profile. The structural challenge is to explain jointly the observed reaction-like dynamics and the influence of unmeasured quantities, including their response to interventions and long-term depletion.

\subsection{Discrete dynamics}

The scenario is a synchronous update process observed through noisy counters. Two to four states follow
\begin{equation}
x_{k+1}=x_k+F(x_k,u;\theta),
\end{equation}
with independently composed nonlinear increments, optional unobserved states, transformed measurements, and an optional external rate. The update-response pool contains saturation, exponentials, logarithmic responses, hyperbolic tangents, and compositions; an additional thresholded response can introduce a regime change. Ordinary and wider preparation ranges follow the signed mechanics profile.

A standard experiment records 31 values, including the initial state and thirty updates. Setting a counter changes its initial value, an impulse adds to a counter during the run, and a clamp pins a coordinate over an interval. A long public probe extends the run to ninety updates; hidden long-horizon tests use 300 updates. Eight OOD, six intervention, and three long-horizon tests evaluate whether the inferred update rules capture the process beyond the observed sequences.

\subsection{Causal systems}

The scenario is a population described by jointly measured attributes. A random acyclic graph contains three to six nodes, with candidate forward-edge probability selected from $\{0.40,0.55,0.70\}$. Non-root nodes receive nonlinear structural equations composed from their parents. With probability 0.5, one eligible non-root, non-sink node is hidden. Root values are sampled uniformly from their ordinary boxes; non-root equations receive independent additive Gaussian structural noise with standard deviation $0.1$ times the node scale, followed by sensor noise.

An observational action buys $n$ independent joint samples. A clamp replaces one visible node's structural equation by a chosen value, then buys $n$ samples of the intervened system. Ordinary observation costs one unit per sample, while intervention costs are disclosed per world. Hidden evaluation uses ten extended root configurations and six interventions on visible non-root nodes, comparing the noiseless structural responses. This domain requires reasoning from noisy observations and interventions to the dependencies and functional mechanisms of the system.

\subsection{Scaling relations}

The scenario is a measurement rig with adjustable inputs and dependent recorded quantities. It uses a three-to-six-node acyclic computation graph with the same compositional static-equation construction as the causal domain. An intermediate node can be unobserved. Here the generating relation is deterministic before sensor noise: each query specifies all root-input values, rather than sampling a population with structural noise.

Ordinary input ranges are $[0.1s,s]$ and extended ranges $[0.1s,2s]$. An ordinary point evaluation costs one unit; wider-regime evaluations, when offered, cost more. The hidden battery contains ten OOD root configurations and six interventions on observed derived quantities. A candidate therefore specifies a complete system of relationships, including how changes to an intermediate quantity propagate, rather than only a separate regression for each output.

\subsection{Ecology}

The scenario is a controlled microcosm with two measured populations and two hidden resource stores. The generator independently selects which resources support growth, which consumers deplete each resource, and which variables contribute feedback. Population equations combine resource-dependent growth, linear loss, and nonlinear density regulation. Resource equations combine input-driven regeneration, turnover, consumption, transfer between stores, and signed feedback. Response terms are composed from linear, saturating, hyperbolic-tangent, logarithmic, and signed-power atoms. Coefficients are shared by process role while source variables and atom choices vary across terms.

Population preparations lie in $[0.15,1.25]$, extended preparations in $[0.05,2.5]$, and replenishment $u$ in $[0.3,1.7]$. Hidden resource initial values lie in $[0.45,1.15]$. Standard trajectories span eight time units with 25 observations and two integration substeps per observation interval. All of the dynamical acquisition types are offered, including assays of the resource stores. Zero or one auxiliary signal is added; the sensor SNR lies in $[40,800]$.

The hidden battery uses eight extended preparations or input conditions, six interventions, and three eighty-unit trajectories. Input extrapolation includes replenishment values in $[1.9,2.8]$. The scientific challenge is to distinguish direct population interactions from effects mediated by resource depletion, regeneration, and memory in hidden stores.

\subsection{Network dynamics}

The standard network contains five directly measured nodes. A directed graph is sampled edge by edge, with edge probability in $\{0.25,0.35,0.45\}$ and acceptance conditional on strong connectivity. Each node combines linear self-activation, a nonlinear restoring response, a bias, and input-scaled interactions with its incoming neighbors. Coupling functions act on differences $x_j-x_i$ and are selected from linear, hyperbolic-tangent, and signed-power responses. Restoring exponents are drawn from $\{1.5,2,3\}$.

Ordinary preparations lie in $[-0.35,0.35]$, probes in $[-1,1]$, and the global coupling input in $[0.4,1.4]$. Standard trajectories span six time units and contain 25 observations, with two integration substeps per observation interval. Every dynamical acquisition type is offered. The five node signals can be accompanied by one auxiliary channel; measurement SNR lies in $[40,800]$. The $8/6/3$ hidden battery includes input values in $[1.8,2.6]$ and sixty-unit trajectories. The agent must identify both the directed graph and the nonlinear response associated with its nodes and edges.

\subsection{Network-S1: twelve-node scaling setting}

Network-S1 increases the system to twelve directly measured nodes. A randomly ordered directed cycle provides a strongly connected backbone. Additional directed edges are sampled independently with probability in $\{0.025,0.04,0.06\}$, retaining graphs with two to five additional edges. Backbone interactions use hyperbolic tangents; the additional interactions select from the network response atoms. Distinct shared coefficients control backbone and additional edges, and node-specific bias signs and restoring exponents create heterogeneous dynamics.

Ordinary preparations lie in $[-0.2,0.2]$, probes in $[-0.7,0.7]$, and the coupling input in $[0.5,1.3]$. A standard experiment spans four time units with 25 observations and two integration substeps per observation interval. The hidden $8/6/3$ battery includes input values in $[1.5,2.1]$ and forty-unit trajectories. This is a larger network domain under the same interaction and submission interface, with graph recovery and functional identification both contributing to performance.

\subsection{Metrology}

The scenario is a calibration rig with two controlled settings, \texttt{u1} and \texttt{u2}, and two anonymous instruments, \texttt{sensor\_a} and \texttt{sensor\_b}. A latent reference response is constructed from an intercept and six to eight terms. The terms include univariate powers and nonlinear responses, products of responses of the two inputs, and nonlinear functions of mixed input combinations. An outer transformation selects the direct response, a hyperbolic tangent, a linear-plus-tangent response, or a sine-modulated response.

The instruments apply different unknown transformations to the same latent response: one has saturating compression and the other a soft dead zone, with their assignment randomized. Both include an unknown gain, bias, scale, and shape exponent. Ordinary settings are in $[-1,1]^2$ and extended settings in $[-2.5,2.5]^2$. Channel order is shuffled. Input values are returned exactly, while instrument readings have Gaussian noise with SNR in $[100,700]$ and standard-deviation floor 0.002.

An ordinary two-instrument measurement costs one unit and an extended measurement costs four. A \texttt{gold\_measure} action costs eight and returns the noiseless latent reference at a chosen extended-range setting. This scalar can guide model construction and parameter initialization; it is separate from the sensor dataset used by the standard fitter. Hidden evaluation comprises sixteen outer-range settings and eight controlled-input changes, scoring the observable response channels. The task combines discovery of a latent relationship with separation of instrument distortion from the underlying signal.

\subsection{Spectroscopy}

The scenario measures intensity as a function of frequency $\nu$, field $F$, temperature $T$, and requested resolution $r$. A spectrum combines three or four generated peak components and a background. Each component independently selects its field-dependent center shift, resolution-dependent width, temperature broadening, thermal amplitude, and line-shape kernel. Center shifts can be linear, polynomial, sinusoidal, or saturating. Width laws combine an intrinsic scale with additive, quadratic, root-sum-square, multiplicative, or saturating resolution responses. Thermal amplitudes use activated or contrast responses. Kernels include Gaussian, super-Gaussian, Lorentzian, squared-Lorentzian, mixed Gaussian--Lorentzian, doublet, and triplet shapes. The background can depend on temperature, field, or resolution.

\begin{table}[H]
\caption{Spectroscopy acquisition ranges.}
\centering\small
\begin{tabular}{@{}lll@{}}
\toprule
Setting & Ordinary & Extended \\
\midrule
Frequency & $[0.5,9]$ & $[0.2,12]$ \\
Field & $[-0.3,0.3]$ & $[-0.8,0.8]$ \\
Temperature & $[0.8,1.3]$ & $[0.5,2]$ \\
Resolution & $[0.03,0.12]$ & $[0.006,0.18]$ \\
\bottomrule
\end{tabular}
\end{table}

Each query returns the four settings and one noisy intensity. The disclosed acquisition law is
\begin{equation}
c(r)=0.25+\frac{0.02}{r},\qquad
\sigma(r)=0.003\sqrt{\frac{0.04}{r}}.
\end{equation}
An ordinary query costs $c(r)$ and an extended query costs $2c(r)$. Input channels are noiseless. The hidden battery contains 24 outer-range conditions and eight controlled changes of one input. Recovering the response requires identifying how spectral components shift, broaden, and change amplitude across conditions, while allocating measurements across resolution and cost.

\subsection{Stochastic dynamics}

The scenario is a fully observed one- or two-state It\^o diffusion,
\begin{equation}
dX_i=f_i(X,u;\theta)\,dt+g_i(X_i;\theta)\,dW_i,
\end{equation}
with independent Wiener drivers. One and two states are proposed with probabilities 0.35 and 0.65. Each drift combines linear decay with two or three independently composed response terms, including a driven term. Responses include linear, hyperbolic-tangent, sinusoidal, signed-logarithmic, signed-saturation, and sublinear-power atoms, and products of two atoms. A diffusion amplitude combines a positive baseline with a bounded state-dependent modulation. The generating amplitudes remain positive; the submitted language represents amplitudes directly.

Ordinary initial states and input $u$ lie in $[-1,1]$. Extended initial states lie in $[-2.5,2.5]$ and extended inputs in $[-2,2]$. Ordinary durations range from 0.25 to 1.5, and probe durations from 0.25 to 3, in increments of 0.05. The public interface specifies one to 256 independent paths per acquisition. A probe must extend at least one initial condition, input, or duration beyond its ordinary range. Cost is the path count times duration. Integration uses Euler--Maruyama with step 0.01; observations are spaced by 0.05 and include independent sensor noise with standard deviation between 0.01 and 0.02.

Hidden evaluation uses four conditions with 512 paths each. Two OOD conditions start at signed magnitudes between 2.7 and 3.6 and run for 2.5 time units. Two first-passage conditions start near a barrier and run for three time units. Each condition selects a target state channel. Distributional comparison uses normalized energy/CRPS excess at recorded-time checkpoints, and first-passage comparison uses excess Brier loss for a barrier crossing. The task requires a model of both the systematic drift and stochastic variability, rather than a fit to a particular noise realization.

\section{Agent tools and experimental actions}
\label{app:tools}

\subsection{Conversation-level tools}

The evaluated interface exposes exactly three tools. The full public experiment description is supplied in the initial prompt, and the action list specifies which experiments are available in that world.

\begin{description}
\item[\texttt{act(\{"action": A\})}.] Execute the action object $A$. Its \texttt{type} selects an acquisition, a fit, a submission, or hypothesis logging. The outer object contains exactly \texttt{action}; arguments such as \texttt{ic} and \texttt{program} are fields of $A$ itself.
\item[\texttt{response\_archive(\{\ldots\})}.] Read previously returned action responses, including complete observations and fitting diagnostics. This read-only tool also provides numerical extraction and summary statistics.
\item[\texttt{finish(\{\})}.] End the episode. The final model is the last accepted \texttt{submit} action. The termination call itself supplies no model or fitted parameters.
\end{description}

Calls and responses use JSON. Experiment prices are deducted from the remaining budget; fitting, submission, archive access, and hypothesis logging are free. A rejected request returns an error without buying measurements or replacing the committed model. Successful action responses report \texttt{ok}, \texttt{cost}, \texttt{remaining}, and \texttt{experiments\_run}, together with the action-specific data.

\subsection{Acquisition actions}

Table~\ref{tab:action-prices} gives the prices in these domains. A requested action must appear in the world's disclosed action list. All numerical settings must lie within the corresponding disclosed bounds, and the complete acquisition cost must fit within the remaining budget.

\begin{table}[H]
\caption{Base acquisition prices across the four worlds of each domain. Ranges summarize world-specific prices; an asterisk means an action is available in only a subset of those four worlds. Causal prices are per joint sample and are multiplied by the requested sample count. For spectroscopy, $c(r)=0.25+0.02/r$; stochastic prices are $PH$ for $P$ paths of duration $H$. The three free action types are described separately.}
\label{tab:action-prices}
\centering\small\setlength{\tabcolsep}{2.6pt}
\input{tables/scenario-action-prices.tex}
\end{table}

\paragraph{\texttt{observe}.}
For ODE and map worlds, \texttt{\{"type":"observe"\}} records a run from the standard preparation. For causal worlds, \texttt{\{"type":"observe","n":8\}} buys eight independent joint samples; \texttt{n} defaults to one and may be any positive integer up to 4,096 that fits the budget. For static functions, \texttt{ic} must provide every controlled input, for example \texttt{\{"type":"observe","ic":\{"u1":0.2,"u2":-0.4\}\}}. For SDE worlds, \texttt{paths} is required, with optional \texttt{ic}, \texttt{inputs}, and \texttt{horizon}; omitted settings use the standard preparation and default duration.

\paragraph{\texttt{set\_initial}.}
The action \texttt{\{"type":"set\_initial","ic":\{"y1":0.4\}\}} changes the initial value of a directly controllable channel and records a fresh dynamical run. A dictionary can specify several channels; unspecified coordinates retain their standard preparations. Hidden initial states are part of the world's preparation and remain outside these controls.

\paragraph{\texttt{set\_input}.}
The action \texttt{\{"type":"set\_input","inputs":\{"u":1.0\}\}} sets an exogenous input for one run. The input is constant during that experiment. An optional \texttt{ic} changes the preparation as well.

\paragraph{\texttt{kick}.}
The action \texttt{\{"type":"kick","channel":"y1","time":1.0,"delta":0.5\}} adds an impulse to a controllable coordinate during a fresh run. The absolute impulse magnitude must lie in the disclosed band, and its time must lie within the standard horizon. An optional \texttt{ic} changes the initial preparation.

\paragraph{\texttt{clamp}.}
For a dynamical world, the action specifies \texttt{channel}, \texttt{value}, \texttt{t0}, and \texttt{t1}. The coordinate is held fixed over the interval, with $0\leq t_0<t_1\leq T$; an optional \texttt{ic} changes the preparation. For a causal world, \texttt{channel} and \texttt{value} define a do-intervention, and optional \texttt{n} specifies the number of independent joint samples. The intervened structural equation is replaced by the requested value.

Combining a custom \texttt{ic} with \texttt{kick}, dynamical \texttt{clamp}, or \texttt{set\_input} adds the preparation surcharge $c_{\mathrm{initial}}-c_{\mathrm{observe}}$ to the listed action price.

\paragraph{\texttt{probe\_regime}.}
For ODE and map worlds, \texttt{ic} selects a preparation in the wider box and \texttt{long:true} triples the horizon and number of observation intervals. For static functions, all controlled inputs must be supplied in \texttt{ic}, within the extended box. Spectroscopy applies its resolution-dependent price. For stochastic worlds, a probe uses \texttt{ic}, \texttt{inputs}, \texttt{horizon}, and \texttt{paths}, with at least one setting outside the ordinary acquisition range.

\paragraph{\texttt{deep\_assay}.}
The action \texttt{\{"type":"deep\_assay","k":1,"t":0.0\}} buys a noiseless value of the $k$th hidden state from the most recent dynamical trajectory. Indices are one-based, and the requested time must lie within that trajectory. The returned value uses the nearest recorded time, which is reported with it. An index beyond the hidden-state list returns \texttt{null} and still incurs the assay price. The scalar response is available for reasoning and parameter initialization; it is separate from the standard observation dataset used by fitting.

\paragraph{\texttt{gold\_measure}.}
Available in metrology, this action supplies both settings in \texttt{ic} and returns the noiseless latent reference as \texttt{reference}, together with the queried \texttt{setting}. It costs eight acquisition units. Like an internal assay, it provides additional scientific evidence without appending an ordinary sensor row to the fitting dataset.

\subsection{Observation formats}

ODE and map acquisitions return \texttt{experiment}, a time vector \texttt{t}, ordered \texttt{channels}, and a matrix \texttt{y} indexed by time and channel. Static-function evaluations use the same fields with a single observation row. Causal acquisitions return \texttt{channels} and \texttt{samples}, a matrix indexed by independently sampled unit and channel. Successful stochastic ensembles return \texttt{experiment}, \texttt{t}, \texttt{channels}, \texttt{paths}, and \texttt{y} indexed by path, time, and channel. The reported channel order specifies the interpretation of every array column.

An ordinary deterministic trajectory contains its initial observation and one observation after each of the disclosed intervals. Impulses and clamps are included in the returned experimental specification, so later fitting uses the conditions under which each trajectory was acquired.

\subsection{Fitting, submission, and hypothesis logging}

\paragraph{\texttt{fit}.}
The action \texttt{\{"type":"fit","program":"<DSL text>"\}} estimates parameters using all purchased fitting data. Optional \texttt{init} supplies a parameter vector and optional \texttt{bounds} supplies one lower/upper pair per parameter; the order follows the program's parameter list, including any added hidden-state initial values. The response provides the ordered \texttt{params} and a \texttt{fit} object containing \texttt{theta}, \texttt{cost}, \texttt{success}, \texttt{n\_starts}, \texttt{per\_sample\_rmse}, and elapsed fitting time.

For deterministic classes, the fitter performs multistart nonlinear least squares on simulated observations. Channel scales come from purchased data, residuals are clipped to $[-10,10]$, and the reported cost is their mean square. Static root inputs and imposed do-values are treated as experimental conditions rather than regression targets. The standard configuration uses six starts with up to sixty least-squares evaluations per start; a supplied initialization occupies one of those starts. A parameter-free program is evaluated directly.

For stochastic programs, fitting matches empirical drift and innovation-variance moments from the acquired path increments, correcting the variance for measurement noise. It uses six starts without an initialization and one supplied start otherwise, again with up to sixty evaluations per start. SDE fitting requires exactly one directly observed channel per state, in the disclosed channel order. Identical program/data/initialization/bounds requests use the same optimizer seed and therefore reproduce the same diagnostic.

The training error uses the channels the candidate predicts. Consequently, a low training error alone does not establish that the candidate predicts every channel required by the held-out evaluation. Observation coverage, faithful execution, and generalization are separate checks.

\paragraph{\texttt{submit}.}
The action accepts the same \texttt{program}, \texttt{init}, and \texttt{bounds} fields, performs the fitting step, and records the resulting program and parameters as the current commitment. Its public response gives acceptance and fitting diagnostics. Hidden errors are withheld. A subsequent accepted submission replaces the current commitment; a rejected submission leaves it unchanged. Calling \texttt{fit} alone does not commit a model.

\paragraph{\texttt{log\_hypotheses}.}
This action takes a \texttt{hypotheses} field containing a list of program strings and records those proposed explanations at the current acquisition spend. It neither fits nor submits the programs and provides no hidden-score feedback.

\subsection{Complete observation access}

Responses exceeding 20,000 characters are represented in the conversation by a bounded message identifying their archive entry. The archive retains the complete response. Its four operations are:
\begin{description}
\item[\texttt{list}.] Accepts optional \texttt{offset} and \texttt{limit}; returns entry indices, action types, and response lengths. Pages contain at most 64 entries.
\item[\texttt{get}.] Requires \texttt{index} and accepts \texttt{offset} and \texttt{limit}; returns an exact JSON-text \texttt{chunk}, its position, total length, and \texttt{next\_offset}. The evaluated configuration uses pages of at most 8,000 characters.
\item[\texttt{values}.] Requires \texttt{index} and exactly one of \texttt{field} or \texttt{path}. A path is a sequence of object keys and array indices. The response reports the array shape and pages its numerical entries in row-major order, with at most 512 values per page.
\item[\texttt{stats}.] Selects a field or path in the same way and accepts an optional \texttt{axis}. It returns count, minimum, maximum, mean, and population standard deviation, either globally or along the selected axis. Result arrays are paged using \texttt{offset} and \texttt{limit}.
\end{description}

For example, the following calls inspect an archived trajectory without buying another experiment:
\begin{lstlisting}[basicstyle=\ttfamily\footnotesize,breaklines=true,frame=single]
response_archive({"operation":"list","offset":0,"limit":8})
response_archive({"operation":"get","index":0,"offset":0,"limit":8000})
response_archive({"operation":"values","index":0,"field":"y","limit":64})
response_archive({"operation":"stats","index":0,"field":"y","axis":0})
\end{lstlisting}
For a causal response the numerical field is \texttt{samples}. For an ensemble, reducing \texttt{y} along axis zero summarizes across paths. Returned \texttt{next\_offset} values permit complete traversal of a paged response.

\section{Executable hypothesis language}
\label{app:dsl}

\subsection{Program structure}

A hypothesis is a UTF-8 program with one statement per line. Blank lines and comments beginning with \texttt{\#} are ignored. State and parameter names are chosen by the agent; observation-channel and exogenous-input names must match the experiment's description. The principal statements are:

\begin{lstlisting}[basicstyle=\ttfamily\footnotesize,breaklines=true,frame=single]
class ODE | Map | SDE | StructuralDAG | StaticFunction
state x1, x2, ...
latent z1, z2, ...
input u, ...
param a, b, k, ...
d x1/dt = <expression>       # ODE or SDE drift
next x1 = <expression>       # complete next value for a Map
diffusion x1 = <expression>  # SDE diffusion amplitude
let x2 = <expression>       # static structural equation
obs y1 = <expression>       # observed-channel prediction
\end{lstlisting}

The first line selects one of the five classes. Comma-separated declarations may span multiple declaration statements. Expressions use declared states and parameters; dynamical right-hand sides can also use disclosed exogenous inputs and the time symbol \texttt{t}. Observation maps are functions of the state and parameters. The language includes numerical constants, parentheses, unary signs, and ordinary arithmetic precedence. Fixed numerical choices such as an exponent are part of the hypothesis structure; named parameters are estimated by the fitter.

Every ODE, map, or SDE state requires an evolution equation. A map's \texttt{next} statement specifies $x_{k+1}$ itself, so an incremental model writes \texttt{next x = x + ...}. Every SDE state additionally requires an explicit \texttt{diffusion} statement, whose value multiplies its own independent Wiener increment. Static classes require at least one root state without a \texttt{let} equation. Other static states are evaluated in their \texttt{state}-declaration order, with parents preceding their consumers; the line order of \texttt{let} statements is not a substitute for this dependency order.

Programs are limited to sixteen fitted parameters and 400 expression-tree nodes. The node count includes all evolution, diffusion, and observation expressions. All referenced state and parameter names must be declared. The parser normalizes accepted notation to program text; submissions use a stable parse--print representation. The examples below are checked to round-trip through that representation.

\subsection{Initial conditions, controls, and readouts}

The declaration \texttt{obs y1 = x} is a direct readout. For a non-latent dynamical state it binds preparations, impulses, and clamps on channel \texttt{y1} to state \texttt{x}. An expression such as \texttt{obs y2 = z/2} instead defines a transformed measurement. State names are internal identifiers; matching a state's spelling to a channel name alone does not create a binding.

Each dynamical state without a direct controllable readout has a fitted initial-value parameter \texttt{<state>\_0}. An explicit \texttt{latent} declaration also creates this parameter. These initial values are shared across experiments and count toward the sixteen-parameter limit. Directly controlled states use the experimental preparation, including standard values for coordinates that are not changed.

For a static program, roots are states without defining equations. A direct observation declaration binds an experimental input to its corresponding root, as in \texttt{obs field = field}. Thus static settings are represented by root states and readouts; the \texttt{input} declaration represents dynamical drive signals. Supplied input and do-values are bound before evaluating dependent equations. A do-intervention fixes its target instead of evaluating the target's structural equation. Unbound roots retain their default zero value. Static intermediate states are algebraic quantities and receive no dynamical initial-value parameters.

These rules connect the scientific model to the experiment. A complete static response program therefore declares its input readouts as well as its predicted outputs. SDE fitting in the evaluated profile uses a one-to-one direct state/channel correspondence, while ODE and map programs can introduce hidden states and transformed observation functions.

\subsection{Expression semantics}

Let $\operatorname{spow}(x,r)=\operatorname{sign}(x)|x|^r$ denote signed power. For negative exponents, the magnitude is floored at $10^{-6}$ before exponentiation; at zero the result is zero. Protected division replaces a denominator whose magnitude is below $10^{-9}$ by a same-sign value of magnitude $10^{-9}$, using the positive value at zero. The executable operators are:

\begin{table}[H]
\caption{Core expression operators. Exponents $r$ and $a$ are fixed numerical constants; $k$, thresholds, and other arguments can be expressions containing fitted parameters.}
\label{tab:dsl-operators}
\centering\small
\begin{tabularx}{\linewidth}{@{}l X@{}}
\toprule
Syntax & Meaning \\
\midrule
\texttt{x+y}, \texttt{x-y}, \texttt{x*y}, \texttt{-x} & Addition, subtraction, multiplication, and negation \\
\texttt{x/y} & Protected division as defined above \\
\texttt{pow(x,r)} & $\operatorname{spow}(x,r)$ \\
\texttt{sat(x,k,a)} & $|x|^a/\bigl((|k|+10^{-9})^a+|x|^a+10^{-12}\bigr)$ \\
\texttt{exp(x)} & $\exp(\operatorname{clip}(x,-50,50))$ \\
\texttt{log1p(x)} & $\log(1+|x|)$ \\
\texttt{tanh(x)}, \texttt{sin(x)} & Hyperbolic tangent and sine \\
\texttt{step(x,th,f,g)} & $f$ when $x>\mathrm{th}$, otherwise $g$ \\
\bottomrule
\end{tabularx}
\end{table}

Signed power is distinct from ordinary even integer power: \texttt{pow(-2,2)} evaluates to $-4$, while \texttt{(-2)*(-2)} evaluates to $4$. In particular, a Gaussian factor uses \texttt{exp(-x*x)}. The expression \texttt{pow(x,0)} is the sign function, with value zero at $x=0$.

The parser also accepts the following operators and spellings:
\begin{table}[H]
\caption{Additional accepted operators and spellings. Power aliases retain signed-power semantics; absolute value returns the ordinary magnitude.}
\centering\small
\begin{tabular}{@{}ll@{}}
\toprule
Spelling & Meaning or expansion \\
\midrule
\texttt{x**r} or \texttt{x\string^r} & \texttt{pow(x,r)} \\
\texttt{sqrt(x)} & \texttt{pow(x,1/2)} \\
\texttt{log(x)} & \texttt{log1p(x)} \\
\texttt{cos(x)} & $\sin(x+\pi/2)$ \\
\texttt{sat(x,k)} & \texttt{sat(x,k,1)} \\
\texttt{abs(x)} & $|x|$ \\
\bottomrule
\end{tabular}
\end{table}
Absolute value can equivalently be expressed by \texttt{step(x,0,x,-x)}. The public grammar supplied to agents lists the core operators in Table~\ref{tab:dsl-operators}.

\subsection{Illustrative programs for the five evaluated classes}

\paragraph{ODE with an unobserved state.}
The second state influences the observed dynamics without a direct measurement. Its initial value becomes the fitted parameter \texttt{z\_0}; the observed channel provides the controllable coordinate.
\begin{lstlisting}[basicstyle=\ttfamily\footnotesize,columns=fullflexible,keepspaces=true,breaklines=true,showstringspaces=false,frame=single]
class ODE
state x, z
latent z
input u
param a, b, c, k
d x/dt = -a*x + b*sat(z,k,2)
d z/dt = u - c*z - x*z
obs y1 = x
\end{lstlisting}

\paragraph{Discrete map.}
The right-hand side is the complete next state, with an externally selected input held fixed over a rollout.
\begin{lstlisting}[basicstyle=\ttfamily\footnotesize,columns=fullflexible,keepspaces=true,breaklines=true,showstringspaces=false,frame=single]
class Map
state x
input u
param a, b, k
next x = a*x + b*tanh(u/k)
obs y1 = x
\end{lstlisting}

\paragraph{Stochastic differential equation.}
The two equations separately specify the drift and the state-dependent diffusion amplitude.
\begin{lstlisting}[basicstyle=\ttfamily\footnotesize,columns=fullflexible,keepspaces=true,breaklines=true,showstringspaces=false,frame=single]
class SDE
state x
input u
param a, b, sigma, q, k
d x/dt = -a*x + b*tanh(u)
diffusion x = sigma + q*tanh(x/k)
obs y1 = x
\end{lstlisting}

\paragraph{Structural causal model.}
The input root \texttt{x} is connected to channel \texttt{y1}. The unobserved intermediate \texttt{z} is computed before \texttt{v}, whose readout is \texttt{y2}. An intervention on \texttt{y2} overrides the structural equation for \texttt{v}.
\begin{lstlisting}[basicstyle=\ttfamily\footnotesize,columns=fullflexible,keepspaces=true,breaklines=true,showstringspaces=false,frame=single]
class StructuralDAG
state x, z, v
latent z
param a, b, c
let z = a*x
let v = b*tanh(z) + c*x*x
obs y1 = x
obs y2 = v
\end{lstlisting}

\paragraph{Static response function.}
The input readouts bind the queried field and temperature. The response can then vary with those settings through the declared expression.
\begin{lstlisting}[basicstyle=\ttfamily\footnotesize,columns=fullflexible,keepspaces=true,breaklines=true,showstringspaces=false,frame=single]
class StaticFunction
state field, temperature, response
param a, b, c
let response = a*tanh(field/b) + c*temperature*temperature
obs field = field
obs temperature = temperature
obs signal = response
\end{lstlisting}

\input{appendix-named-models}

%% file: tables/scenario-panel.tex
\begin{tabular}{@{}lrrrrrrr@{}}
\toprule
Setting & Channels & $T$ & Samples/run & $B_1$ & $B_2$ & $B_3$ & $B_4$ \\
\midrule
Mechanics & 2--4 & 1.46--16.9 & 41 & 337 & 89 & 179 & 187 \\
Kinetics & 2--3 & 0.655--12.2 & 41 & 120 & 178 & 322 & 102 \\
Discrete & 3--5 & 30 & 31 & 121 & 246 & 203 & 181 \\
Causal & 5--6 & --- & Point/sample & 160 & 2,135 & 171 & 1,181 \\
Scaling & 4--7 & --- & Point/sample & 215 & 149 & 109 & 88 \\
Ecology & 2--3 & 8 & 25 & 243 & 166 & 192 & 144 \\
Network & 5--6 & 6 & 25 & 236 & 196 & 111 & 182 \\
Network-S1 & 12 & 4 & 25 & 156 & 167 & 140 & 141 \\
Metrology & 4 & --- & Point/sample & 134 & 148 & 179 & 199 \\
Spectroscopy & 5 & --- & Point/sample & 180 & 77 & 171 & 138 \\
SDE & 2 & 1.5 & 31/path & 5,660 & 6,289 & 2,950 & 5,024 \\
\bottomrule
\end{tabular}

%% file: tables/scenario-action-prices.tex
\begin{tabular}{@{}l*{8}{r}@{}}
\toprule
Setting & Observe & Initial & Kick & Clamp & Input & Probe & Assay & Gold \\
\midrule
Mechanics & 1--2 & 2--5 & 4--6$^{\ast}$ & 8--11$^{\ast}$ & 9$^{\ast}$ & 6--14 & 15--40 & --- \\
Kinetics & 1--2 & 2--4 & 4--6$^{\ast}$ & 7--11$^{\ast}$ & 4--10$^{\ast}$ & 16$^{\ast}$ & 15--40 & --- \\
Discrete & 1--2 & 2--4 & 3--5$^{\ast}$ & 5--11$^{\ast}$ & 6--9$^{\ast}$ & 9--10$^{\ast}$ & 15--40 & --- \\
Causal & 1 & --- & --- & 3--8 & --- & --- & --- & --- \\
Scaling & 1 & --- & --- & --- & --- & 4--10 & --- & --- \\
Ecology & 1--2 & 2--4 & 4--6 & 6--9 & 4--10 & 10--14 & 15--40 & --- \\
Network & 1--2 & 3--4 & 3--6 & 8--10 & 5--9 & 8--13 & 15--40 & --- \\
Network-S1 & 1--2 & 3--5 & 4--6 & 6--10 & 6--7 & 7--14 & 15--40 & --- \\
Metrology & 1 & --- & --- & --- & --- & 4 & --- & 8 \\
Spectroscopy & $c(r)$ & --- & --- & --- & --- & $2c(r)$ & --- & --- \\
SDE & $PH$ & --- & --- & --- & --- & $PH$ & --- & --- \\
\bottomrule
\end{tabular}

%% file: appendix-named-models.tex
\subsection{Named models within the limits of a program}
\label{app:coverage}

A program may have at most 16 fitted coefficients, including the initial value of each latent state, and at most 400 expression nodes (1{,}200 in Network-S2; Table~\ref{tab:protocol}). This part measures what the two limits exclude.

\paragraph{Named models.} Each of the 300 laws of the library (Appendix~\ref{app:named-laws}) is a valid program of the modeling language, and 298 are within the limits. The two that are not are the Hodgkin--Huxley equations, with 26 coefficients, and a five-variable model of the circadian oscillator, with 17. A further 68 named models that the library does not contain were written as programs: 38 spatial, 13 delay, 9 SDE and 8 agent-rule models. Of these, 64 are valid programs within the limits. The other four, the Kuramoto--Sivashinsky, Swift--Hohenberg, Cahn--Hilliard and Korteweg--de Vries equations, contain a spatial derivative of order above two, which the modeling language does not have. Altogether 362 of the 368 named models can be submitted.

\paragraph{Programs of the agents.} The true programs have 4 to 15 fitted coefficients, with a median of 7. Of the 35{,}917 programs that the agents sent to the fitter or submitted, 104 (0.3\%), in 64 episodes, were rejected for having more than 16 coefficients, and 62 for their size. The 692 final programs have a median of 9 coefficients, and 99 of them have 16. A final program with 16 coefficients has more than the true program of its world, and it passes less often than a final program with fewer: 13 of the 99 pass (13\%), against 217 of the other 593 (37\%).

%% file: appendix-extensions.tex
\section{Delay, spatial, stochastic-process, reservoir, larger-network, and transfer domains}
\label{app:extensions}

This appendix specifies the delay, spatial, stochastic-process, reservoir, 32-node network, and transfer domains of Table~\ref{tab:domains}. Each uses the common experiment budget, accumulated observations, executable submissions, and held-out behavioral tests. The language adds five model classes: delay differential equations (\texttt{DDE}), spatial field equations (\texttt{PDEGrid}), continuous-time jump processes (\texttt{Jump}), microscopic update rules (\texttt{AgentRule}), and local-energy lattice dynamics (\texttt{LatticeMC}). Reservoir, larger-network, and transfer systems use the existing ODE class. Their held-out long-horizon tests last four times the standard run in the delay and spatial domains and three times in the jump, agent, and lattice domains, against ten times in the ODE and map domains.

\subsection{Delayed feedback}

A delayed system depends on past states as well as its present state. The generator composes instantaneous and delayed responses, choosing the source coordinate, feedback sign, nonlinear response, and lag. A hypothesis can therefore distinguish a delayed influence from an additional instantaneous coupling. The expression \texttt{delay(x,tau)} denotes $x(t-\tau)$, where $\tau$ is constant during an experiment and lies in $[0.01,8]$. Its value can be a fitted parameter.

Each experiment resets the full prehistory. The default history is constant at the preparation; a disclosed \texttt{history\_slopes} context allows linear histories of controlled channels. Subsequent impulses and clamps are retained in delayed reads. The solver uses the method of steps, splitting intervals at interventions and their delayed effects. Clamps apply on half-open intervals. Each complete simulation permits $10^5$ right-hand-side evaluations across its integration intervals; exhaustion is a numerical failure for that parameter trial. Before a world is accepted, a numerical check refines the integration and requires the delayed dependence to affect observable predictions. The following example proposes delayed feedback through an unobserved state:
\begin{lstlisting}[basicstyle=\ttfamily\footnotesize,breaklines=true,frame=single]
class DDE
state x, z
latent z
input u
param a, b, c, tau
d x/dt = -a*x + b*tanh(delay(z,tau)) + u
d z/dt = c*(x-z)
obs y1 = x
\end{lstlisting}

\subsection{Spatially distributed fields}

The three PDE settings describe one-dimensional fields on a physical interval. Transport systems combine generated fluxes, diffusion, losses, forcing, and optional hidden storage. Reaction--diffusion systems combine nonlinear reactions, inter-field dependencies, and distinct diffusivities. Wave systems evolve displacement and rate fields with composed restoring and damping responses, spatial stiffness that may depend on displacement, and optional memory. The generator chooses these responses as expression trees; the agent must construct the field equations.

The operators \texttt{dx(f)}, \texttt{dxx(f)}, and \texttt{diffuse(D,f)} denote $\partial_x f$, $\partial_{xx}f$, and $\partial_x(D\partial_x f)$, respectively. Expressions may depend on the fields, the physical coordinate \texttt{space}, and interval \texttt{length}; total spatial derivative order is at most two. A latent field begins uniformly at its fitted initial-value parameter, so increasing numerical resolution introduces no additional fitted coefficients. For example:
\begin{lstlisting}[basicstyle=\ttfamily\footnotesize,breaklines=true,frame=single]
class PDEGrid
state c, r
latent r
input u
param D, v, a, k
d c/dt = -v*dx(c) + diffuse(D,c) - a*c*r + u
d r/dt = k*(c-r)
obs y1 = c
\end{lstlisting}

An acquisition's \texttt{context} specifies interval length, boundary conditions, initial profiles, forcing profiles, and sensor coordinates. Ordinary lengths range from 0.8 to 1.2; probes extend this to $[0.7,1.7]$. Available boundaries are periodic, zero-gradient Neumann, and zero-value Dirichlet. Preparations and forcing use constant, sinusoidal, or Gaussian profiles. Sinusoids have modes one through four and a chosen phase; Gaussian centers and widths are specified as fractions of interval length, with widths in $[0.08,0.3]$. Periodic Gaussians are smoothly periodized. An optional profile offset lies in $[-1,1]$. A pulse profile specifies the shape of an impulse or whole-field clamp.

The agent chooses four to 32 physical sensor locations. Responses return aligned time, position, and channel rows, retaining early-time measurements as well as later checkpoints. Acquisition price scales with sensor count and duration. Sensor density is distinct from the evaluator's numerical grid: an agent purchases observations, not integration cells.

The simulator uses second-order cell-centered spatial approximations, linear sensor reconstruction, and sparse adaptive backward-differentiation integration with relative and absolute tolerances $10^{-7}$ and $10^{-9}$. Ordinary simulations use 64 cells and hidden tests use 96--256. Before a world is accepted, a check doubles the numerical resolution and compares predictions at the same physical sensors, requiring a discrepancy below 0.01 NRMSE for every held-out condition. Removing spatial operators must materially alter observable behavior. Hidden experiments vary preparations, boundaries, physical length, controls, duration, and numerical resolution.

A simulation permits at most $10^5$ right-hand-side evaluations, including numerical Jacobian probes and all intervention segments. Exceeding this allowance marks the parameter trial as a numerical failure, allowing fitting to continue with other trials. Only requested observation checkpoints are retained. This limit on numerical work is separate from the agent's experiment budget.

\subsection{Finite-activity jump processes}

Jump systems represent discrete events with state-dependent rates. The generator varies nonlinear rate expressions, inhibition, branching, conversion, and joint events. The agent supplies each event's rate and simultaneous state increments. For example, reproduction, loss, and conversion are expressed as:
\begin{lstlisting}[basicstyle=\ttfamily\footnotesize,breaklines=true,frame=single]
class Jump
state n, m
param birth, death, transfer
rate reproduce = birth*n
change reproduce n = 1
rate loss = death*n
change loss n = -1
rate convert = transfer*n
change convert n = -1
change convert m = 1
obs y1 = n
obs y2 = m
\end{lstlisting}

Rates depend on current states and constant inputs. Event increments are evaluated on the pre-event state and applied together. The Gillespie direct method samples event times exactly for this finite-activity class; negative or nonfinite rates invalidate a simulation. Preparations and clamps set nonnegative integer counts, and impulses introduce positive integer counts. Ordinary durations lie in $[0.5,8]$, with probes extending to twelve. The task is to recover event mechanisms from ensembles of count trajectories, rather than match individual random realizations.

\subsection{Microscopic agent rules}

The microscopic setting asks for local update rules that explain population-level measurements. Individuals occupy nodes of a selected directed graph. The generator varies nonlinear response composition, aggregation order, persistent individual types, and hidden memory. A program specifies synchronous updates and optional independent Gaussian noise amplitudes. The operator \texttt{neighbors(f)} averages $f$ over incoming neighbors, returning zero for an isolated node; \texttt{mean(f)} and \texttt{variance(f)} summarize the population. For example:
\begin{lstlisting}[basicstyle=\ttfamily\footnotesize,breaklines=true,frame=single]
class AgentRule
state opinion, memory
latent memory
input u
param a, b, c, r, sigma
types 1
next opinion = tanh(a*opinion+b*neighbors(opinion)-c*memory+u)
next memory = r*memory+(1-r)*opinion
noise opinion = sigma
obs y1 = mean(opinion)
obs y2 = variance(opinion)
\end{lstlisting}

The declaration \texttt{types K} allows one to five persistent types. Each individual's unobserved \texttt{kind} is drawn uniformly from $0,\ldots,K-1$ and can enter its update rule. Noise draws are independent across individuals, states, and updates. Initial individual offsets are centered on the selected population mean; hidden states share a fitted initial value.

Experimental context selects graph edges, population size, initial spread, and intervention nodes. Ordinary populations contain eight to 48 individuals and probes reach 96. Runs last two to 32 updates, or up to 48 in probes. In addition to aggregate observations, a \texttt{deep\_assay} purchases one individual's measured-state trace from an acquired path, priced per recorded time. The trace supplies evidence about microscopic behavior without revealing the individual's hidden type. As with scalar internal assays, it informs the agent's reasoning while the standard fitter uses the aggregate observations.

\subsection{Local-energy lattice systems}

The lattice setting asks an agent to infer local interactions from collective stochastic dynamics. A program specifies a local energy expression on a periodic square lattice of spins $s\in\{-1,+1\}$. The total Hamiltonian sums this expression over all lattice anchors. The operator \texttt{shift(f,i,j)} translates a local expression by integer offsets $i,j\in[-4,4]$. The generator samples interaction supports, orders from two to five spins, signs, strengths, and external-field responses, withholding the interaction list. An illustrative hypothesis is:
\begin{lstlisting}[basicstyle=\ttfamily\footnotesize,breaklines=true,frame=single]
class LatticeMC
state s
input temperature, field
param J, K
energy = -J*s*shift(s,1,0)-K*s*shift(s,0,1)*shift(s,1,1)-field*s
obs y1 = mean(s)
obs y2 = mean(s*shift(s,1,0))
\end{lstlisting}

Each sweep makes $L^2$ random-site visits with replacement on a side-$L$ lattice. A visit proposes a spin flip with probability one half and accepts it according to the Metropolis rule using the full Hamiltonian change and bath temperature. The lazy proposal removes parity periodicity. Initial magnetization sets independent spin probabilities; clamps pin selected sites to either spin value. Measurements include transients rather than presupposing equilibrium. Ordinary side lengths range from four to ten and probes reach sixteen; runs last one to twelve sweeps, extending to 24 in probes. Unlike PDE grid density, lattice size is a physical system-size choice.

\subsection{Stochastic acquisition and evaluation}

Jump, agent, and lattice acquisitions purchase one to 256 independent paths, subject to any disclosed per-action allocation limit. Measurements of the simulated counts and aggregates are exact; randomness arises from the process itself. Acquisition, fitting, and hidden evaluation use separate seeded random streams. Prices scale with path count and duration, additionally with population size or lattice area for microscopic systems.

The fitter minimizes ensemble discrepancies over channel distributions, pairwise channel sums, and increments, using fixed fitting streams. It uses the derivative-free Nelder--Mead method with six random starts, or one supplied starting point, and sixty objective evaluations per start. Hidden evaluation uses 512 independent reference and candidate paths per condition. Error is normalized energy/CRPS excess over the disclosed projections and six time checkpoints; each test group must remain below 0.04. Native distributional errors are retained alongside the binary outcome. These comparisons assess predictive distributions, not pathwise agreement between independent random samples.

\subsection{Reservoirs and larger networks}

The reservoir setting samples directed exchanges between two observed stores and two hidden stores, including a slow store. Internal transfer terms conserve total stored material, while recharge and spill change it. Both the exchange graph and the composed flux functions are unknown. Agents use the same preparations, interventions, scalar internal assays, and numerical fitter as in the other ODE settings.

Network-S2 extends the sparse network construction to 32 measured nodes. A randomly ordered directed cycle provides the backbone, with four to nine additional edges. Five shared fitted coefficients govern the node and edge responses. Its disclosed expression limit is 1,200 nodes, accommodating the larger graph; the sixteen-parameter limit is unchanged. The five- and twelve-node domains retain their separate generation distributions and limits.

\subsection{Transfer problems}

Transfer tests whether a discovered scientific mechanism helps explain a different system. Each study contains a first world, a related world, and a control world. The first and related worlds contain the same generated expression subtree, attached to different variables and embedded in different surrounding equations; their fitted coefficients are independent. The control world changes that subtree while matching the related world's remaining equations, coefficients, preparations, measurement-noise distribution, and hidden conditions. A study is accepted only if the subtree is literally shared and both removing and replacing it change behavior materially.

One agent first investigates the first world. Its last accepted program, canonically printed without comments, is given to the later instances that receive it; its fitted coefficient vector, observations, and conversation are not supplied. Four fresh investigations then compare agents given the program with agents starting without it, on the related and on the control world, in randomized order. The related and control worlds both receive the larger of their independently calibrated budgets, and each comparison with and without the program uses the same world, noise seed, and inference configuration. This paired design distinguishes reuse of a shared mechanism from a general benefit of seeing another submitted model. A third starting point, the true program of the first world, separates the value of a correct mechanism from the agent's success in finding it (Appendix~\ref{app:transfer-true}).

\subsection{Transfer results}
\label{app:extension-results}

Table~\ref{tab:extension-transfer} gives the paired transfer results summarized in Section~\ref{sec:extension-results}; Table~\ref{tab:transfer-true} adds the agents given the true program.

\begin{table}[H]
\caption{Paired transfer results. In each study, an agent given the first world's final program and an agent starting without it investigate the related world, which shares the mechanism, and the control world, which does not, with the same noise seed and budget. ``Studies'' counts the studies whose first investigation produced a submitted program (one GLM study did not), and ``First world'' those in which it was solved. Entries are solve counts.}
\label{tab:extension-transfer}
\centering\footnotesize\setlength{\tabcolsep}{4pt}
\input{tables/extension-transfer.tex}
\end{table}

\input{appendix-transfer-true}

%% file: tables/extension-transfer.tex
\begin{tabular}{@{}lrrrrrr@{}}
\toprule
 & & & \multicolumn{2}{c}{Related world} & \multicolumn{2}{c}{Control world} \\
\cmidrule(lr){4-5}\cmidrule(l){6-7}
Model & Studies & First world & Fresh & Given program & Fresh & Given program \\
\midrule
GPT 6 Astra & 4 & 1 & 2 & 2 & 1 & 0 \\
Claude Fable 5.1 & 4 & 2 & 0 & 2 & 0 & 0 \\
DeepSeek v4.1 Flash & 4 & 0 & 0 & 0 & 0 & 0 \\
Qwen 3.8 Max & 4 & 1 & 1 & 1 & 0 & 0 \\
Muse Spark 1.3 & 4 & 1 & 0 & 1 & 0 & 0 \\
Kimi K3 & 4 & 0 & 1 & 0 & 0 & 0 \\
Grok 4.6 & 4 & 0 & 0 & 0 & 0 & 0 \\
GLM 5.3 & 3 & 1 & 0 & 0 & 0 & 0 \\
Gemini 3.8 Flash & 4 & 0 & 0 & 0 & 0 & 0 \\
\bottomrule
\end{tabular}

%% file: appendix-transfer-true.tex
\subsection{Transfer with the true program}
\label{app:transfer-true}

In the transfer studies (Section~\ref{sec:transfer}), an agent that starts the related world from a program receives the final program of its own investigation of the first world, whether or not that program is correct (Table~\ref{tab:extension-transfer}). A third starting point separates the value of a correct mechanism from the agent's success in finding it: a fresh instance of the agent is given the true program of the first world, with the same instruction that accompanies the agent's own program, and investigates the related world and the control world. All 36 studies have this starting point, which raises the number of related-world episodes from 70 to 106.

Agents given the true program solve the related world in 10 of 36 studies, agents given their own program in 6 of 35, and agents starting fresh in 4 of 35 (Table~\ref{tab:transfer-true}). The advantage of the true program over the agent's own lies in the studies whose first world was not solved. There the agent's own program is not a correct mechanism and leads to a solved related world in 1 of 29 studies, while the true program does in 7 of 30, which include the study without a first program. On the control world, which lacks the mechanism, agents given the true program solve 2 of 36, agents given their own program 0 of 35, and agents starting fresh 1 of 35. In the four transfer problems, the related world is solved by 1, 3, 5 and 1 of the nine agents when they are given the true program, by 0, 2, 4 and 0 when they are given their own program, and by 0, 1, 3 and 0 when they start fresh. The true program is ahead of a fresh start in each of the four problems.

\paragraph{Statistical test.} We pool the related-world episodes of the three starting points by whether the agent holds a correct program of the first world, either its own after solving that world or the true program. Agents with one solve the related world in 15 of 42 episodes, and agents without one, those starting fresh included, in 5 of 64 (exact permutation test within study, two-sided $p=0.002$). The test permutes, inside each study, which of its episodes are solved; 9 studies, those with solved and unsolved episodes and with episodes on both sides of the comparison, are informative. On the control world the counts are 2 of 42 and 1 of 64 ($p=0.44$). A correct program helps on the world that shares the mechanism and not on the world that lacks it.

\paragraph{Episodes ended by the provider.} An episode ends without a decision of the agent when the provider's response contains no tool call, because of an error, a content filter, or a turn that the model completes without calling a tool. Such an episode is scored like any other, on its last accepted submission. In the first run of the true-program starting point this happened in 9 of 72 episodes, in 8 studies. The true-program episodes of these studies, on the related and the control world, were run again from the start, on the same worlds, noise seed and budgets. The studies were chosen for a rerun by the reason for termination alone, not by outcome. Three studies in which the new run again contained such an episode were run once more, and that run is reported whatever its terminations. The counts reported in the paper use the new runs: of the 72 reported episodes, one is ended by the provider, and it is not solved. In the 8 studies, the related world is solved in one first run and in two reported runs, none of them in the study solved in its first run. Of the 140 episodes of the two earlier starting points, one was ended by the provider and is kept. With the first runs in place of the new ones, agents given the true program solve the related world in 9 of 36 studies and the control world in 2 of 36, and the pooled counts on the related world are 14 of 42 and 5 of 64 ($p=0.006$).

\begin{table}[H]
\caption{Related and control worlds of the 36 transfer studies under three starting points: worlds solved, with 95\% Wilson intervals in percent. The last two columns split the related world by whether the agent's first investigation solved the first world. One study has no program from the first investigation, which leaves 35 studies for the first two rows.}
\label{tab:transfer-true}
\centering\footnotesize\setlength{\tabcolsep}{4pt}
\input{tables/transfer.tex}
\end{table}

%% file: tables/transfer.tex
\begin{tabular}{@{}lrrrrrr@{}}
\toprule
 & \multicolumn{2}{c}{Related world} & \multicolumn{2}{c}{Control world} & \multicolumn{2}{c}{Related world, first world} \\
\cmidrule(lr){2-3}\cmidrule(lr){4-5}\cmidrule(lr){6-7}
Agent's starting point & Solved & 95\% interval & Solved & 95\% interval & solved & not solved \\
\midrule
Starting fresh & 4/35 & 5--26 & 1/35 & 1--15 & 2/6 & 2/29 \\
Given its own program & 6/35 & 8--33 & 0/35 & 0--10 & 5/6 & 1/29 \\
Given the true program & 10/36 & 16--44 & 2/36 & 2--18 & 3/6 & 7/30 \\
\bottomrule
\end{tabular}

%% file: tables/configuration.tex
\begin{tabular}{@{}llrrcc@{}}
\toprule
Model & Provider & Context & Output & Effort & Cache \\
\midrule
GPT 6 Astra & OpenAI & 1,050,000 & 128,000 & \texttt{max} & --- \\
Claude Fable 5.1 & Anthropic & 1,000,000 & 128,000 & \texttt{max} & 1h \\
GLM 5.3 & Z.AI & 1,048,576 & 131,072 & \texttt{max} & --- \\
Grok 4.6 & xAI & 500,000 & 450,000 & \texttt{max} & --- \\
Muse Spark 1.3 & Meta & 1,048,576 & 943,718 & \texttt{max} & --- \\
Qwen 3.8 Max & Alibaba & 1,000,000 & 131,072 & \texttt{xhigh} & --- \\
Gemini 3.8 Flash & Google AI Studio & 1,048,576 & 65,536 & \texttt{high} & --- \\
Kimi K3 & Moonshot AI & 1,048,576 & 943,718 & \texttt{max} & --- \\
DeepSeek v4.1 Flash & Fireworks & 1,048,576 & 943,718 & \texttt{max} & --- \\
\bottomrule
\end{tabular}

%% file: tables/model-ids.tex
\begin{tabular}{@{}ll@{}}
\toprule
Label & OpenRouter identifier \\
\midrule
GPT 6 Astra & \texttt{openai/gpt-6-astra} \\
Claude Fable 5.1 & \texttt{anthropic/claude-fable-5.1} \\
GLM 5.3 & \texttt{z-ai/glm-5.3} \\
Grok 4.6 & \texttt{x-ai/grok-4.6} \\
Qwen 3.8 Max & \texttt{qwen/qwen3.8-max-0902} \\
Muse Spark 1.3 & \texttt{meta/muse-spark-1.3} \\
Gemini 3.8 Flash & \texttt{google/gemini-3.8-flash} \\
Kimi K3 & \texttt{moonshotai/kimi-k3} \\
DeepSeek v4.1 Flash & \texttt{deepseek/deepseek-v4.1-flash} \\
\bottomrule
\end{tabular}

%% file: tables/deterministic-errors.tex
\begin{tabular}{@{}lr*{9}{r}@{}}
\toprule
Setting & World & GPT & Claude & GLM & Grok & Muse & Qwen & Gemini & Kimi & DeepSeek \\
\midrule
Mechanics & 1 & 0.458 & 0.389 & 0.259 & 0.597 & 0.357 & 0.469 & 0.495 & 0.844 & 0.369 \\
 & 2 & 0.005 & 0.005 & 0.056 & 0.073 & 0.005 & 0.005 & 0.210 & 0.230 & I \\
 & 3 & 0.293 & 0.208 & 0.130 & 0.512 & 0.432 & 0.210 & 0.542 & 0.633 & 0.730 \\
 & 4 & 3.063 & 3.037 & 3.704 & D & 1.106 & 3.806 & 3.942 & 3.082 & 3.948 \\
Kinetics & 1 & D & 0.429 & D & 0.659 & 0.854 & 0.400 & 0.956 & 0.515 & 0.355 \\
 & 2 & 0.300 & D & 1.405 & D & 109.923 & 0.273 & 2.491 & 0.365 & D \\
 & 3 & 0.691 & 0.813 & 0.866 & 0.849 & D & 0.736 & 0.748 & 0.854 & 0.660 \\
 & 4 & 0.030 & 0.208 & 0.476 & 12.982 & 1.659 & 0.506 & D & 1.116 & 2.494 \\
Discrete & 1 & 0.061 & 0.017 & 0.050 & 0.369 & 0.281 & 0.074 & D & 0.433 & 0.402 \\
 & 2 & 0.364 & 0.276 & 0.491 & 0.495 & 0.631 & 0.502 & 0.527 & D & D \\
 & 3 & 0.019 & 0.226 & 0.285 & 0.765 & 0.598 & 0.279 & 0.494 & D & 0.576 \\
 & 4 & 1.749 & 1.109 & 3.214 & 1.825 & 0.531 & 4.580 & D & 1.475 & 35.185 \\
Causal & 1 & 0.002 & 0.005 & 0.007 & 0.018 & 0.043 & 0.261 & 0.148 & 0.092 & 0.256 \\
 & 2 & 0.012 & 0.014 & 0.017 & 0.062 & 0.094 & 0.116 & 0.005 & 0.128 & 0.156 \\
 & 3 & 0.014 & 0.014 & 0.025 & 0.018 & 0.024 & 0.127 & 0.036 & 0.130 & 0.287 \\
 & 4 & 0.127 & 0.108 & N & 0.129 & 0.147 & N & 0.286 & 0.186 & 0.489 \\
Scaling & 1 & 0.258 & 0.530 & 0.168 & 2.333 & 0.975 & 0.354 & 0.296 & 0.447 & 3.773 \\
 & 2 & 0.020 & 0.017 & 0.363 & 0.363 & 0.363 & 0.363 & 0.029 & 0.419 & 1.366 \\
 & 3 & 0.409 & 0.882 & 0.336 & 0.183 & 0.098 & 0.334 & 0.434 & 0.536 & 1.080 \\
 & 4 & 0.064 & 0.449 & 0.448 & 0.449 & 0.952 & 0.725 & 6.451 & 0.447 & 0.268 \\
Ecology & 1 & 0.091 & 0.087 & 0.094 & 0.171 & 0.195 & 0.066 & 0.722 & 0.391 & 0.230 \\
 & 2 & 0.267 & 0.158 & 1.225 & 0.333 & 0.152 & D & 0.292 & 0.159 & 2.219 \\
 & 3 & 0.153 & 1.675 & 0.171 & 1.228 & 0.806 & 0.131 & 1.104 & 0.602 & D \\
 & 4 & 0.113 & 0.334 & 0.072 & 0.152 & 0.099 & 0.059 & 0.131 & 0.132 & 0.203 \\
Network & 1 & 0.278 & 0.129 & 0.519 & 0.186 & 0.540 & 0.579 & 0.719 & 0.450 & 0.266 \\
 & 2 & 0.088 & 0.069 & 0.254 & 0.075 & 0.202 & 0.732 & 0.139 & 0.144 & 0.422 \\
 & 3 & 0.061 & 0.076 & 0.270 & 0.118 & 0.283 & 0.732 & 0.263 & 0.321 & 0.243 \\
 & 4 & 0.214 & 0.204 & 0.272 & 0.320 & 0.744 & 0.491 & 0.400 & 1.001 & 0.335 \\
Network-S1 & 1 & 0.094 & 0.105 & 0.312 & 0.478 & 0.181 & 0.585 & 0.732 & 0.779 & 0.966 \\
 & 2 & 0.023 & 0.281 & 0.314 & 0.573 & 0.647 & N & 0.663 & 0.238 & 1.782 \\
 & 3 & 0.005 & 0.305 & 0.144 & 0.466 & 0.234 & 0.195 & 1.175 & 0.431 & 1.073 \\
 & 4 & 0.021 & 0.051 & 0.610 & 0.651 & 0.201 & 5.351 & 12.890 & 10.634 & 0.707 \\
Metrology & 1 & 1.147 & 0.179 & 0.876 & 0.282 & 0.759 & I & 2.319 & 0.491 & 0.308 \\
 & 2 & 0.103 & 0.091 & 0.184 & 0.157 & 2.162 & 0.158 & 0.480 & I & I \\
 & 3 & 0.994 & 0.136 & 0.171 & 0.359 & 0.219 & 0.210 & I & 0.250 & 0.161 \\
 & 4 & 1.288 & I & I & 0.491 & 0.129 & 0.090 & I & I & 0.330 \\
Spectroscopy & 1 & 0.012 & 0.009 & I & 0.076 & 0.025 & 0.012 & 0.056 & 0.013 & I \\
 & 2 & 0.009 & 0.015 & 0.008 & 0.023 & 0.015 & I & 0.028 & 0.019 & I \\
 & 3 & 0.012 & 0.088 & 0.072 & 0.231 & 0.080 & 0.145 & 0.249 & I & 0.111 \\
 & 4 & 0.012 & 0.016 & 0.022 & 0.074 & 0.141 & 0.023 & 0.060 & I & I \\
\bottomrule
\end{tabular}

%% file: tables/stochastic-errors.tex
\begin{tabular}{@{}l*{8}{r}@{}}
\toprule
Model & 1: CRPS & 1: Brier & 2: CRPS & 2: Brier & 3: CRPS & 3: Brier & 4: CRPS & 4: Brier \\
\midrule
GPT 6 Astra & 0.1122 & 0.0004 & 0.0327 & 0.0316 & 0.0090 & 0.0014 & 0.0144 & 0.0015 \\
Claude Fable 5.1 & 0.0857 & 0.0006 & 0.0563 & 0.0011 & 0.0236 & 0.0037 & 0.0282 & 0.0006 \\
GLM 5.3 & --- & --- & --- & --- & --- & --- & 0.0294 & 0.0017 \\
Grok 4.6 & --- & --- & 0.3768 & 0.0009 & 0.0377 & 0.0030 & 0.0116 & 0.0011 \\
Muse Spark 1.3 & 0.0710 & 0.0007 & 0.0832 & 0.0003 & 0.0087 & 0.0018 & 0.0426 & 0.0010 \\
Qwen 3.8 Max & 0.0548 & 0.0009 & --- & --- & 0.0088 & 0.0020 & 0.0338 & 0.0014 \\
Gemini 3.8 Flash & 0.4257 & 0.0007 & 0.5982 & 0.0011 & 0.0196 & 0.0011 & 0.0741 & 0.0017 \\
Kimi K3 & 0.0848 & 0.0007 & 0.4539 & 0.0010 & 0.0650 & 0.0018 & --- & --- \\
DeepSeek v4.1 Flash & 0.0655 & 0.0005 & 0.0568 & 0.0012 & 0.0133 & 0.0026 & 0.0140 & 0.0020 \\
\bottomrule
\end{tabular}

%% file: tables/calibration-errors.tex
\begin{tabular}{@{}lrrrr@{}}
\toprule
Setting & World 1 & World 2 & World 3 & World 4 \\
\midrule
Mechanics & 0.0087 & 0.0023 & 0.0471 & 0.0115 \\
Kinetics & 0.0738 & 0.0169 & 0.0751 & 0.0129 \\
Discrete & 0.0103 & 0.0575 & 0.0159 & 0.0069 \\
Causal & 0.0651 & 0.0372 & 0.0294 & 0.0565 \\
Scaling & 0.0474 & 0.0570 & 0.0144 & 0.0223 \\
Ecology & 0.0569 & 0.0196 & 0.0178 & 0.0125 \\
Network & 0.0061 & 0.0071 & 0.0113 & 0.0175 \\
Network-S1 & 0.0041 & 0.0016 & 0.0080 & 0.0062 \\
Metrology & 0.0078 & 0.0100 & 0.0171 & 0.0073 \\
Spectroscopy & 0.0076 & 0.0090 & 0.0119 & 0.0200 \\
\bottomrule
\end{tabular}

%% file: appendix-classical-all.tex
\subsection{Classical methods on every applicable world}
\label{app:classical-all}

This part specifies the classical comparison of Section~\ref{sec:classical} and gives its full results. Every method runs on every world it applies to and on three datasets, and its output is scored on the same hidden tests as an agent's submission. Appendix~\ref{app:classical-matched} reports a smaller comparison, on 44 worlds under one random design, together with the known-structure baseline.

\paragraph{Methods.} \emph{Sparse regression with an operator library} extends SINDy \citep{brunton2016sindy} to the modeling language. Its library holds monomials, signed powers, saturations, hyperbolic tangents, exponentials, logarithms, absolute values, threshold switches, compositions of depth two, couplings through pairwise differences, oscillation in time, and delays. For a differential equation it regresses the integral form \citep{schaeffer2017integral}, so that no time derivative of noisy data is taken; for a map it regresses the next value. Sequentially thresholded ridge regression runs on 20 bootstrap resamples of the purchased experiments, a term is kept when it is selected in at least 60\% of them \citep{fasel2022ensemble}, and the threshold and the number of terms are chosen on a held-out quarter of the purchased experiments. A second variant adds one latent coordinate to the regression. \emph{Polynomial SINDy} is the observed-state SINDy of Table~\ref{tab:classical}: a library of degree 3 on the measured channels with at most 6 terms. \emph{PDE-FIND} \citep{rudy2017pdefind} adds spatial operators to the operator library. \emph{PySR} \citep{cranmer2023pysr} runs 1{,}000 iterations with 8 populations of 32 and a maximum expression size of 30 on at most 500 rows. It regresses the outputs of a static function, each variable of a causal graph on the parents found by intervention, the estimated derivatives of a differential equation, and the next-step differences of a map; the submitted expressions are selected within the limit of 16 coefficients. The \emph{MLP} is trained for 2{,}000 steps on static functions. The \emph{neural ODE} \citep{chen2018neuralode} and the \emph{latent-state neural ODE}, which adds two latent coordinates \citep{dupont2019augmented}, are trained for 1{,}500 steps on ODE and map worlds. The \emph{enlarged grammar search} runs 4{,}096 proposals over 7 generations, eight times the 512 proposals over 3 generations of Table~\ref{tab:classical}. The last two rows of Table~\ref{tab:classical-datasets} score the expressions found by the PySR runs of Table~\ref{tab:classical} with the same pipeline, and rerun the grammar search of Table~\ref{tab:classical}, with its settings, on five of its static and causal worlds.

\paragraph{Scoring.} A method that produces a program submits it as an agent does. The program belongs to the world's class in the modeling language and has at most 16 parameters; the fitter refits it to the method's dataset from six starts of 60 residual evaluations, one of them the method's own coefficients (Section~\ref{sec:fitter}), and the result is scored on the hidden tests. The three neural models are scored on the same hidden tests from their own predictions. None of these methods applies to the 16 stochastic worlds: each produces a deterministic model, and the hidden tests of a stochastic world score the distribution of outcomes of a submitted program.

\paragraph{Datasets.} A classical method cannot choose experiments, so each method receives three datasets, all bought through the experiment interface under the world's budget: the \emph{calibration design}, which is the design of the calibration check (Section~\ref{sec:calibration}) bought until the budget is spent; the experiments GPT purchased before its final submission on the world; and a random design (Section~\ref{sec:setup}). GPT's experiments are those of its one episode. The calibration and random designs are drawn under three noise seeds for the three sparse regressions, PDE-FIND, the MLP and the two neural ODEs. PySR runs under the first noise seed and, on the 12 static-function worlds, under the other two as well; the enlarged grammar search runs under the first, and the last two rows of Table~\ref{tab:classical-datasets} exist for the random design under the first.

A fourth dataset serves as a control: 120 experiments of the calibration design, bought without a budget and with the measurement noise removed. The methods received it on 7 worlds, in 24 runs. One of these runs solves its world, which is also solved on one of the three datasets. The 24 runs are included in Table~\ref{tab:classical-failures} and in no count of solved worlds.

\begin{table}[H]
\caption{Classical methods on every world they apply to. The three dataset columns give the worlds solved on the calibration design, on GPT's experiments and on the random design, each under its first noise seed; the denominator is the number of worlds on which the method ran on that dataset. \emph{Any} counts the worlds solved on any dataset under any noise seed. The last two columns give the worlds that GPT and Claude solve among the method's worlds, each in one episode.}
\label{tab:classical-datasets}
\centering\footnotesize\setlength{\tabcolsep}{3pt}
\input{tables/classical-datasets.tex}
\end{table}

\paragraph{Results.} On the calibration design under its first noise seed, no method solves more than 2 of its worlds, and all methods together solve 3 of the 64 worlds to which a classical method applies (Table~\ref{tab:classical-datasets}); counting all three noise seeds of the calibration design, they solve 6. The best result of any method on any dataset is PySR on GPT's experiments, 5 of 40 worlds; GPT solves 17 of those 40. Pooled over all methods, datasets and noise seeds, the classical methods solve 14 of the 64 worlds. GPT solves 39 of them and Claude 31, each in a single episode per world.

Table~\ref{tab:classical-kinds} separates the kinds of system. The classical successes lie in the static-function and causal worlds, where the task reduces to regressing each output on its inputs: over all datasets and noise seeds, the methods that apply to these worlds reach 10 of the 16, and GPT and Claude reach 9 each in a single episode. On the 48 worlds with dynamics, the first two rows of Table~\ref{tab:classical-kinds}, the classical methods solve 4, GPT 30 and Claude 22. On the 16 stochastic worlds GPT solves 9 and Claude 7.

\begin{table}[H]
\caption{Worlds solved by kind of system. \emph{Classical methods} counts a world when any method solves it on any dataset under any noise seed; each agent has one episode per world.}
\label{tab:classical-kinds}
\centering\footnotesize\setlength{\tabcolsep}{4pt}
\input{tables/classical-kinds.tex}
\end{table}

\paragraph{How the submissions fail.} Over all datasets, the control dataset included, and all noise seeds, 1{,}734 of the 1{,}785 runs of a classical method end in a scored submission that fails, 25 end without a scored submission, and the others pass (Table~\ref{tab:classical-failures}). Of these, 96\% fail the out-of-range tests $T_1$, and 30\% simulate to a divergent trajectory on a hidden test.

\begin{table}[H]
\caption{Failed submissions of the classical methods over all datasets and noise seeds. \emph{Fail $T_1$}: the error on the out-of-range tests is at least 0.1. \emph{Diverge}: a simulation of the submission diverges on a hidden test. A further 25 runs end without a scored submission.}
\label{tab:classical-failures}
\centering\footnotesize\setlength{\tabcolsep}{4pt}
\input{tables/classical-failures.tex}
\end{table}

\paragraph{PDE-FIND on the spatial worlds.} PDE-FIND solves 1 of the 12 spatial worlds; GPT and Claude each solve all 12 (Table~\ref{tab:classical-datasets}). The integral form removes the time derivative, but the spatial operators have to be evaluated on the data: PDE-FIND forms them by finite differences of the noisy readings at the purchased sensors, of which an experiment has at most 32 (Appendix~\ref{app:extensions}), whereas a submitted program is simulated on the evaluator's grid. Of its 79 failed submissions, 59 have a root training error, the square root of $L$ in Eq.~\ref{eq:fit}, of at least 0.1 on their own dataset, and all 79 fail the long-horizon tests $T_3$, which are their worst test group in 67. On the control dataset without measurement noise, PDE-FIND finished on one spatial world; its submission has a root training error of 0.037, passes $T_1$ and $T_2$, and fails $T_3$ with an error of 0.146.

\paragraph{Controls.} Table~\ref{tab:classical-controls} sends the true program through the same pipeline. Scored directly with its true coefficients, the true program passes the hidden tests of all 80 worlds. Fitted from its true coefficients on the calibration design, it passes on all 64 worlds to which a classical method applies. Submitted through the path the classical methods use, with one of the six starts at the true coefficients and the lowest training error kept, it passes on 63, 63 and 64 of the 64 worlds on the three datasets.

\begin{table}[H]
\caption{Controls with the true program on the datasets given to the classical methods. The first row covers all 80 worlds, the others the 64 worlds to which a classical method applies.}
\label{tab:classical-controls}
\centering\footnotesize\setlength{\tabcolsep}{4pt}
\input{tables/classical-controls.tex}
\end{table}

%% file: tables/classical-datasets.tex
\begin{tabular}{@{}lrrrrrrr@{}}
\toprule
 & & \multicolumn{4}{c}{Classical method, solved} & \multicolumn{2}{c}{Agent, solved} \\
\cmidrule(lr){3-6}\cmidrule(lr){7-8}
Method & Worlds & Calibration & GPT's & Random & Any & GPT & Claude \\
\midrule
Sparse regression, operator library & 52 & 2/52 & 2/52 & 1/52 & 4 & 27 & 19 \\
Sparse regression, latent coordinate & 32 & 0/32 & 0/32 & 0/32 & 0 & 14 & 6 \\
Polynomial SINDy & 32 & 0/32 & 0/32 & 0/32 & 0 & 14 & 6 \\
PDE-FIND & 12 & 0/12 & 0/12 & 1/12 & 1 & 12 & 12 \\
PySR & 44 & 0/44 & 5/40 & 0/40 & 7 & 21 & 15 \\
MLP & 12 & 0/12 & 0/12 & 0/12 & 0 & 6 & 6 \\
Neural ODE & 32 & 0/32 & 0/32 & 0/32 & 0 & 14 & 6 \\
Latent-state neural ODE & 32 & 0/32 & 0/32 & 0/32 & 0 & 14 & 6 \\
Grammar search, enlarged & 48 & 1/48 & 0/48 & 1/48 & 2 & 23 & 15 \\
PySR runs of Table~\ref{tab:classical} & 12 & --- & --- & 3/12 & 3 & 6 & 6 \\
Grammar search of Table~\ref{tab:classical} & 5 & --- & --- & 0/5 & 0 & 3 & 4 \\
\midrule
Any method & 64 & 3/64 & 6/64 & 6/64 & 14 & 39 & 31 \\
\bottomrule
\end{tabular}

%% file: tables/classical-kinds.tex
\begin{tabular}{@{}lrrrr@{}}
\toprule
Kind of system & Worlds & Classical methods & GPT 6 Astra & Claude Fable 5.1 \\
\midrule
Dynamics & 36 & 3 & 18 & 10 \\
Spatial fields & 12 & 1 & 12 & 12 \\
Static and causal & 16 & 10 & 9 & 9 \\
\midrule
Worlds a classical method applies to & 64 & 14 & 39 & 31 \\
\midrule
Stochastic & 16 & --- & 9 & 7 \\
\bottomrule
\end{tabular}

%% file: tables/classical-failures.tex
\begin{tabular}{@{}lrrr@{}}
\toprule
Method & Failed submissions & Fail $T_1$ & Diverge \\
\midrule
Sparse regression, operator library & 357 & 339 (95\%) & 138 (39\%) \\
Sparse regression, latent coordinate & 223 & 222 (100\%) & 156 (70\%) \\
Polynomial SINDy & 226 & 226 (100\%) & 181 (80\%) \\
PDE-FIND & 79 & 54 (68\%) & 16 (20\%) \\
PySR & 164 & 152 (93\%) & 16 (10\%) \\
MLP & 85 & 71 (84\%) & 0 (0\%) \\
Neural ODE & 226 & 226 (100\%) & 0 (0\%) \\
Latent-state neural ODE & 226 & 226 (100\%) & 0 (0\%) \\
Grammar search, enlarged & 134 & 134 (100\%) & 17 (13\%) \\
PySR runs of Table~\ref{tab:classical} & 9 & 7 (78\%) & 0 (0\%) \\
Grammar search of Table~\ref{tab:classical} & 5 & 5 (100\%) & 0 (0\%) \\
\midrule
All methods & 1{,}734 & 1{,}662 (96\%) & 524 (30\%) \\
\bottomrule
\end{tabular}

%% file: tables/classical-controls.tex
\begin{tabular}{@{}llr@{}}
\toprule
Control & Dataset & Solved \\
\midrule
True program with its true coefficients, scored directly & --- & 80/80 \\
True program fitted from its true coefficients & Calibration design & 64/64 \\
True program through the submission path & Calibration design & 63/64 \\
 & GPT's experiments & 63/64 \\
 & Random design & 64/64 \\
\bottomrule
\end{tabular}

%% file: tables/classical-comparison.tex
\begin{tabular}{@{}l*{6}{r}@{}}
\toprule
Method & Dynamical & NRMSE & Static & NRMSE & All deterministic & NRMSE \\
\midrule
GPT 6 Astra & 11/24 & 0.133 & 6/12 & 0.084 & 20/40 & 0.099 \\
Claude Fable 5.1 & 6/24 & 0.217 & 6/12 & 0.113 & 15/40 & 0.168 \\
GLM 5.3 & 4/24 & 0.299 & 3/12 & 0.260 & 10/40 & 0.278 \\
Grok 4.6 & 2/24 & 0.504 & 3/12 & 0.256 & 8/40 & 0.361 \\
Muse Spark 1.3 & 2/24 & 0.481 & 4/12 & 0.180 & 9/40 & 0.282 \\
Qwen 3.8 Max & 4/24 & 0.497 & 3/12 & 0.272 & 7/40 & 0.358 \\
Gemini 3.8 Flash & 0/24 & 0.720 & 4/12 & 0.365 & 6/40 & 0.511 \\
Kimi K3 & 0/24 & 0.559 & 2/12 & 0.469 & 3/40 & 0.449 \\
DeepSeek v4.1 Flash & 0/24 & 0.718 & 0/12 & 1.223 & 0/40 & 0.683 \\
Observed-state SINDy & 0/24 & $\infty$ & --- & --- & --- & --- \\
PySR & --- & --- & 3/12 & 0.146 & --- & --- \\
Grammar search & 0/24 & 2.192 & 0/12 & 0.438 & 0/40 & 1.439 \\
Known structure & 16/24 & 0.015 & 8/12 & 0.020 & 27/40 & 0.019 \\
\bottomrule
\end{tabular}

%% file: tables/known-structure-sde.tex
\begin{tabular}{@{}lrrc@{}}
\toprule
World & CRPS excess & Brier excess & Pass \\
\midrule
1 & 0.0170 & 0.0005 & Yes \\
2 & 0.0571 & 0.0012 & No \\
3 & 0.0084 & 0.0020 & Yes \\
4 & 0.0228 & 0.0011 & Yes \\
\bottomrule
\end{tabular}

%% file: tables/error-profiles.tex
\begin{tabular}{@{}lrrrrr@{}}
\toprule
Model & Pass & Finite error & Simulation & Missing channel & No submission \\
\midrule
GPT 6 Astra & 22 & 21 & 1 & 0 & 0 \\
Claude Fable 5.1 & 17 & 25 & 1 & 1 & 0 \\
GLM 5.3 & 11 & 26 & 1 & 2 & 4 \\
Grok 4.6 & 10 & 31 & 2 & 0 & 1 \\
Muse Spark 1.3 & 10 & 33 & 1 & 0 & 0 \\
Qwen 3.8 Max & 9 & 29 & 1 & 2 & 3 \\
Gemini 3.8 Flash & 7 & 32 & 3 & 2 & 0 \\
Kimi K3 & 3 & 34 & 2 & 4 & 1 \\
DeepSeek v4.1 Flash & 2 & 34 & 3 & 5 & 0 \\
\bottomrule
\end{tabular}

%% file: tables/spectroscopy-error-comparison.tex
\begin{tabular}{@{}lrrll@{}}
\toprule
Model & Measurements & Fit cost & Predicted intensity range & Hidden error \\
\midrule
GPT 6 Astra & 157 & 0.00398 & [0.0354, 0.2037] & 0.0121 \\
Claude Fable 5.1 & 162 & 0.00323 & [0.0346, 0.1941] & 0.0157 \\
Kimi K3 & 192 & 2.13896 & [0.0340, 0.0340] & Missing readouts \\
DeepSeek v4.1 Flash & 222 & 4.81083 & [0.0000, 0.0000] & Missing readouts \\
\bottomrule
\end{tabular}

%% file: tables/settings-solved.tex
\begin{tabular}{@{}lrrrrrrrrrr@{}}
\toprule
Domain & GPT & Claude & Muse & Grok & Qwen & GLM & DeepSeek & Gemini & Kimi & All \\
\midrule
\multicolumn{11}{@{}l}{\emph{Dynamics}} \\
Mechanics & 1 & 1 & 1 & 1 & 1 & 1 & 0 & 0 & 0 & 6 \\
Kinetics & 1 & 0 & 0 & 0 & 0 & 0 & 0 & 0 & 0 & 1 \\
Discrete & 2 & 1 & 0 & 0 & 1 & 1 & 0 & 0 & 0 & 5 \\
Ecology & 1 & 1 & 1 & 0 & 2 & 2 & 0 & 0 & 0 & 7 \\
Network & 2 & 2 & 0 & 1 & 0 & 0 & 0 & 0 & 0 & 5 \\
Network-S1 & 4 & 1 & 0 & 0 & 0 & 0 & 0 & 0 & 0 & 5 \\
Network-S2 & 2 & 0 & 0 & 0 & 0 & 0 & 0 & 0 & 0 & 2 \\
Reservoir & 1 & 0 & 0 & 0 & 0 & 0 & 0 & 0 & 0 & 1 \\
Delay & 4 & 4 & 2 & 3 & 4 & 3 & 4 & 3 & 4 & 31 \\
\multicolumn{11}{@{}l}{\emph{Spatial fields}} \\
Transport & 4 & 4 & 4 & 3 & 3 & 4 & 2 & 3 & 4 & 31 \\
Reaction--diffusion & 4 & 4 & 4 & 4 & 3 & 3 & 4 & 4 & 4 & 34 \\
Wave & 4 & 4 & 4 & 2 & 4 & 1 & 3 & 0 & 1 & 23 \\
\multicolumn{11}{@{}l}{\emph{Static and causal}} \\
Causal & 3 & 3 & 3 & 3 & 0 & 3 & 0 & 2 & 1 & 18 \\
Scaling & 2 & 1 & 1 & 0 & 0 & 0 & 0 & 1 & 0 & 5 \\
Metrology & 0 & 1 & 0 & 0 & 1 & 0 & 0 & 0 & 0 & 2 \\
Spectroscopy & 4 & 4 & 3 & 3 & 2 & 3 & 0 & 3 & 2 & 24 \\
\multicolumn{11}{@{}l}{\emph{Stochastic}} \\
Stochastic dynamics & 2 & 2 & 1 & 2 & 2 & 1 & 2 & 1 & 0 & 13 \\
Jump & 3 & 3 & 0 & 1 & 0 & 0 & 2 & 0 & 1 & 10 \\
Agents & 3 & 2 & 0 & 0 & 0 & 0 & 1 & 0 & 0 & 6 \\
Lattice & 1 & 0 & 0 & 0 & 0 & 0 & 0 & 0 & 0 & 1 \\
\bottomrule
\end{tabular}

%% file: tables/selection.tex
\begin{tabular}{@{}lrrrrrr@{}}
\toprule
Model & Budget used (\%) & Truncated & No submission & Final worse & Solve lost & Cost (\$) \\
\midrule
GPT 6 Astra & 93 & 0 & 0 & 10 & 1 & 959 \\
Claude Fable 5.1 & 84 & 1 & 0 & 34 & 2 & 1687 \\
Muse Spark 1.3 & 88 & 7 & 4 & 27 & 1 & 618 \\
Grok 4.6 & 77 & 3 & 2 & 23 & 0 & 397 \\
Qwen 3.8 Max & 66 & 40 & 7 & 34 & 0 & 498 \\
GLM 5.3 & 83 & 20 & 11 & 25 & 1 & 446 \\
DeepSeek v4.1 Flash & 90 & 3 & 2 & 32 & 0 & 43 \\
Gemini 3.8 Flash & 53 & 0 & 1 & 11 & 1 & 83 \\
Kimi K3 & 90 & 12 & 1 & 39 & 1 & 729 \\
\bottomrule
\end{tabular}

%% file: appendix-analyses.tex
\newcommand{\RunToRunShare}{63\%}

\section{Why episodes fail}
\label{app:failures}

Three analyses locate the difficulty of an episode: whether the experiments the agent purchased determine the world (Appendix~\ref{app:experiments-hypothesis}), how the final programs relate to the true programs (Appendix~\ref{app:programs-truth}), and how the agents fare when every latent state is recorded (Appendix~\ref{app:latent-recorded}).

\subsection{Experiments and hypotheses in failed episodes}
\label{app:experiments-hypothesis}

An episode fails either because the purchased experiments do not determine the world or because the agent's hypothesis is wrong. To separate the two, we take the experiments each agent had purchased at its final submission and fit the \emph{true structure} to them: the true program with its coefficients free, fitted by the protocol fitter as a submission is. When this fit passes the hidden tests, the experiments were sufficient and the episode failed on its hypothesis.

Table~\ref{tab:structure-fit} reports the fits for the 490 failed episodes. In 268 of them (55\%; 95\% interval 44--65\% from 10{,}000 resamples of worlds) the true structure passes from the fitter's six random starts. In a further 180 (37\%) it passes when the fit starts from the true coefficients, and that fit has a training error more than 1\% below the fit from random starts: the experiments identify the coefficients, and the random starts do not reach them. The remaining 42 (9\%) are undetermined: either a failing fit reaches the training error of the fit from the true coefficients, so that the experiments do not distinguish the two, or the fit from the true coefficients fails as well. The experiments of failed episodes are nearly as informative as those of solved episodes: the same fits on the 230 solved episodes give 140 (61\%), 84 (37\%) and 6 (3\%).

\begin{table}[H]
\caption{The true structure fitted to the experiments each agent had purchased at its final submission, for failed episodes by agent and, in the last row, for solved episodes. \emph{Passes}: the fit from the fitter's six random starts passes the hidden tests. \emph{Passes from the true coefficients}: only the fit started at the true coefficients passes, with a training error more than 1\% below the fit from random starts. \emph{Not determined}: neither holds. Percentages are of the row's episodes.}
\label{tab:structure-fit}
\centering\footnotesize\setlength{\tabcolsep}{4pt}
\input{tables/structure-fit.tex}
\end{table}

\paragraph{Exchange of structures and experiments.} The same separation follows without the true program. For each failed episode in a world that GPT solved, we fit GPT's final structure to the failing agent's experiments and the failing agent's final structure to GPT's experiments, and we do the same with Claude's final structures in the worlds Claude solved. Each fit runs from random starts and again with one start replaced by the coefficients submitted with that structure, the start an agent may supply (Section~\ref{sec:fitter}). GPT's structure passes on the failing agent's experiments in 193 of 210 episodes (136 from random starts alone), and Claude's in 123 of 132 (59 from random starts alone). The failing agent's structure passes on GPT's experiments in 11 of 197 (7 from random starts alone). Restricted to the 197 failed episodes in worlds GPT solved in which the failing agent has a final program, GPT's structure passes in 180, and the failing agent's structure on GPT's experiments in 11, all of them among the 180 (Table~\ref{tab:exchange}).

\begin{table}[H]
\caption{Fits that pass the hidden tests when structures and experiments are exchanged, over the failed episodes of each agent in worlds that GPT (first two columns) or Claude (last column) solved. A fit counts when it passes from random starts or with one start replaced by the submitted coefficients; the last row counts random starts alone. The first column covers the episodes with a final program.}
\label{tab:exchange}
\centering\footnotesize\setlength{\tabcolsep}{4pt}
\input{tables/exchange.tex}
\end{table}

\subsection{Final programs and the true program}
\label{app:programs-truth}

A world is solved when the final program passes the hidden tests; the form of the program is not scored. This part compares the 692 final programs with the true programs of their worlds.

\paragraph{Comparison.} Each program is reduced to a dependency graph. Its nodes are the scored channels, the inputs and the latent states, and it has an edge from $a$ to $b$ when the equation of $b$ (its drift, its noise amplitude or its readout) reads $a$. A state takes the name of the channel that reads it, so that the graph does not depend on the names an agent chooses, and a latent state from which no scored channel can be reached is removed. The latent states of a final program are matched one to one with those of the true program so that the $F_1$ score of the edges is highest. Every dependence carries its \emph{term types}: linear, power, product, quotient, or the outermost function through which the variable enters, such as a hyperbolic tangent, a delay or a spatial derivative. Two programs have \emph{identical terms} when their equations contain the same terms once the values of the coefficients are dropped, up to the names of states and parameters.

\paragraph{Passing final programs.} Table~\ref{tab:programs-truth} classifies the 230 final programs that pass the hidden tests. Six have identical terms and 224 do not. Of the 224, 171 have the dependency graph of the true program, 28 have additional or missing dependencies, and 25 have a different number of latent states. In 200 of the 230, across all rows of Table~\ref{tab:programs-truth}, at least one dependence shared with the true program has a different set of term types. In 105, the final program lacks a transcendental function that the true program calls (an exponential, a logarithm, a hyperbolic tangent or a sine) and every function from which it can be written exactly; these 105 programs solve 30 of the 54 solved worlds, in 15 of the 20 domains.

\begin{table}[H]
\caption{The 230 final programs that pass the hidden tests, by their relation to the true program of the world. The rows are disjoint; the three rows on additional and missing dependencies hold programs with the number of latent states of the true program. Percentages are of the 230 programs.}
\label{tab:programs-truth}
\centering\footnotesize\setlength{\tabcolsep}{4pt}
\input{tables/programs-truth.tex}
\end{table}

\paragraph{Dependency graph and hidden tests.} The dependency graph of the true program is more frequent among the final programs that pass: 177 of 230 (77\%) have it, against 197 of 462 (43\%) of those that fail, a difference of 34 points (95\% interval 22 to 47, from 2{,}000 resamples of the 80 worlds). Conversely, 177 of the 374 final programs with the true graph pass (47\%), against 53 of the 318 without it (17\%). The relation holds inside worlds. Of the 47 worlds with both passing and failing final programs, the mean $F_1$ score of the edges is higher for the passing programs in 31, equal in 11 and lower in 5.

\paragraph{Failing final programs.} Appendix~\ref{app:domain-results} gives the worst test group of the 321 executable final programs that fail in deterministic worlds: $T_1$ in 154, $T_3$ in 145 and $T_2$ in 22. Of these programs, 163 reproduce the purchased experiments with a root training error, the square root of $L$ in Eq.~\ref{eq:fit}, below 0.1. Their worst test group is $T_1$ in 74, $T_3$ in 78 and $T_2$ in 11. Half of these 321 programs thus reproduce their own data and fail outside it.

\subsection{Latent states recorded}
\label{app:latent-recorded}

Latent states are one of the difficulties the generator sets (Section~\ref{sec:generation}). To separate them from the other differences between worlds, the 18 ODE and map worlds with latent states were run a second time with every latent state recorded on an additional channel and nothing else changed: the program, its coefficients, the budget, the prices and the hidden tests are those of the original world. Four agents investigated these worlds, and each episode is paired with the same agent's episode on the original world.

With the latent states recorded, the agents solve 17 of the 72 episodes, against 8 with the latent states hidden: 10 episodes are gained and 1 is lost (Table~\ref{tab:latent-recorded}). The order of the agents is unchanged: GPT solves 9 worlds (5 with the latent states hidden), Claude 6 (3), DeepSeek 2 (0) and Gemini 0 (0). With every state recorded, 55 of the 72 episodes fail, and the four agents solve 24\% of the episodes, against 75\% (45 of 60) for the same four agents on the deterministic dynamical worlds without a latent state, networks excluded (Section~\ref{sec:case-study}).

\begin{table}[H]
\caption{The 18 ODE and map worlds with latent states, investigated with the latent states hidden and with every latent state recorded on an additional channel. \emph{Gained} and \emph{Lost} count the worlds an agent solves only with the latent states recorded and only with them hidden; \emph{Episodes} counts each agent's episodes with the latent states recorded.}
\label{tab:latent-recorded}
\centering\footnotesize\setlength{\tabcolsep}{4pt}
\input{tables/latent-recorded.tex}
\end{table}

\section{Uncertainty, stability of the ranking, and repeated episodes}
\label{app:ranking-stability}

The first part gives the uncertainty of the percentage of worlds each agent solves and the stability of the ranking (Appendix~\ref{app:ranking}); the second runs further episodes on the same worlds to measure the variation between two episodes of an agent on one world (Appendix~\ref{app:repeats}).

\subsection{Uncertainty of the results and stability of the ranking}
\label{app:ranking}

The benchmark measures an agent's ability in aggregate over the 80 worlds. This part gives the uncertainty of the percentage of worlds each agent solves and shows that the ranking of Table~\ref{tab:headline} is stable under the choice of domains, the settings of the generator, the vocabulary of responses, and the scoring rule.

\paragraph{Intervals.} Table~\ref{tab:uncertainty} resamples the results in two ways: resampling the worlds inside each domain measures the uncertainty for the 20 domains of the benchmark, and resampling whole domains with their worlds measures it for a new sample of domains. With the domains fixed, the standard errors of the percentages solved are 2.6 to 4.8 points, against 4.6 to 5.6 when the 80 worlds are treated as independent, as in Table~\ref{tab:headline}. When domains are resampled, GPT is first in 99.7\% of the 200{,}000 resamples and Claude is ahead of the other seven agents in 99.9\%. GPT solves 10 worlds more than Claude (95\% interval 3 to 18) and Claude 14 more than Muse (7 to 22). A paired $t$ test on the two agents' differences in the 20 domains gives $p=0.021$ and $p=0.003$, in agreement with the exact McNemar tests of Section~\ref{sec:llm-results} ($p=0.013$ and $p=0.001$).

\begin{table}[H]
\caption{Uncertainty of the percentage of worlds solved and stability of the ranking on the 80 worlds. \emph{Worlds solved} counts the worlds out of 80; the intervals are for the percentage solved. Intervals are 95\% percentile intervals from 200{,}000 resamples: of the four worlds inside each domain (domains fixed), and of the 20 domains with their worlds (domains resampled). Ranks are over the nine agents, with ties sharing the mean rank; the rank interval is from the resamples of domains. The last two columns give the range of each agent's rank over all 20 removals of one domain and all 190 removals of two.}
\label{tab:uncertainty}
\centering\footnotesize\setlength{\tabcolsep}{3pt}
\input{tables/uncertainty.tex}
\end{table}

\paragraph{Domains and classes.} GPT remains first and Claude second under every removal of one domain and of two domains (Table~\ref{tab:uncertainty}), and under all 21{,}699 removals of one to five domains. Across six groups of worlds by program class (28 ODE, 4 map, 16 delay and PDE, 12 static-function, 4 causal and 16 stochastic), GPT is first or tied for first in every group, and the ranking inside each group correlates positively with the ranking on the remaining worlds (Spearman 0.39 to 0.84).

\paragraph{Settings of the generator.} The generator varies each of its settings across the 80 worlds, so the sensitivity of the ranking to a setting can be read from the results. We consider 13 recorded attributes of a world: the number of latent states, the number of auxiliary channels, the signal-to-noise ratio, the budget factor, the budget, the reference cost, the numbers of coefficients, state variables and recorded channels, the size and the description length of the true program, the bits of its structural choices, and the presence of an external input. For each attribute we split the worlds in two, once over all worlds and once inside each domain so that both halves contain the same domains. The presence of an external input does not vary inside a domain and is split over all worlds only, which gives 25 splits and 50 halves. GPT is first or tied for first in all 50 halves. The top two agents are never behind any of the other seven, and in 49 of the 50 halves they are strictly ahead. Inside domains, no attribute reorders the agents significantly more than a random split of the same worlds does: the rank correlation between the two halves averages 0.71, against 0.73 for random halves (20{,}000 random splits), and the smallest $p$-value over the 12 attributes is 0.15. The budget factor illustrates the result. In the 40 worlds with the smaller factors GPT solves 23 and Claude 18, in the 40 worlds with the larger factors they solve 25 and 20, and no other agent solves more than 14 in either half. The latent states give the same picture: in the 45 worlds without a latent state GPT solves 32 and Claude 25, and no other agent more than 15; in the 35 worlds with latent states they solve 16 and 13, and no other agent more than 10.

\paragraph{Vocabulary of responses and sensors.} Table~\ref{tab:vocabulary} restricts the results to the worlds whose true program avoids part of the vocabulary. Seven reduced sets of 11 to 72 worlds each exclude one response. GPT and Claude are ahead of the other seven agents in all seven sets; GPT is first in 6 and tied with Claude in the set without hyperbolic tangents. In the 28 worlds whose true program uses at most one response function, GPT solves 21 and Claude 19, and no other agent more than 15. In the 36 worlds that use signed powers and hyperbolic tangents only, the counts are 26, 19 and 14. The ranking also holds for both kinds of sensor: in the 65 worlds in which every sensor reads one state directly the counts are 42, 35 and 23, and in the 15 worlds in which a sensor mixes or saturates states they are 6, 3 and 1.

\begin{table}[H]
\caption{Worlds solved when the results are restricted to worlds whose true program avoids a response, uses few response functions, or has a given kind of sensor. Each row keeps the worlds named in its first column, and \emph{Domains} counts the domains of these worlds; the last column gives the most worlds solved by any of the other seven agents.}
\label{tab:vocabulary}
\centering\footnotesize\setlength{\tabcolsep}{4pt}
\input{tables/vocabulary.tex}
\end{table}

\paragraph{Scoring rule.} A world is solved when $E<0.1$, and $E$ is the worst error over the test groups. Table~\ref{tab:threshold} rescores the episodes at thresholds from 0.02 to 1.0: GPT is first and Claude second at every threshold, and both are ahead of the other seven agents. Table~\ref{tab:test-groups} rescores the 692 final programs with each test group alone, with each pair of groups, and with the mean of the groups in place of the worst; the long-horizon group $T_3$ exists in 68 worlds. GPT is first and Claude second under every rule.

\begin{table}[H]
\caption{Worlds solved when the threshold on $E$ is varied. The protocol uses 0.1. \emph{Solved episodes} counts the solved episodes of all nine agents, out of 720.}
\label{tab:threshold}
\centering\footnotesize\setlength{\tabcolsep}{4pt}
\input{tables/threshold.tex}
\end{table}

\begin{table}[H]
\caption{Worlds solved when the score of a world is computed from other combinations of the test groups $T_1$ (out-of-range), $T_2$ (interventions) and $T_3$ (long horizon). The second column gives the worlds scored under the rule of the row; in the 12 worlds without $T_3$, a rule that includes $T_3$ uses the other groups of the row.}
\label{tab:test-groups}
\centering\footnotesize\setlength{\tabcolsep}{4pt}
\input{tables/test-groups.tex}
\end{table}

\paragraph{One episode per world.} For the same number of episodes, one episode on each of more worlds measures an agent at least as precisely as repeated episodes on fewer worlds. We decomposed the variance of the 720 outcomes into the difficulty of the domain, the difficulty of the world, the interaction of agent and domain, and a part specific to the agent on the world. For one agent, 53\% of the variance of an episode's outcome is the difficulty of the domain (41.5\%; 95\% interval 19 to 56\% when domains are resampled) and of the world within its domain (11.2\%; 3 to 24\%), which every episode on that world shares and which repetition cannot average away. The domains differ in difficulty by more than their worlds do: with the number of agents that solve a world as its score, the mean square between domains is 11.7 times that between the worlds of a domain, and none of 1{,}000{,}000 random reassignments of the 80 worlds to the domains reaches this ratio. The interaction of agent and domain accounts for 6\%, and the remaining 41\% is specific to the agent on the world; pooled over the nine agents, this part has a variance of 0.086. The repeated episodes of Appendix~\ref{app:repeats} measure how much of it varies between episodes: \RunToRunShare{} for DeepSeek and Gemini, and all of it for GPT and Claude, whose variance between episodes (0.125) exceeds 0.086. Three episodes on each of 27 worlds, about as many episodes as one on each of 80 worlds, therefore estimate the percentage of worlds an agent solves with 1.4 to 2.0 times the variance of one episode on each of 80 worlds, and a difference between two agents with 1.0 to 1.7 times the variance; the lower values hold for the variation measured for GPT and Claude, the higher for DeepSeek and Gemini. With 80 worlds and the 20 domains fixed, the standard error of the percentage solved, from the variance pooled over the agents, is 3.7 points and that of a difference between two agents 4.6 points; the differences in Table~\ref{tab:headline} are 12.5 points between GPT and Claude and 17.5 points between Claude and Muse, 2.7 and 3.8 times this standard error. Three episodes on each of the 80 worlds, 1{,}440 episodes more for the nine agents, would lower these standard errors to 2.5 to 3.0 points and 2.7 to 3.5 points. As a measurement of the nine agents, the percentages solved have a reliability, the share of their variance that reflects differences between the agents rather than the sampling of worlds or domains, of 0.94 on these domains and 0.90 for a new sample of domains.

\subsection{Repeated episodes}
\label{app:repeats}

The main results run each agent once on each world. To measure the variation between episodes, four agents were run again under two further noise seeds, so that each has three independent episodes on a world: DeepSeek and Gemini, the two agents with the lowest cost, on all 80 worlds (320 new episodes), and GPT and Claude, the two leading agents, on one world of each domain, chosen before the runs by a hash of the domain's name (80 new episodes). Table~\ref{tab:repeats} compares each further episode with the agent's first episode, and Table~\ref{tab:repeats-worlds} counts the worlds by the number of episodes that solve them.

The percentage of worlds an agent solves changes little from one episode to the next. DeepSeek solves 18, 18 and 16 of the 80 worlds in its three episodes and Gemini 17, 16 and 15, a range of 2.5 points for each agent, against a standard error of 3.7 points for one episode on each of 80 worlds (Appendix~\ref{app:ranking-stability}). World by world, DeepSeek's second and third episodes have the outcome of its first on 68 and on 70 of the 80 worlds and Gemini's on 71 and on 74, and 66 and 68 worlds have the same outcome in all three episodes. Pooled over both agents and both further episodes, a further episode solves the world in 49 of the 70 cases in which the agent's first episode solved it (70\%) and in 16 of the 250 in which it did not (6\%). The outcomes of an agent's three episodes on a world have an intraclass correlation of 0.66 for DeepSeek and 0.69 for Gemini.

On their 20 worlds, GPT solves 11, 13 and 13 in its three episodes and Claude 9, 8 and 12, so GPT solves more worlds than Claude in each episode; Cochran's $Q$ test finds no difference between the three episodes of either agent ($p=0.61$ and $0.16$). World by world, the outcomes of these two agents vary more than those of DeepSeek and Gemini, which fail most worlds in every episode: a further episode has the outcome of the first in 62 of 80 cases (78\%), and solves the world in 34 of the 40 cases in which the first episode solved it (85\%) and in 12 of the 40 in which it did not (30\%), and 12 and 13 of the 20 worlds have the same outcome in all three episodes. The intraclass correlations are 0.45 for GPT and 0.55 for Claude.

The variance between the three episodes of an agent on a world is 0.054 averaged over DeepSeek and Gemini and 0.125 averaged over GPT and Claude, whose solve rates near one half give a single outcome its largest variance. In the decomposition of Appendix~\ref{app:ranking-stability}, the variance specific to an agent on a world, pooled over the nine agents, is 0.086, so the variation between repeated episodes is \RunToRunShare{} of it for DeepSeek and Gemini and exceeds it for GPT and Claude.

Each further episode of GPT and Claude was to use the provider route of the agent's first episode on that world, OpenRouter or the provider's own interface, with the same model and settings. The direct accounts ran out of credit during the runs, and 34 of the 80 episodes were moved from the provider's own interface to OpenRouter. One episode, of Claude, was ended by the provider and is scored on its last accepted submission, as in the main results.

\begin{table}[H]
\caption{Repeated episodes of four agents: DeepSeek and Gemini on the 80 worlds, and GPT and Claude on one world of each domain. The last three columns compare each further episode with the agent's first episode: worlds with the same outcome in both, worlds gained and worlds lost.}
\label{tab:repeats}
\centering\footnotesize\setlength{\tabcolsep}{4pt}
\input{tables/repeats.tex}
\end{table}

\begin{table}[H]
\caption{The worlds of Table~\ref{tab:repeats} by the number of an agent's three episodes that solve them.}
\label{tab:repeats-worlds}
\centering\footnotesize\setlength{\tabcolsep}{4pt}
\input{tables/repeats-worlds.tex}
\end{table}

\section{Named laws in the worlds and in the agents' programs}
\label{app:named-laws-all}

Programs are compared with named laws at two points: before a world is accepted, the generator compares its program with a library of named laws (Appendix~\ref{app:named-laws}), and after the episodes, every program an agent wrote is compared with named laws (Appendix~\ref{app:recall}).

\subsection{Probability that a generated world is a named law}
\label{app:named-laws}

Before a world is accepted (Section~\ref{sec:generation}), its program is compared with a library of 300 named laws, among them Lotka--Volterra, van der Pol, Duffing, Lorenz, SIR and Michaelis--Menten, and a candidate that matches one is discarded. The comparison reduces each equation to the set of its term shapes with the coefficients dropped. It is therefore invariant to the values and signs of the coefficients, to the order of terms and factors, to the expanded or factored form of an expression, and to every relabelling of the state variables. It runs in thirteen generators, those of twelve domains and of the transfer problems (Section~\ref{sec:generation}); they also discard a candidate whose equations are all linear, and a candidate with several state variables of which none depends on another. The comparison rarely acts, because the generators place little probability on any named law. This part derives that probability for one law, reports it for the others, and checks it against the generators.

\paragraph{Worked example: Lotka--Volterra.} The Lotka--Volterra equations are $\dot x_1=\alpha x_1-\beta x_1x_2$ and $\dot x_2=\delta x_1x_2-\gamma x_2$. A raw draw of the kinetics generator, the program it samples for a fresh seed before any check, counts as Lotka--Volterra when its two equations have the term shapes $\{x_1,\,x_1x_2\}$ and $\{x_2,\,x_1x_2\}$ under either labelling of the states, whatever the coefficients. The generator reaches this event through three independent choices. (i)~It draws two states with probability 0.333. (ii)~It adds no threshold-switched term with probability 0.6. (iii)~It writes each equation as an optional decay term plus one to four sampled products of responses, and a sampled product is one of 18{,}060 choice paths, or of 71{,}400 when the equation receives the external input. Summing the probabilities of all paths that produce the two sets of term shapes, and averaging over the presence of the input and the number of coefficients, gives $2.27\times10^{-7}$ for the pair of equations. The product of the three factors is $4.5\times10^{-8}=2^{-24.4}$: one raw draw in 22 million is Lotka--Volterra up to its coefficients. The most probable single path has probability $1.1\times10^{-9}$, and the event is 42 times as probable because many paths write the same equations. The same computation gives $1.1\times10^{-7}$, one raw draw in 9 million, for the van der Pol oscillator.

\paragraph{All two-equation laws.} Table~\ref{tab:named-laws} applies the computation to the 123 two-equation laws of the library, for the kinetics and the mechanics generator. Summed over the laws, the probability that a raw draw is a named law is $5.5\times10^{-5}$ for the kinetics generator and $4.9\times10^{-5}$ for the mechanics generator, one raw draw in 18{,}000 and one in 21{,}000. The generators agree with the computation: 2 million raw draws of each contain 113 and 96 matches, where the computation expects 110 and 97 (Poisson test, $p=0.84$ and $0.94$).

\begin{table}[H]
\caption{Probability that a raw draw of the kinetics and of the mechanics generator is a named law, with the number of matches expected and observed in 2 million raw draws of each generator. The table lists the 22 laws with positive probability; the last row sums over the 123 two-equation laws of the library.}
\label{tab:named-laws}
\centering\scriptsize\setlength{\tabcolsep}{4pt}
\input{tables/named-laws.tex}
\end{table}

\paragraph{Static relations.} A static world has one relation per output, and a short named form is more probable there. For the single-equation laws of the library (the power law, the Langmuir isotherm and the Hill curves), the same computation gives 0.40\% of the raw draws of the causal and scaling generators. In 10 million raw draws, half from each generator, 40{,}266 match against 40{,}479 expected ($p=0.29$; Table~\ref{tab:named-laws-static}). The comparison discards each of them.

\begin{table}[H]
\caption{Single-equation laws in the causal and scaling generators: probability that a raw draw matches, with the number of matches expected and observed in 10 million raw draws, half from each generator.}
\label{tab:named-laws-static}
\centering\footnotesize\setlength{\tabcolsep}{4pt}
\input{tables/named-laws-static.tex}
\end{table}

\paragraph{Recorded raw draws.} Table~\ref{tab:raw-draws} counts the matches among 105{,}600 raw draws recorded from fresh seeds of the thirteen generators. There are 166, all of them single-equation laws in the causal and scaling generators; the 69{,}600 raw draws of the other eleven generators contain none. The recorded draws also show whether a generator repeats a structure. A candidate is discarded when the information recorded for its structural choices is below 48 bits, and across twelve of the thirteen generators (all but Network-S1, for whose twelve-state programs an exact comparison is infeasible), the 76{,}249 recorded raw draws that meet this minimum contain no two draws of one generator with the same structure, in 401 million pairs compared exactly.

\begin{table}[H]
\caption{Recorded raw draws that match a named law, by generator. Rates carry 95\% intervals; for a generator without a match the rate is a one-sided 95\% upper bound.}
\label{tab:raw-draws}
\centering\footnotesize\setlength{\tabcolsep}{4pt}
\input{tables/raw-draws.tex}
\end{table}

\paragraph{Provenance of the evaluated worlds.} The generator draws its candidates from private seed streams derived from a secret that stays with the evaluator, and the evaluated worlds are the first candidates of their streams that pass admission (Section~\ref{sec:generation}). The evaluated worlds were fixed when they were generated, before any agent ran on them, and every one of them is evaluated: no world was selected, exchanged or removed after an agent had run on it. During an episode the agent reaches the world only through the interface for experiments, fits and submissions; the true program, its coefficients and the hidden tests are not sent to it.

\subsection{Named laws in the agents' programs}
\label{app:recall}

This part measures how often an agent proposes a named law and how such programs fare on the hidden tests. It covers the 42{,}088 programs of the 720 episodes: the 35{,}917 that an agent sent to the fitter or submitted, and those it recorded as hypotheses.

\paragraph{Comparisons.} Three comparisons are used, from the narrowest to the broadest. The first is the check that the generator applies to a candidate world, with its library of 300 named laws (Appendix~\ref{app:named-laws}). It applies to ODE, map, SDE, causal and static-function programs of one to six equations. The second, the \emph{looser comparison}, removes differences of notation first: it writes a saturation as a quotient, distributes products over sums, treats numbers, parameters, inputs and time as coefficients, drops additive terms that contain no state, and accepts any one-to-one renaming of the states. It adds textbook models that the library does not contain, among them the Mackey--Glass and Burgers equations, the Ising model, and Lorentzian and Gaussian peaks; it also compares a delay program with its delays removed, a spatial program through its reaction terms, and a static program response by response. The third comparison calls a program \emph{familiar} when it matches a named law under the looser comparison, when every one of its equations is linear in the states, or when two or three states of a larger program form a nonlinear named law.

\paragraph{Named laws during an episode.} Of the 717 episodes with at least one program, the first program matches a named law under the looser comparison in 189, and at least one program does in 449. In 360 episodes the agent fitted such a program and ended with a final program that matches no named law. In 267 of these, the lowest training error the fitter returned for a matching program is above the training error of the final program, each on the experiments purchased at the time of its fit; the median root training errors are 0.162 against 0.070.

\paragraph{Final programs.} The generator's check applies to 307 of the 692 final programs and decides 297 of them within 30 CPU seconds. It names a law in 2, and neither passes the hidden tests. Under the looser comparison, 68 final programs match a named law and 19 of them pass, against 211 of the other 624. Half of the 68 match a linear law (exponential decay, the damped harmonic oscillator, and linear chains of compartments or reactions), and 17 of these 34 pass, 7 in the SDE domain and 10 in the three spatial domains. The other 34 match a nonlinear law, and 2 of them pass, against 228 of the other 658 final programs. Of the 128 familiar final programs, 27 pass (21\%), against 203 of the other 564 (36\%).

\paragraph{Models named in the text.} We searched the text of each episode for the names of ten textbook models: Lotka--Volterra, logistic growth, Michaelis--Menten, SIR, van der Pol, Hill, Monod, Fisher--KPP, Mackey--Glass and Ising. The agents' messages contain text in 516 episodes, and 80 of these name one of the models. In 14 of the 80 the final program is familiar, and none of the 14 passes; in the other 66 it is not, and 22 pass. With the summaries of the agents' reasoning included, 714 episodes have text and 387 name one of the models. Of these, 74 have a familiar final program, of which 10 pass, and 294 a final program that is not familiar, of which 90 pass; the other 19 end without a final program.

\paragraph{True programs.} The generator's check was applied to the true programs of every world whose class it covers: those of 13 domains (the twelve of Appendix~\ref{app:named-laws} and the stochastic-dynamics domain) and the 12 worlds of the transfer problems, 64 in all. It names a law in none of them, and none is linear. The library does not cover the delay, spatial, lattice and agent-rule classes. The 24 true programs of these classes were compared term by term with 70 named models of their own classes, the 59 spatial, delay and agent-rule models of Appendix~\ref{app:coverage} and 11 lattice models, among them the Mackey--Glass, Ikeda, Fisher--KPP, Allen--Cahn, Burgers, FitzHugh--Nagumo and Gray--Scott equations and the Ising family. None of the 24 equals a named model. The 4 true programs of the jump domain are continuous-time jump processes, a class that neither comparison covers.

%% file: tables/structure-fit.tex
\begin{tabular}{@{}lrrrr@{}}
\toprule
 & & \multicolumn{3}{c}{True structure fitted to the agent's experiments} \\
\cmidrule(lr){3-5}
Agent & Episodes & Passes & Passes from the true coefficients & Not determined \\
\midrule
GPT 6 Astra & 32 & 18 (56\%) & 14 (44\%) & 0 (0\%) \\
Claude Fable 5.1 & 42 & 22 (52\%) & 16 (38\%) & 4 (10\%) \\
Muse Spark 1.3 & 56 & 29 (52\%) & 21 (38\%) & 6 (11\%) \\
Grok 4.6 & 57 & 31 (54\%) & 20 (35\%) & 6 (11\%) \\
Qwen 3.8 Max & 57 & 34 (60\%) & 18 (32\%) & 5 (9\%) \\
GLM 5.3 & 58 & 34 (59\%) & 23 (40\%) & 1 (2\%) \\
DeepSeek v4.1 Flash & 62 & 31 (50\%) & 26 (42\%) & 5 (8\%) \\
Gemini 3.8 Flash & 63 & 33 (52\%) & 21 (33\%) & 9 (14\%) \\
Kimi K3 & 63 & 36 (57\%) & 21 (33\%) & 6 (10\%) \\
\midrule
All failed episodes & 490 & 268 (55\%) & 180 (37\%) & 42 (9\%) \\
Solved episodes & 230 & 140 (61\%) & 84 (37\%) & 6 (3\%) \\
\bottomrule
\end{tabular}

%% file: tables/exchange.tex
\begin{tabular}{@{}lrrr@{}}
\toprule
 & Agent's structure, & GPT's structure, & Claude's structure, \\
Failing agent & GPT's experiments & agent's experiments & agent's experiments \\
\midrule
GPT 6 Astra & --- & --- & 2/2 \\
Claude Fable 5.1 & 2/12 & 12/12 & --- \\
Muse Spark 1.3 & 2/23 & 24/26 & 15/16 \\
Grok 4.6 & 0/25 & 22/25 & 14/15 \\
Qwen 3.8 Max & 1/24 & 25/27 & 14/17 \\
GLM 5.3 & 1/24 & 27/27 & 17/17 \\
DeepSeek v4.1 Flash & 2/29 & 29/31 & 22/22 \\
Gemini 3.8 Flash & 3/30 & 26/31 & 20/21 \\
Kimi K3 & 0/30 & 28/31 & 19/22 \\
\midrule
All agents & 11/197 & 193/210 & 123/132 \\
All agents, random starts only & 7/197 & 136/210 & 59/132 \\
\bottomrule
\end{tabular}

%% file: tables/programs-truth.tex
\begin{tabular}{@{}lr@{}}
\toprule
Final program against the true program & Passing final programs \\
\midrule
Identical terms & 6 (3\%) \\
True dependency graph, same term types & 18 (8\%) \\
True dependency graph, a different term type & 153 (67\%) \\
Additional dependencies only & 10 (4\%) \\
Missing dependencies only & 14 (6\%) \\
Additional and missing dependencies & 4 (2\%) \\
Different number of latent states & 25 (11\%) \\
\midrule
All passing final programs & 230 \\
\bottomrule
\end{tabular}

%% file: tables/latent-recorded.tex
\begin{tabular}{@{}lrrrrr@{}}
\toprule
 & & \multicolumn{2}{c}{Solved} & & \\
\cmidrule(lr){3-4}
Agent & Episodes & Latent states hidden & Latent states recorded & Gained & Lost \\
\midrule
GPT 6 Astra & 18 & 5 & 9 & 4 & 0 \\
Claude Fable 5.1 & 18 & 3 & 6 & 4 & 1 \\
DeepSeek v4.1 Flash & 18 & 0 & 2 & 2 & 0 \\
Gemini 3.8 Flash & 18 & 0 & 0 & 0 & 0 \\
\midrule
All four agents & 72 & 8 & 17 & 10 & 1 \\
\bottomrule
\end{tabular}

%% file: tables/uncertainty.tex
\begin{tabular}{@{}lrrrrrr@{}}
\toprule
 & & \multicolumn{2}{c}{Percentage solved, 95\% interval} & \multicolumn{3}{c}{Rank} \\
\cmidrule(lr){3-4}\cmidrule(lr){5-7}
Agent & Worlds solved & Domains fixed & Domains resampled & 95\% interval & 1 removed & 2 removed \\
\midrule
GPT 6 Astra & 48 & 52.5--67.5 & 46.3--73.8 & 1--1 & 1 & 1 \\
Claude Fable 5.1 & 38 & 40.0--55.0 & 31.3--63.8 & 2--2 & 2 & 2 \\
Muse Spark 1.3 & 24 & 25.0--36.3 & 15.0--47.5 & 3--8 & 3--5 & 3--6 \\
Grok 4.6 & 23 & 22.5--35.0 & 15.0--43.8 & 3--7 & 3.5--5.5 & 3--6 \\
Qwen 3.8 Max & 23 & 22.5--35.0 & 13.8--45.0 & 3--8.5 & 3--6 & 3--6.5 \\
GLM 5.3 & 22 & 21.3--33.8 & 13.8--42.5 & 3--8 & 3.5--6 & 3--7 \\
DeepSeek v4.1 Flash & 18 & 17.5--27.5 & 8.8--38.8 & 3--9 & 7--9 & 4.5--9 \\
Gemini 3.8 Flash & 17 & 16.3--26.3 & 7.5--36.3 & 5.5--9 & 7--9 & 6.5--9 \\
Kimi K3 & 17 & 17.5--26.3 & 7.5--37.5 & 5--9 & 7--9 & 7--9 \\
\bottomrule
\end{tabular}

%% file: tables/vocabulary.tex
\begin{tabular}{@{}lrrrrr@{}}
\toprule
Worlds retained & Worlds & Domains & GPT & Claude & Best of the other seven \\
\midrule
All worlds & 80 & 20 & 48 & 38 & 24 \\
\midrule
No signed powers & 42 & 12 & 29 & 26 & 18 \\
No saturations & 47 & 13 & 34 & 26 & 18 \\
No exponentials & 64 & 19 & 38 & 30 & 20 \\
No logarithms & 61 & 19 & 42 & 33 & 21 \\
No hyperbolic tangents & 11 & 7 & 5 & 5 & 3 \\
No sines & 71 & 19 & 44 & 33 & 22 \\
No threshold switches & 72 & 19 & 45 & 36 & 24 \\
\midrule
At most one response function & 28 & 7 & 21 & 19 & 15 \\
At most two response functions & 49 & 15 & 35 & 27 & 18 \\
Signed powers and hyperbolic tangents only & 36 & 9 & 26 & 19 & 14 \\
\midrule
Every sensor reads one state directly & 65 & 18 & 42 & 35 & 23 \\
A sensor mixes or saturates states & 15 & 5 & 6 & 3 & 1 \\
\bottomrule
\end{tabular}

%% file: tables/threshold.tex
\begin{tabular}{@{}rrrrr@{}}
\toprule
Threshold on $E$ & Solved episodes & GPT & Claude & Best of the other seven \\
\midrule
0.02 & 115 & 30 & 23 & 13 \\
0.05 & 165 & 38 & 27 & 20 \\
0.075 & 196 & 45 & 31 & 21 \\
0.1 & 230 & 48 & 38 & 24 \\
0.15 & 282 & 56 & 44 & 30 \\
0.2 & 320 & 59 & 49 & 35 \\
0.3 & 381 & 66 & 59 & 43 \\
0.5 & 462 & 71 & 68 & 50 \\
1.0 & 548 & 74 & 72 & 66 \\
\bottomrule
\end{tabular}

%% file: tables/test-groups.tex
\begin{tabular}{@{}lrrrr@{}}
\toprule
Score of a world & Worlds & GPT & Claude & Best of the other seven \\
\midrule
Worst of the groups ($E$) & 80 & 48 & 38 & 24 \\
$T_1$ alone & 80 & 57 & 46 & 32 \\
$T_2$ alone & 80 & 68 & 64 & 50 \\
$T_3$ alone & 68 & 46 & 39 & 32 \\
Worst of $T_2$ and $T_3$ & 80 & 58 & 50 & 40 \\
Worst of $T_1$ and $T_3$ & 80 & 50 & 38 & 26 \\
Worst of $T_1$ and $T_2$ & 80 & 55 & 46 & 27 \\
Mean of the groups & 80 & 58 & 46 & 33 \\
\bottomrule
\end{tabular}

%% file: tables/repeats.tex
\begin{tabular}{@{}lrlrrrr@{}}
\toprule
 & & & & \multicolumn{3}{c}{Against the first episode} \\
\cmidrule(lr){5-7}
Agent & Worlds & Episode & Solved & Same outcome & Gained & Lost \\
\midrule
GPT 6 Astra & 20 & First & 11 & --- & --- & --- \\
 &  & Second & 13 & 16 & 3 & 1 \\
 &  & Third & 13 & 14 & 4 & 2 \\
\midrule
Claude Fable 5.1 & 20 & First & 9 & --- & --- & --- \\
 &  & Second & 8 & 19 & 0 & 1 \\
 &  & Third & 12 & 13 & 5 & 2 \\
\midrule
DeepSeek v4.1 Flash & 80 & First & 18 & --- & --- & --- \\
 &  & Second & 18 & 68 & 6 & 6 \\
 &  & Third & 16 & 70 & 4 & 6 \\
\midrule
Gemini 3.8 Flash & 80 & First & 17 & --- & --- & --- \\
 &  & Second & 16 & 71 & 4 & 5 \\
 &  & Third & 15 & 74 & 2 & 4 \\
\bottomrule
\end{tabular}

%% file: tables/repeats-worlds.tex
\begin{tabular}{@{}lrrrrrr@{}}
\toprule
 & & \multicolumn{4}{c}{Worlds solved in} & \\
\cmidrule(lr){3-6}
Agent & Worlds & 0 of 3 & 1 of 3 & 2 of 3 & 3 of 3 & Same outcome in all three \\
\midrule
GPT 6 Astra & 20 & 4 & 3 & 5 & 8 & 12 \\
Claude Fable 5.1 & 20 & 6 & 6 & 1 & 7 & 13 \\
DeepSeek v4.1 Flash & 80 & 55 & 9 & 5 & 11 & 66 \\
Gemini 3.8 Flash & 80 & 58 & 6 & 6 & 10 & 68 \\
\bottomrule
\end{tabular}

%% file: tables/named-laws.tex
\begin{tabular}{@{}lrrrrrr@{}}
\toprule
 & \multicolumn{3}{c}{Kinetics generator} & \multicolumn{3}{c}{Mechanics generator} \\
\cmidrule(lr){2-4}\cmidrule(lr){5-7}
Named law & Probability & Expected & Observed & Probability & Expected & Observed \\
\midrule
Griffith positive feedback & $2.2\times10^{-5}$ & 44.2 & 42 & $2.0\times10^{-5}$ & 39.2 & 41 \\
Reversible mass action & $1.7\times10^{-5}$ & 34.7 & 32 & $1.5\times10^{-5}$ & 30.7 & 30 \\
Damped harmonic oscillator & $1.0\times10^{-5}$ & 20.4 & 24 & $9.0\times10^{-6}$ & 18.0 & 16 \\
Helmholtz oscillator & $1.8\times10^{-6}$ & 3.7 & 7 & $1.5\times10^{-6}$ & 3.1 & 5 \\
Harmonic oscillator & $1.5\times10^{-6}$ & 3.0 & 3 & $1.3\times10^{-6}$ & 2.7 & 0 \\
Hertz contact oscillator & $1.1\times10^{-6}$ & 2.2 & 4 & $9.8\times10^{-7}$ & 2.0 & 2 \\
Duffing oscillator & $5.4\times10^{-7}$ & 1.1 & 0 & $4.6\times10^{-7}$ & 0.91 & 1 \\
Double-well oscillator & $1.6\times10^{-7}$ & 0.32 & 1 & $1.3\times10^{-7}$ & 0.27 & 0 \\
Lanchester mixed combat & $1.5\times10^{-7}$ & 0.31 & 0 & $1.3\times10^{-7}$ & 0.26 & 1 \\
van der Pol oscillator & $1.1\times10^{-7}$ & 0.22 & 0 & $8.7\times10^{-8}$ & 0.17 & 0 \\
Lotka--Volterra & $4.5\times10^{-8}$ & 0.091 & 0 & $3.6\times10^{-8}$ & 0.073 & 0 \\
Holling type II predator--prey & $2.9\times10^{-8}$ & 0.059 & 0 & $2.4\times10^{-8}$ & 0.047 & 0 \\
Autocatalysis with decay & $2.7\times10^{-8}$ & 0.053 & 0 & $2.1\times10^{-8}$ & 0.043 & 0 \\
Cubic autocatalysis & $1.2\times10^{-8}$ & 0.024 & 0 & $9.5\times10^{-9}$ & 0.019 & 0 \\
SIS epidemic & $8.4\times10^{-9}$ & 0.017 & 0 & $6.4\times10^{-9}$ & 0.013 & 0 \\
Hunt--Crossley contact & $7.5\times10^{-9}$ & 0.015 & 0 & $6.0\times10^{-9}$ & 0.012 & 0 \\
Ross--Macdonald & $4.8\times10^{-9}$ & 0.0095 & 0 & $3.4\times10^{-9}$ & 0.0068 & 0 \\
Tessier growth & $4.3\times10^{-9}$ & 0.0086 & 0 & $3.3\times10^{-9}$ & 0.0065 & 0 \\
SI epidemic & $3.9\times10^{-9}$ & 0.0079 & 0 & $3.2\times10^{-9}$ & 0.0063 & 0 \\
Moser growth & $3.5\times10^{-9}$ & 0.0069 & 0 & $2.8\times10^{-9}$ & 0.0056 & 0 \\
Monod batch growth & $2.5\times10^{-9}$ & 0.0051 & 0 & $2.0\times10^{-9}$ & 0.0041 & 0 \\
Duffing--van der Pol & $1.9\times10^{-9}$ & 0.0037 & 0 & $1.4\times10^{-9}$ & 0.0028 & 0 \\
\midrule
Any of the 123 two-equation laws & $5.5\times10^{-5}$ & 110.5 & 113 & $4.9\times10^{-5}$ & 97.4 & 96 \\
\bottomrule
\end{tabular}

%% file: tables/named-laws-static.tex
\begin{tabular}{@{}lrrr@{}}
\toprule
Named law & Probability & Expected & Observed \\
\midrule
Power law & $1.5\times10^{-3}$ & 15{,}211 & 15{,}046 \\
Hill binding ($n=2$) & $1.2\times10^{-3}$ & 11{,}855 & 11{,}742 \\
Langmuir isotherm & $1.0\times10^{-3}$ & 10{,}061 & 10{,}030 \\
Hill binding ($n=4$) & $3.2\times10^{-4}$ & 3{,}222 & 3{,}306 \\
Hill repression & $1.3\times10^{-5}$ & 131 & 142 \\
\midrule
Any single-equation law & $4.0\times10^{-3}$ & 40{,}479 & 40{,}266 \\
\bottomrule
\end{tabular}

%% file: tables/raw-draws.tex
\begin{tabular}{@{}lrrr@{}}
\toprule
Generator & Raw draws & Matches & Rate (\%) \\
\midrule
Mechanics & 17{,}000 & 0 & $<$0.018 \\
Kinetics & 18{,}000 & 0 & $<$0.017 \\
Discrete & 17{,}000 & 0 & $<$0.018 \\
Causal & 18{,}000 & 67 & 0.37 (0.29--0.47) \\
Scaling & 18{,}000 & 99 & 0.55 (0.45--0.67) \\
Ecology & 6{,}000 & 0 & $<$0.050 \\
Network & 4{,}000 & 0 & $<$0.075 \\
Metrology & 3{,}000 & 0 & $<$0.10 \\
Spectroscopy & 1{,}500 & 0 & $<$0.20 \\
Reservoir & 600 & 0 & $<$0.50 \\
Transfer problems & 1{,}500 & 0 & $<$0.20 \\
Network S1 & 800 & 0 & $<$0.37 \\
Network S2 & 200 & 0 & $<$1.5 \\
\midrule
All thirteen generators & 105{,}600 & 166 & 0.16 (0.13--0.18) \\
\bottomrule
\end{tabular}